\pdfoutput=1
\documentclass[titlepage, a4paper, 11pt]{article}
\usepackage{geometry}
\usepackage{amsmath}
\usepackage{amsfonts}
\usepackage{type1cm}
\usepackage{bm}
\usepackage[dvipdfmx]{graphicx}
\usepackage{titlesec}
\usepackage{multirow}
\usepackage{booktabs}
\usepackage{setspace}

\usepackage{hyperref}
\usepackage{cite}
\usepackage{xcolor}
\hypersetup{
    colorlinks=true,
    citecolor=black,
    linkcolor=black,
    urlcolor=black,
}

\makeatletter
\@addtoreset{equation}{section}
\@addtoreset{figure}{section}
\@addtoreset{table}{section}
\makeatother

\titleformat*{\section}{\huge\bfseries}

\begin{document}
\title{
    \fontsize{22pt}{30pt}\selectfont
    \centering
    \vspace{-90pt}
    Toyota Technological Institute Ph.D. Thesis\\
    \vspace{100pt}
    Integrating Heterogeneous Domain Information into Relation Extraction: 
    A Case Study on Drug-Drug Interaction Extraction
    \vspace{170pt}
}
\author{
    \fontsize{17pt}{3pt}\selectfont
    \begin{minipage}{10truecm}
        \centering
        December 2022\\[50pt]
        Masaki Asada\\[50pt]
        Computational Intelligence Laboratory
    \end{minipage}
}
\date{}
\maketitle
\abstract{

Relation extraction from the literature is a crucial task in natural language processing.
It is very important to extract relationships between named entities from news articles and academic papers in various fields and to summarize these relations into structured data. However, the number of digital documents on the Internet is increasing daily, and it is impossible to perform relation extraction for all of the texts manually.
Therefore, there is a need to study on automatic relation extraction from the literature using machine learning.

The development of deep neural networks has improved representation learning in various domains, including textual, graph structural, and relational triple representations.
This development opened the door to new relation extraction beyond the traditional text-oriented relation extraction.
However, research on the effectiveness of considering multiple heterogeneous domain information simultaneously is still under exploration, and if a model can take an advantage of integrating heterogeneous information, it is expected to exhibit a significant contribution to many problems in the world.

This thesis works on Drug-Drug Interactions (DDIs) from the literature as a case study and realizes relation extraction utilizing heterogeneous domain information.
Drugs are related to information about various aspects, including textual information, information of molecular structures, categorical information, and related protein information. 
Therefore, extracting DDIs from the literature is a suitable target to verify the effectiveness of considering heterogeneous domain information in the relation extraction task.

First, a deep neural relation extraction model is prepared and its attention mechanism is analyzed.
Next, a method to combine the drug molecular structure information and drug description information to the input sentence information is proposed, and the effectiveness of utilizing drug molecular structures and drug descriptions for the relation extraction task is shown.

Then, in order to further exploit the heterogeneous information, drug-related items, such as protein entries, medical terms and pathways are collected from multiple existing databases and a new data set in the form of a knowledge graph (KG) is constructed.
A link prediction task on the constructed data set is conducted to obtain embedding representations of drugs that contain the heterogeneous domain information.

Finally, a method that integrates the input sentence information and the heterogeneous KG information is proposed.
The proposed model is trained and evaluated on a widely used data set, and as a result, it is shown that utilizing heterogeneous domain information significantly improves the performance of relation extraction from the literature.
Overall, this study demonstrates the importance of considering heterogeneous domain items in the information extraction task beyond the text-oriented information extraction.
}
\newpage
\fontsize{12pt}{27.5pt}\selectfont
\thispagestyle{empty}
\paragraph{Acknowledgments}~

First, and most of all, I would like to express my gratitude to my supervisor, Prof. Yutaka Sasaki.
I have been indebted to him for more than five years, and it is impossible to get to this point without his continual support.
He gave me sincere guidance for my research plan, for writing papers, and for preparing presentation slides.

I would also like to thank Assoc. Prof. Makoto Miwa, who has given me a lot of extremely important advice in my research on neural relation extraction.
He is a great motivator in my decision to pursue a Ph.D. degree.
During our daily discussions, he gave me very thoughtful comments that made my Ph.D. research even greater.

I would like to thank my co-supervisor, Prof. Kazuo Hotate for his great advice in the introduction part of my research presentation.
I would like also to thank the members of my Ph.D. thesis review committee. I would like to thank Prof. Yukihiro Motoyama for giving me insightful comments about drugs.
I would like to thank Prof. Norimichi Ukita for the great discussions about deep learning methods that helps me to improve my Ph.D. thesis.
I would like to thank Prof. Yoshimasa Tsuruoka for his extremely insightful comments on natural language processing.

Another big thank you goes to all of my colleagues at Toyota Technological Institute, who have supported me.
As a senior member of the Ph.D. program, Tomoki Tsujimura provided me with a lot of advice on my research. We had a lot of fun conversations outside of work.

My deepest gratitude goes to my family for their endless love and support during all these years. I would like to thank my parents for supporting my decisions with generosity.

\newpage
\fontsize{12pt}{27.5pt}\selectfont
\pagestyle{empty}
\tableofcontents

\clearpage
\listoftables
\clearpage
\listoffigures

\clearpage
\setcounter{page}{1}
\pagestyle{plain}

\section{Introduction}
\subsection{Motivations and Objectives}
\begin{figure}[t]
\begin{center}
\includegraphics[width=0.95\linewidth]{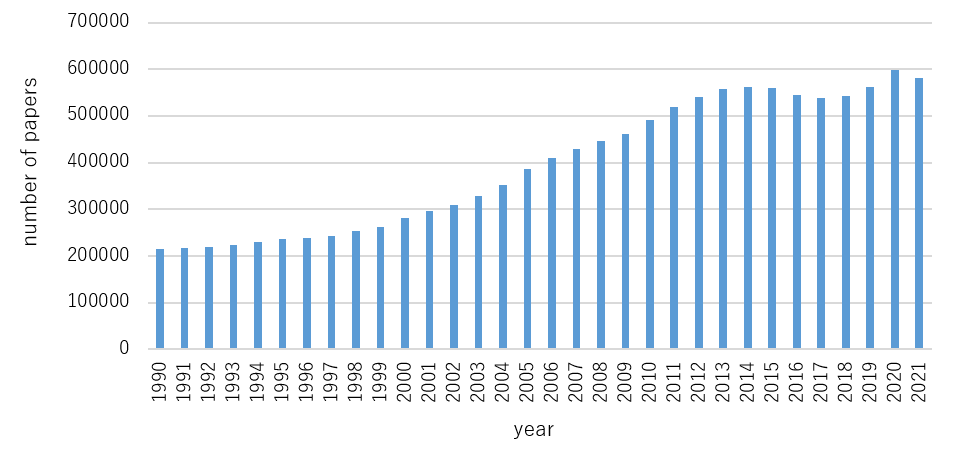}
\end{center}
\caption[Yearly number of papers with the MeSH category of \textit{"Chemicals and Drugs"} in PubMed]%
{Yearly number of papers with the MeSH category of \textit{``Chemicals and Drugs''} in the medicine and biomedical journal literature search engine PubMed. The number of papers is generally increasing year by year.}
\label{fig:intro:num_papers}
\end{figure}

Relation extraction from the literature is a crucial task in natural language processing.
It is very important to extract relationships between named entities from the literature and summarize these relations into structured data. However, the number of digital documents on the Internet is increasing daily, and it is impossible to conduct relation extraction for all of the texts manually. Figure~\ref{fig:intro:num_papers} shows the yearly number of papers in the medicine and biomedical journal literature search engine PubMed.
The number of papers is generally increasing year by year, and there is a need to study on automatic relation extraction from the literature using machine learning.

It was considered very difficult to utilize different kinds of information at the same time.
However, with the advent of deep learning~\cite{ian2016deeplearning}, it has become possible to represent each attribute as fixed-size numerical vectors using methods such as word2vec~\cite{mikolov2013distributed} and node2vec~\cite{grover2016node2vec}.
As for text representation, the advent of Bidirectional Encoder Representations from Transformer (BERT)~\cite{devlin-etal-2019-bert} has made it more powerful.
Although these developments have opened the door to new relation extraction beyond the traditional text-oriented relation extraction, the research on simultaneously considering multiple heterogeneous domain information is still under exploration.
The author believes that the new techniques will provide clues for the integrated use of heterogeneous domain information.
This thesis elaborates to establish deep learning-based methods that can comprehensively handle heterogeneous information relevant to mentions in relation extraction from the literature.

\begin{figure}[t]
\begin{center}
\includegraphics[width=0.9\linewidth]{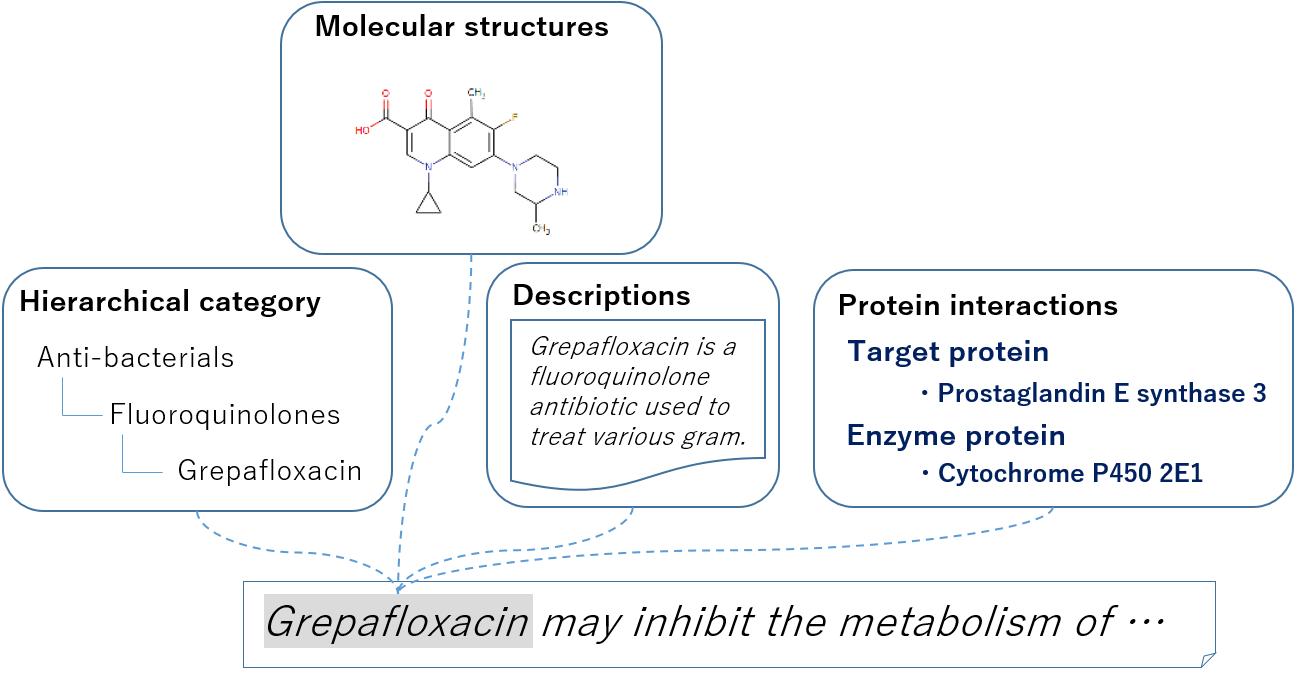}
\end{center}
\caption[Heterogeneous domain information about a mention in the literature]%
{Heterogeneous domain information about a mention in the literature}
\label{fig:intro:hetero_info}
\end{figure}

Similar to this motivation is multimodal learning~\cite{ngiam2011multi} and multi-view learning~\cite{zhao2017multi-view}.
Multimodal learning is a method that simultaneously considers information in two or three different modalities, such as text and image, or text and audio.
Multi-view learning is an emerging direction in machine learning which considers learning with multiple views to improve the generalization performance. Multi-view learning is also known as data fusion or data integration from multiple feature sets.
Figure~\ref{fig:intro:hetero_info} shows an example of the heterogeneous domain information of an entity in the literature. 
The examples of drug entities in the literature show multiple kinds of domain information related to drugs. Drugs contain various kinds of information, such as molecular structures, descriptions, and categorical hierarchies.
This thesis further develops multimodal learning and incorporates multi-view learning to integrate a lot of domain information in the unified vector space, and utilizes them to conduct the relation extraction task.

One thing which must be clear here is that the relation extraction task consists of two parts: (i) recognition of named entities in input sentences and (ii) extraction of the relation between identified named mentions, and this thesis focuses only on the relation extraction part.
This is because many heterogeneous factors need to be considered in order to extract relations from the literature.

Extracting relations between drugs from the literature is an appropriate problem in verifying the effectiveness in considering heterogeneous domain information.
Medicines help us feel better and stay healthy. But sometimes drug interactions can cause problems. 
Drug interactions can be classified into the following three broad categories~\cite{drug-interactions}:
\begin{description}
    \item[Drug-drug interaction] A reaction between two (or more) drugs. 
    \item[Drug-food interaction] A reaction between a drug and a food or beverage.
    \item[Drug-condition interaction] A reaction that occurs when taking a drug while having a certain medical condition.
\end{description}
This thesis chooses the drug-drug interaction (DDI) extraction task as a case study.
When considering DDIs, it is necessary to refer to the descriptions in drug articles for achieving evidence-based medicine~\cite{sackett1997evidence}; therefore it is important to extract DDIs from the literature.
Hence, the author emphasizes that the goal is to determine whether a sentence in the articles represents that two drugs interact or not. It is not a goal to determine whether the two drugs actually interact. For instance, when a sentence just mentions a pair of drug entities without justifying their interaction, the model cannot extract the DDI of the two drugs in the context of the sentence.

\subsection{Contributions}
The contributions of this thesis are shown as follows:
\begin{itemize}
    \item To utilize heterogeneous domain information in relation extraction.
    \item To construct a new benchmark data set in the form of knowledge graphs to use heterogeneous information.
    \item To propose a neural architecture that integrally uses sentence information and domain information.
    \item To propose methods to utilize the heterogeneous domain representation of drugs trained from the constructed data set to extract DDIs from texts.
    \item To evaluate the advantage of the proposed method using drug-drug interaction extraction as a case study.
    \item To show that several kinds of heterogeneous drug information are complementary and their effective combinations can largely improve the DDI extraction performance.
    \item To show that the proposed neural architecture can effectively utilize heterogeneous domain information for DDI extraction. 
\end{itemize}

\subsection{Thesis Structure}
The remainder of this thesis is organized as follows.

Chapter 2 provides the novelties of the thesis based on an overview of relation extraction from the literature. 
First, how the relation extraction corpus was created and the relation extraction task settings, evaluation metrics and data set statistics are described. Then, the history of how machine learning-based automatic relation extraction from the literature is described.
Finally, a new approach that integrates heterogeneous domain information into relation extraction tasks is introduced.

Chapter 3 provides a base neural model for  relation extraction as a preliminary, and reviews the effectiveness of the method of employing an attention mechanism in the CNN-based model on relation extraction tasks. 
This model is a text-oriented relation extraction model and is not capable of domain information.

Chapter 4 reviews the study on relation extraction from texts considering molecular structures of drugs in addition to the context of input sentences.
A baseline model with the word2vec algorithm and Convolutional Neural Networks (CNNs) is addressed in this section.
Since this model cannot deal with heterogeneous domain information, the following chapters develop relation extraction models that can consider heterogeneous domain information.

Chapter 5 proposes a relation extraction model that can 
consider two types of domain information: the molecular structure information of the drug and the description information of the drug in a case study.
In addition, the relation extraction model is applied to the drug-protein interaction (DPI) extraction task which was held in the BioCreative VII shared task. 
This section and later sections employ Bidirectional Encoder Representations from Transformer (BERT)~\cite{devlin-etal-2019-bert} as the baseline model.
This chapter includes work from the published paper \cite{asada-bioinformatics}.

Chapter 6 builds a novel data set in the form of a Knowledge Graph (KG) to consider further heterogeneous information about drugs.
The link prediction task is conducted on the created heterogeneous KG and the embedding representations of the drugs are obtained.
This chapter is based on the published work~\cite{Asada2021-uv}.

Chapter 7 utilizes the heterogeneous KG embeddings of the drugs for DDI extraction task.
The effectiveness of combining the input sentence representation and KG representation is described.
This chapter is based on~\cite{asada-bioinformatics2}.

Chapter 8 summarizes the findings in this thesis and concludes the thesis.
Finally, the further developments of the model and the future plan are discussed.

\section{New Approach to Relation Extraction: Unified Use of Heterogeneous Domain Information}

This chapter firstly provides an overview of relation extraction tasks.
Then a new approach that integrates heterogeneous domain information into relation extraction tasks is introduced.

\subsection{Relation Extraction: An Overview}
\subsubsection{Task Definition}

With the expanding use of the Internet, a large amount of the literature becomes available every day in the form of news articles, research publications, question answering forums, and social media.
It is important to develop techniques for extracting information automatically from these documents and summarizing relations into structured data.
The entity like the person and the organization is the most basic unit of information. Occurrences of entities in a sentence are often linked through well-defined relations; e.g., occurrences of person and organization in a sentence may be linked through relations such as {\em employed\_at}.
The task of relation extraction is to identify such relations automatically.~\cite{DBLP:journals/corr/abs-1712-05191}

\begin{figure}[t]
\begin{center}
\includegraphics[width=0.95\linewidth]{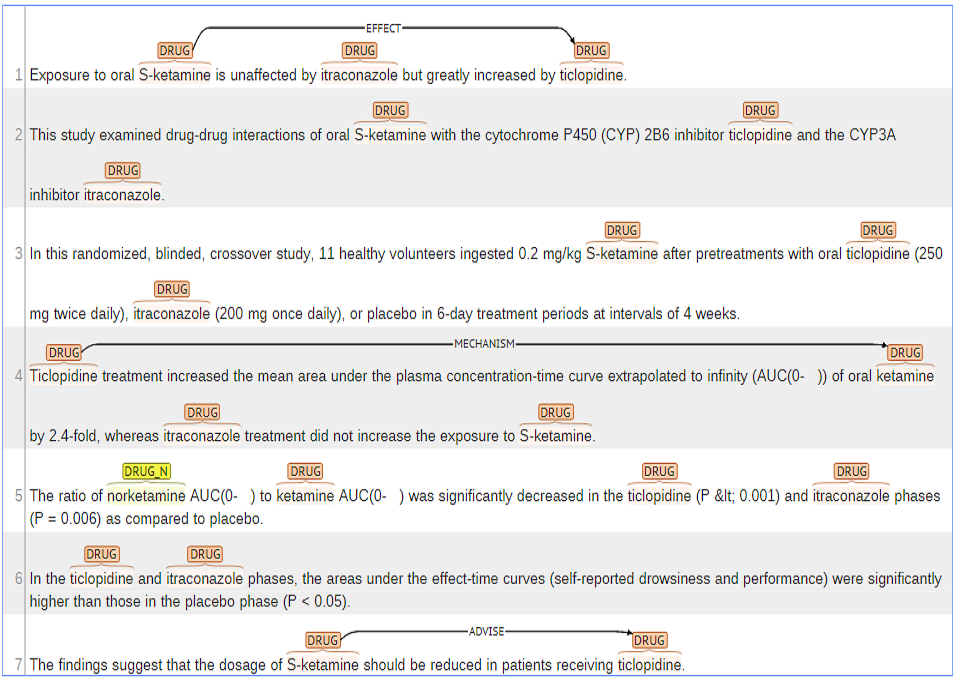}
\end{center}
\caption[The visualization of some instances in the DDIExtraction-2013 data set]%
{The visualization of some instances in the DDIExtraction-2013 data set}
\label{fig:relate:semeval2013_brat}
\end{figure}

Figure~\ref{fig:relate:semeval2013_brat} shows some examples of annotated sentences.
These sentences are included in the abstract of the PubMed article.
The first sentence indicates that there are three entities in the sentence, and the entity \textit{S-ketamine} and the entity \textit{ticlopidine} have relation \textit{Effect}.
The aim of the relation extraction task is to classify the given all possible entity pairs into well-defined relations.

\subsubsection{Data Sets and Corpora}
The construction of a common data set is essential for the development of research of automatic relation extraction from the literature and several data sets have been created to assist the development of relation extraction methods.
These data sets were constructed by trusted annotators who are experts in their fields.

For the relation extraction task in the general field, Automatic Content Extraction (ACE)~\cite{doddington2004automatic} data set which contains named entities, relations, and events in various languages, mostly from news articles was constructed. Another widely used data set is CoNLL04~\cite{carreras-marquez-2004-introduction}, which also contains entities and relations for the general field.
SemEval-2010 task 8~\cite{hendrickx-etal-2010-semeval} data set was created a few years later and used widely to develop relation extraction models. 

For the biomedical field, protein-protein interaction was the first target for relation extraction task and several protein-protein interaction extraction data sets such as AIMed~\cite{bunescu2005comparative} and BioInfer~\cite{pyysalo2008comparative} data sets were created.
Drug-drug interaction (DDI) extraction has also been studied.
DDIExtraction-2013~\cite{segura-bedmar-etal-2013-semeval} data set was constructed and it is used as a standard benchmark data set.
BioCreative VI ChemProt~\cite{Krallinger2017OverviewOT} data set targeted Chemical-Protein relations and is a large relation extraction data set with multiple fine-grained relations.
Later, BioCreative VII DrugProt~\cite{miranda2021overview} task data set was constructed, which contains 13 types of drug-protein interactions.

This thesis focuses on the DDIExtraction-2013 task and investigates the effectiveness of heterogeneous domain information about drugs.

\subsubsection{Evaluation Metrics}
Evaluation is relation-oriented and based on the standard Precision, Recall, and micro-averaged F-score metrics.
Precision is also called Positive Predictive Value (PPV) and Recall is also called Sensitivity.
$TP~(True~Positives)$ correspond to the number of pairs correctly identified by the model as having a relation, $TN~(True~Negatives)$ correspond to the total number of pairs correctly identified as not having a relation.
$FP~(False~Positives)$ correspond to the total number of missed pairs which have a relation and $FN~(False~Negatives)$ correspond to the total number of missed pairs which do not have a relation.
Table~\ref{tab:sec2:contingency_bin} shows the contingency table of TP, FP, TN and FN for a binary classification task.

\begin{table}[h!]
    \centering
    \begin{tabular}{c|c|c|c|}
        \multicolumn{2}{c}{} & \multicolumn{2}{c}{Ground Truth}  \\\cline{3-4}
        \multicolumn{1}{c}{} & & Relation & No Relation \\\cline{2-4}
    \multirow{2}{*}{Prediction} & Relation & TP & FP \\\cline{2-4}
     & No Relation & FN & TN \\\cline{2-4}
    \end{tabular}
    \caption{Contingency table of true positives (TP), false positives (FP), true negatives (TN) and false negative (FN) for binary DDI classification}
    \label{tab:sec2:contingency_bin}
\end{table}

Using these statistics, The evaluation metrics Precision (P), Recall (R) and F1-score (F1) are calculated as follows:

\begin{equation}
    P = \frac{TP}{TP+FP}
\end{equation}
\begin{equation}
    R = \frac{TP}{TP+FN}
\end{equation}
\begin{equation}
    F1 = \frac{2 P R}{P+R}
\end{equation}

The above described statistics and metrics are typically used in the case of binary classification. However, the DDIExtraction-2013 data set has four types of positive labels.
When the F-score is used as an evaluation metric for multi-class classification, there are two types of F-scores, micro-averaged and macro-averaged F-scores.
While micro-averaged F-score is calculated by constructing a global contingency table and then calculating precision and recall, macro-averaged F-score is calculated by first calculating precision and recall for each type and then taking the average of these results.
The micro- and macro-averaged metrics for Precision, Recall and F-score are described in the following equations, where $c$ indicates the interaction category.

\begin{equation}
    P_{micro} = \frac{\sum_c TP_c}{\sum_c TP_c + \sum_c FP_c},~~~~~R_{micro} = \frac{\sum_c TP_c}{\sum_c TP_c + \sum_c FN_c}
\end{equation}
\begin{equation}
    F1_{micro} = \frac{2 P_{micro} R_{micro}}{P_{micro}+R_{micro}}
\end{equation}
\begin{equation}
    P_{macro} = \frac{1}{|c|}\sum_c P_c,~~~~~R_{macro} = \frac{1}{|c|}\sum_c R_c
\end{equation}
\begin{equation}
    F1_{macro} = \frac{2 P_{macro} R_{macro}}{P_{macro}+R_{macro}}
\end{equation}
In this task, the micro-averaged F1-score is used as the main evaluation metric.

\subsubsection{Conventional Approaches Dealing with Relation Extraction from the Literature}

\paragraph{CNN-based Relation Extraction}
The first application of deep Convolutional Neural Networks (CNNs) to the relation extraction task was the CNNs model proposed by Zeng et al.~\cite{zeng-etal-2014-relation}. CNN models take a candidate relation instance that is represented by the word embeddings and position embeddings as input.
Then, the input is fed to the convolutional layer in order to extract features with filters of different sizes. Subsequently, the pooling layer performs down-sampling on the feature maps, which provides two advantages: firstly, it can identify the most relevant and essential local features; and secondly, it reduces the computational complexity which the reduced resolution. Finally, for classifying the relation type, the feature vector generated by pooling is fed into a fully connected softmax layer. The experiments showed that this CNN-based method can capture the semantic information and relative position of words better than conventional Support Vector Machines (SVMs)~\cite{boser1992training} methods.

\paragraph{RNNs for Relation Extraction}
Recurrent Neural Networks (RNNs) mainly address sequential data. Each output depends on the previous values in this structure, which is similar to the human memory mechanism. Generally, these networks are composed of an input layer, a hidden layer that is connected to itself, and an output layer. 
In the standard formulation, RNNs suffer from exploding and vanishing gradient problems. To address these problems, Long Short-Term Memory (LSTM)~\cite{hochreiter1997lstm} units and Gated Recurrent Units (GRU)~\cite{cho-etal-2014-properties} networks model the hidden layer with a memory cell that controls the extent of what is to be forgotten (and to be incorporated) given the input, the previous state, and the current state. These networks can explicitly learn when to forget information and when to store it, which is fit for an input of arbitrary length. Undoubtedly, these variants are efficient in the relation extraction task.
As for RNN-based approaches, Socher et al.~\cite{socher-etal-2012-semantic} presented a novel relation classification model that learns vectors in the syntactic tree path that connects two nominals to determine their semantic relationship.


\paragraph{Transformers and BERT Family for Relation Extraction}
Many neural network-based methods have been proposed since Liu et al.~\cite{liu2016drug} first tacked the DDI extraction task with a neural network-based method.
RNNs and CNNs have been widely used for NLP tasks, but both methods have the problem that it is difficult to capture relationships between distant words in a sentence.
Transformer~\cite{vaswani2017attention}, a model that utilizes the attention mechanism, has been newly devised and has been believed to solve this problem.
The attention is a mechanism for calculating the importance of each word in a sentence.
Given a single input sentence, a contextualized word representation can be obtained by a self-attention mechanism that calculates the importance of each word in the sentence.
While CNNs are unable to consider context outside the window size and RNNs are thought to be difficult to capture relationships between distant words, Transformers are thought to overcome these shortcomings.

Bidirectional Encoder Representation from Transformers (BERT)~\cite{devlin-etal-2019-bert} is a Transformer model that is pre-trained on a large unlabeled corpus.
The approach of fine-tuning the pre-trained BERT shows high performance on various downstream tasks, including the relation extraction task.

\subsection{Integrating Heterogeneous Domain Information into Relation Extraction}

This section outlines the novel approach to integrating heterogeneous domain information into the relation extraction task. 
This thesis develops in three steps.

\paragraph{Using individual domain information}

\begin{figure}[t]
\begin{center}
\includegraphics[width=0.95\linewidth]{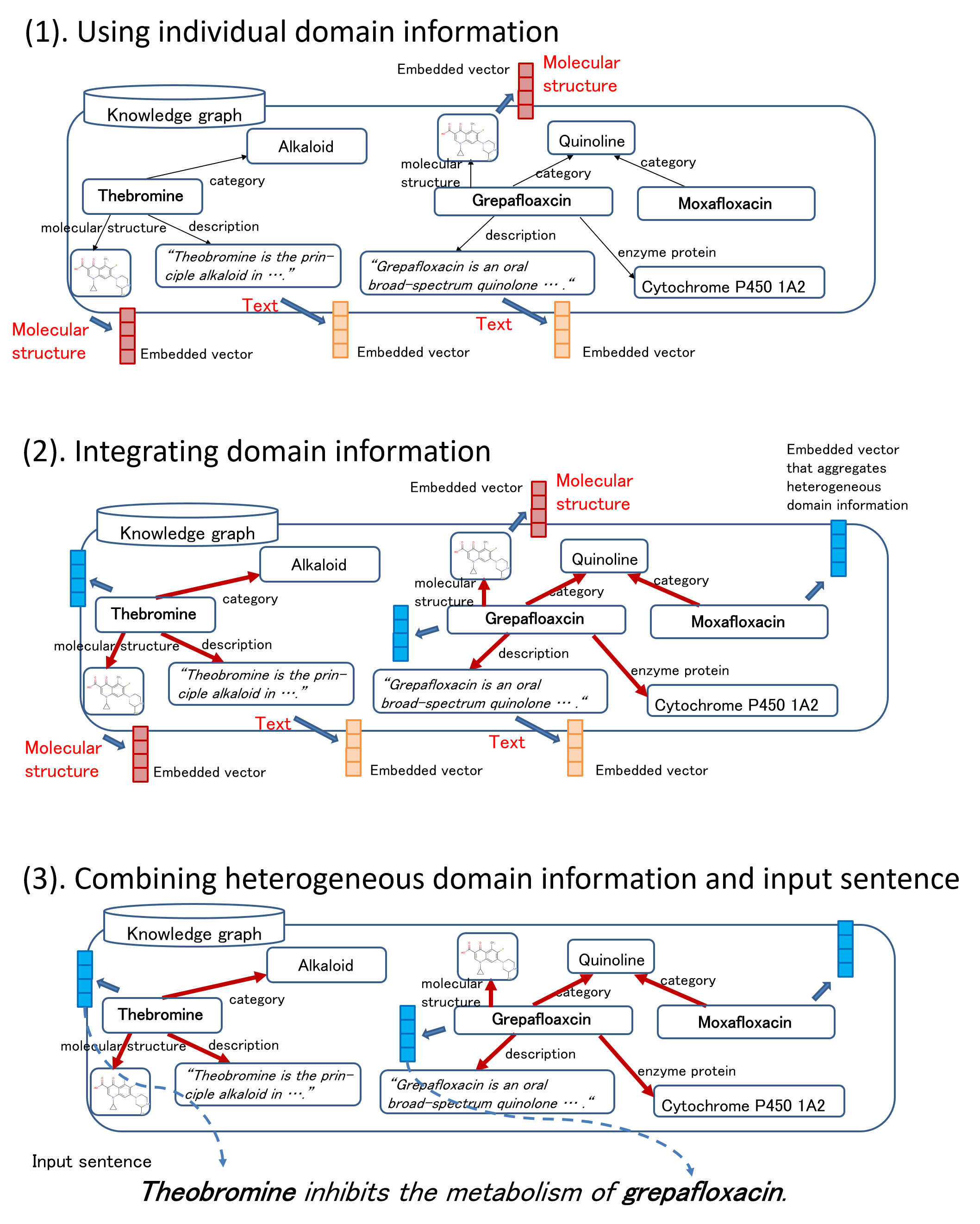}
\end{center}
\caption{The outline of integrating heterogeneous domain information into the relation extraction}
\label{fig:sec2:approach}
\end{figure}

Figure~\ref{fig:sec2:approach}~(1) shows the first step of the proposed method.
Here, the relation extraction that utilizes the drug-related information from external databases are addressed.
The molecular structures and descriptions of drugs are represented as vectors separately by a neural network model.
Details based on case studies are provided in Chapter 4 and Chapter 5.

\paragraph{Integrating Heterogeneous domain information}

Figure~\ref{fig:sec2:approach}~(2) shows the second step of the development.
Heterogeneous drug-related knowledge, including the molecular structures, is represented in a unified vector space. Diverse information is integrated and the vectors that aggregate heterogeneous domain information are obtained.


\paragraph{Combining heterogeneous domain information and input sentences}

Figure~\ref{fig:sec2:approach}~(3) displays the main novelty of this thesis.
In this figure, the vectors of aggregated heterogeneous domain information are effectively utilized for relation extraction from the literature.
The heterogeneous domain information is represented in a unified vector space.
This approach enables to integrate heterogeneous knowledge in relation extraction.


\section{Relation Extraction from Texts with Attention CNNs}\label{sec:attention_cnn}
As a preliminary, this chapter introduces a baseline relation extraction model with an attention mechanism for a CNN-based relation extraction model~\cite{asada-etal-2017-extracting} and reviews the usefulness of the attention mechanism.
This verification is the preliminary experiment to a method that utilizes heterogeneous information.
The technique of preprocessing data set and detailed statistics of the DDIExtraction-2013 article, which are common in subsequent chapters in this thesis, are discussed.

The overview of this chapter is as follows:
\begin{itemize}
    \item This chapter provides an attention mechanism that can boost the performance on CNN-based DDI extraction.
    \item The DDI extraction model with the attention mechanism achieves the performance with an F-score of 69.12\%, which is competitive with other state-of-the-art DDI extraction models when 
    the performance without negative instance filtering~\cite{chowdhury2013fbk} is compared.  
\end{itemize}

\subsection{Background}

For the DDI extraction, deep neural network-based methods have recently drawn a considerable attention~\cite{liu2016drug,zhao2016drug,sahu2018drug}. 
Deep neural networks have been widely used in the NLP field. They show high performance on several NLP tasks without requiring manual feature engineering. 
CNNs and RNNs are often employed for network architectures.
Among these, CNNs have the advantage that they can be easily parallelized and the calculation is thus fast with recent Graphical Processing Units (GPUs). 

Liu et al.~\cite{liu2016drug} showed that a CNN-based model can achieve a high accuracy on the task of DDI extraction. Sahu et al.~\cite{sahu2018drug} proposed an RNN-based model with the attention mechanism to tackle the DDI extraction task and show state-of-the-art performance.
The integration of an attention mechanism into a CNN-based relation extraction is proposed by Wang et al.~\cite{wang2016relation}. This is applied to a general domain relation extraction  task SemEval-2010 Task 8~\cite{hendrickx-etal-2010-semeval}. Their model showed state-of-the-art performance on the task. 
CNNs with attention mechanisms, however, are not evaluated on the task of DDI extraction. 

This chapter reviews a attention mechanism that is integrated into a CNN-based DDI extraction model).
The attention mechanism extends the attention mechanism by \cite{wang2016relation} in that it deals with anonymized entities separately from other words and incorporates a smoothing parameter. 
A CNN-based relation extraction model is implemented and the mechanism into the model is integrated. 
The model is evaluated on the DDIExtraction-2013 data set~\cite{segura-bedmar-etal-2013-semeval}.

\subsection{Preprocessing}\label{sec:preprocessing}
When three or more drug mentions appear in an input sentence, the sentence is duplicated for each drug mention pair. Specifically, if an input sentence contains $n$ drug mentions, $\binom{n}{2}$ input sentences with different drug mention pairs are prepared.

According to the settings of many existing methods~\cite{liu2016drug, sahu2018drug, lim2018drug}, before tokenizing an input sentence, the mentions of the target drugs in the pair are replaced with DRUG1 and DRUG2 according to their order of appearance. 
The other mentions of drugs are replaced with DRUGOTHER.

Table~\ref{tab:sec3:preprocess} shows an example of preprocessing when the input sentence \textit{Exposure to oral S-ketamine is unaffected by itraconazole but greatly increased by ticlopidine} is given with a target entity pair. By performing preprocessing, it is possible to prevent the DDI extraction model to be specialized for the surface forms of the drugs in a training data set and to perform DDI extraction using the information of the whole context. 

\begin{table*}
\centering
\begin{tabular}{ccp{10cm}}\hline
Entity1 & Entity2 & Preprocessed input sentence\\\hline
\textit{S-ketamine} & \textit{itraconazole} & \textit{Exposure to oral \textbf{DRUG1} is unaffected by 
\textbf{DRUG2} but greatly increased by DRUGOTHER.} \\\hline
\textit{S-ketamine} & \textit{ticlopidine} & \textit{Exposure to oral \textbf{DRUG1} is unaffected by DRUGOTHER but greatly increased by \textbf{DRUG2}.} \\\hline
\textit{itraconazole} & \textit{ticlopidine} & \textit{Exposure to oral DRUGOTHER is unaffected by \textbf{DRUG1} but greatly increased by \textbf{DRUG2}.} \\\hline
\end{tabular}
\caption[An example of the preprocessed input sentence]{An example of preprocessing on the sentence \textit{Exposure to oral S-ketamine is unaffected by itraconazole but greatly increased by ticlopidine} for each target pair.}
\label{tab:sec3:preprocess}
\end{table*}

\subsection{Methods}

\begin{figure}[t]
\begin{center}
\includegraphics[width=0.9\linewidth]{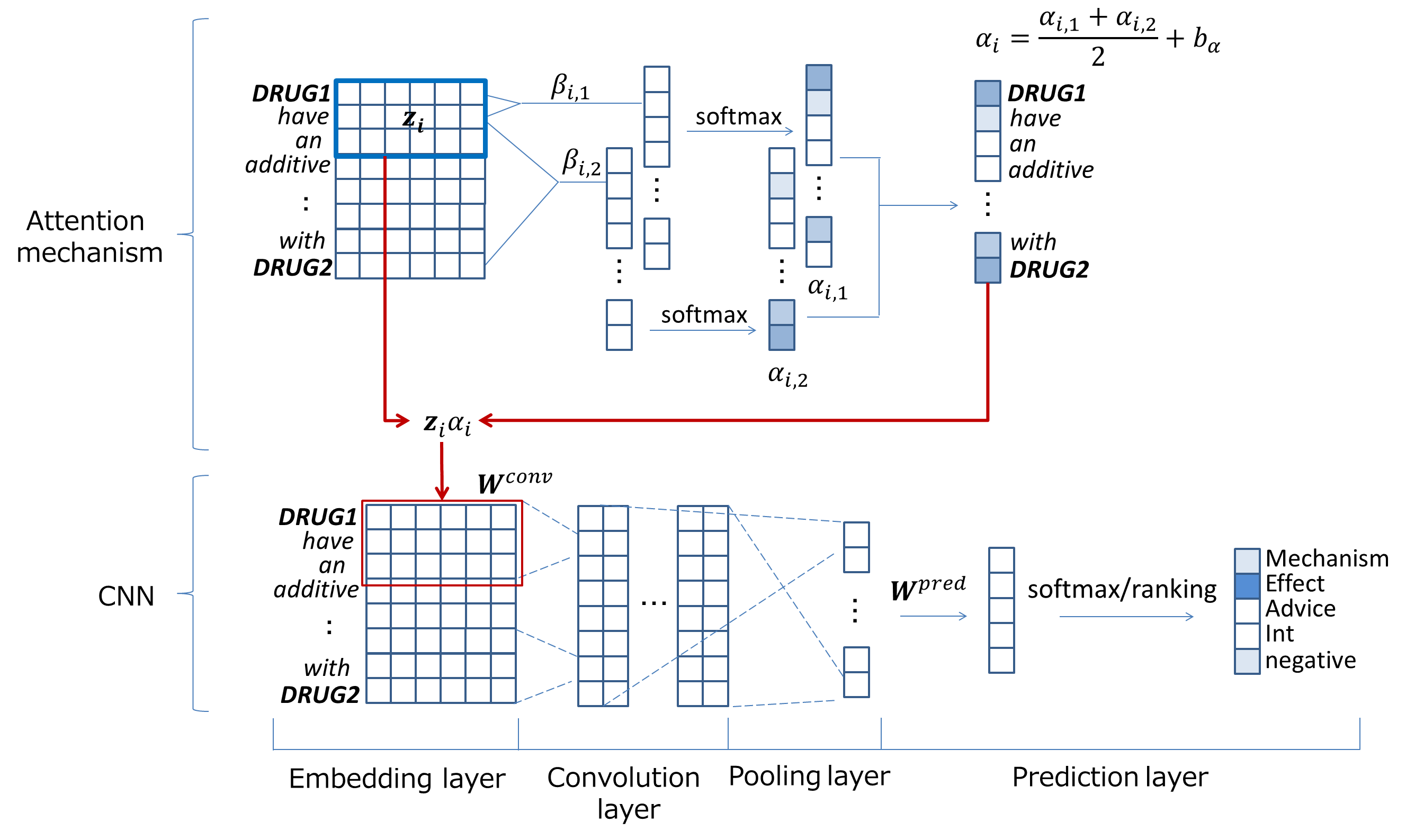}
\end{center}
\caption{The overview of CNN with attention mechanism}
\label{fig:sec3:attention_cnn}
\end{figure}

An attention mechanism for a CNN-based DDI extraction model is described. 
The overview of the DDI extraction model is illustrated in Figure~\ref{fig:sec3:attention_cnn}. 
The model extracts interactions from sentences with drugs are given.
This section first presents preprocessing of input sentences, then introduces the base CNN model and explains the attention mechanism. 
Finally, the training method is described.

\subsubsection{Base CNN Model}

The CNN model for extracting DDIs is based on Zeng et al.~\cite{zeng-etal-2014-relation}. In addition to the original objective function, the ranking-based objective function by dos-Santos et al.~\cite{dos-santos-etal-2015-classifying} is employed. The model consists of four layers: embedding, convolution, pooling, and prediction layers.
The CNN model is shown at the lower half of Figure~\ref{fig:sec3:attention_cnn}.

\subsubsection{Embedding Layer}

In the embedding layer, each word in the input sentence is mapped to a real-valued vector representation using an embedding matrix that is initialized with pre-trained embeddings.
Given an input sentence $S=(w_1, \cdots, w_n)$ with drug entities $e_1$ and $e_2$, each word $w_i$ is first converted into a real-valued vector $\bm{w}^w_i$ by an embedding matrix $\bm{W}^{emb} \in \mathbb{R}^{d_w{\times}|V|}$ as follows: 
\begin{eqnarray}
\bm{w}^w_i=\bm{W}^{emb}\bm{v}^w_i, 
\label{eq:9}
\end{eqnarray}
where $d_w$ is the number of dimensions of the word embeddings, 
$V$ is the vocabulary in the training data set and the pre-trained word embeddings, 
and $\bm{v}^w_i$ is a one-hot vector that represents the index of word embedding in $\bm{W}^{emb}$. 
$\bm{v}^w_i$ thus extracts the corresponding word embedding from $\bm{W}^{emb}$. 
The word embedding matrix $\bm{W}^{emb}$ is fine-tuned during training. 

$d_{wp}$-dimensional word position embeddings $\bm{w}^p_{i, 1}$ and $\bm {w}^p_{i,2}$ that correspond to the relative positions from first and second target entities are prepared, respectively. 
The word embedding $\bm {w}^w_i$ and these word position embeddings $\bm{w}^p_{i, 1}$ and $\bm {w}^p_{i,2}$ as in the following Equation~(\ref{eq:sec3:10}) are concatenated, and the resulting vector is used as the input to the subsequent convolution layer:
\begin{eqnarray}
\bm{w}_i=[\bm{w}^w_i; \bm{w}^p_{i,1}; \bm{w}^p_{i,2}].
\label{eq:sec3:10}
\end{eqnarray}

\subsubsection{Convolution Layer}

A weight tensor for convolution is defined as $\bm{W}_{k}^{conv}{\in}\mathbb{R}^{d_c\times (d_w+2d_{wp})\times k}$ and represents the $j$-th column of $\bm{W}_{k}^{conv}$ is defined as $\bm{W}^{conv}_{k,j}{\in}\mathbb{R}^{(d_w+2d_{wp})\times k}$. Here, $d_c$ denotes the number of filters for each window size, $k$ is a window size, and $K$ is a set of the window sizes of the filters. 
$\bm{z}_{i,k}$ that is concatenated $k$ word embeddings is also introduced: 
\begin{equation}
\bm{z}_{i,k}=[\bm{w}^{\mathrm{T}}_{\lfloor{i-(k-1)/2}\rfloor}; \ldots; \bm{w}^{\mathrm{T}}_{\lfloor{i-(k+1)/2}\rfloor}]^{\mathrm{T}}.
\label{eq:sec3:11}
\end{equation}
The convolution to the embedding matrix is applied as follows:
\begin{equation}\label{eq:sec3:m}
m_{i,j,k}=f(\bm{W}^{conv}_{k,j} \odot \bm{z}_{i,k}+b),
\end{equation}
where $\odot$ is an element-wise product, $b$ is the bias term, and $f$ is the ReLU function defined as: 
\begin{eqnarray}
  f(x)=
  \begin{cases}
    x, & \mbox{if}\ x > 0\\
    0, & \mbox{otherwise}.
  \end{cases}
\end{eqnarray}

\subsubsection{Pooling Layer}

The max pooling~\cite{maxpooling} is employed to convert the output of each filter in the convolution layer into a fixed-size vector as follows:
\begin{equation}
\bm{c}_k=[c_{1,k},\cdots ,c_{d_c,k}],\ c_{j,k}=\max_i m_{i,j,k}. 
\end{equation}
Then the $d_p$-dimensional output of this pooling layer is obtained, where $d_p$ equals to $d_c{\times}|K|$, by concatenating the obtained outputs $\bm{c}_k$ for all the window sizes $k_1, \cdots, k_K (\in K$):
\begin{eqnarray}
    \bm{c}=[\bm{c}_{k_1}; \ldots; \bm{c}_{k_i}; \ldots; \bm{c}_{k_K}].
\end{eqnarray}

\subsubsection{Prediction Layer}

The relation types are predicted using the output of the pooling layer.
First $\bm{c}$ is converted into scores using a weight matrix $\bm{W}^{pred}\in \mathbb{R}^{o{\times}d_p}$:
\begin{equation}\label{eq:sec3:score}
	\bm{s}=\bm{W}^{pred}\bm{c},
\end{equation}
where $o$ is the total number of relationships to be classified and $ \bm{s} = [s_1, \cdots, s_o]$. 
Then the following two different objective functions are employed for prediction.

\paragraph{Softmax}
$\bm{s}$ is converted into the probability of possible relations $\bm{p}$ by a softmax function:
\begin{equation}
	\bm{p}=[p_1,\cdots ,p_o],\ p_j=\frac{\exp{(s_j)}}{\sum^o_{l=1} \exp{(s_l)}}.
\end{equation}
The cross-entropy loss function $L_{CE}$ is defined as in the Equation~(\ref{eq:sec3:loss}) when the gold type distribution $\bm{y}$ is given. $\bm{y}$ is a one-hot vector where the probability of the gold label is 1 and the others are 0.
\begin{equation}\label{eq:sec3:loss}
	L_{CE}=-\sum\bm{y}\log{\bm{p}}
\end{equation}

\paragraph{Ranking}

The ranking-based objective function is employed  following \cite{dos-santos-etal-2015-classifying}.
Using the scores $\bm{s}$ in Equation~(\ref{eq:sec3:score}), the loss is calculated as follows:  
\begin{eqnarray}
    L_{ranking} = \log (1+\exp (\gamma (m^+-s_{y}))  \nonumber\\
    +\log (1+\exp (\gamma (m^-+s_{c})),
\end{eqnarray}
where $m^+$ and $m^-$ are margins, $\gamma$ is a scaling factor, $y$ is a gold label, and $c$ $(
\neq y)$ is a negative label with the highest score in $\bm{s}$.
$\gamma$ to 2, $m^+$ is set to 2.5 and $m^-$ is set to 0.5 following \cite{dos-santos-etal-2015-classifying}. 

\subsubsection{Attention Mechanism}

The attention mechanism is based on the input attention by Wang et al.~\cite{wang2016relation}\footnote{The attention-based pooling in \cite{wang2016relation} is not incorporated. This is left for future work.}. 
The attention mechanism is different from the base one in which separate attentions are prepared for entities and a bias term is incorporated to adjust the smoothness of attentions. 
The attention mechanism is illustrated at the upper half of Figure~\ref{fig:sec3:attention_cnn}.

The word indexes of the first and second target drug entities in the sentence are defined as $e_1$ and $e_2$, respectively.
The set of indices are denoted as $E = \{e_1, e_2\}$ and the indexes of the entities as $j \in \{1,2\}$. 
The attentions are calculated using:
\begin{eqnarray}
\beta_{i,j}&=&\bm{w}_{e_j}\cdot\bm{w}_i \\
\alpha_{i,j}&=&
 \begin{cases}
  \frac{\exp{(\beta_{i,j})}}{\sum_{1\leq l\leq n, l \notin E}\exp{(\beta_{l,j})}}, & 
  \mbox{if}\ i \notin E \\
  a_{drug}, & \mbox{otherwise}
 \end{cases}\\
\alpha_i&=&\frac{\alpha_{i,1}+\alpha_{i,2}}{2}+b_{\alpha}.
\end{eqnarray}
Here, $a_{drug}$ is an attention parameter for entities and $b_{\alpha}$ is the bias term. 
$a_{drug}$ and $b_{\alpha}$ are tuned during training. If $E$ is set to empty and $b_{\alpha}$ is set to zero, the attention will be the same as one by Wang et al.~\cite{wang2016relation}. 
The attentions $\alpha_i$ are incorporated into the CNN model by replacing Equation~(\ref{eq:sec3:m}) with the following equation: 
\begin{eqnarray}
m_{i,j,k}=f(\bm{W}^{conv}_j \odot \bm{z}_{i,k}\alpha_i+b).
\end{eqnarray}

\subsubsection{Training Method}

L2 regularization~\cite{hoerl1970ridge} is used to avoid over-fitting. The following objective functions $L'_{*}$ ($L'_{softmax}$ or $L'_{ranking}$) is used by incorporating the L2 regularization on weights to Equation (\ref{eq:loss}).
\begin{eqnarray}
L'_{*}=L_{*}+\lambda (||\bm{W}^{emb}||^2_F+||\bm{W}^{conv}||^2_F\\\nonumber
+||\bm{W}^{pred}||^2_F)
\label{eq:l2}
\end{eqnarray}
Here, $\lambda$ is a regularization parameter and $||\cdot||_F$ denotes the Frobenius norm. 
All the parameters including the weights $\bm{W}^{emb}$, $\bm{W}^{conv}$, and $\bm{W}^{pred}$, biases $b$ and $b_{\alpha}$, and the attention parameter $a_{drug}$ are updated to minimize $L'_{*}$. The adaptive moment estimation (Adam)~\cite{kingma2014adam} is used for the optimizer.
Training data set is randomly shuffled and divided into mini-batch samples in each epoch.

\subsection{Experimental Settings}

\subsubsection{DDI Data Sets}
The DDIExtraction-2011~\cite{segura20111st} workshop (First Challenge Task on Drug-Drug Interaction Extraction) focuses on the extraction of DDIs from biomedical texts and aims to promote the development of text mining and information extraction systems applied to the pharmacological domain in order to reduce time spent by the medical experts reviewing the literature for potential DDIs.
For performance comparisons between machine learning models, it is important to evaluate models on the data set created by trusted annotators with shared problem settings.
The construction of a common data set is essential for the development of research on automatic DDI extraction from the literature.
The main goal of this shared task is to have a benchmark for the comparison of advanced techniques, rather than competitive aspects.
The DDIExtraction-2013~\cite{segura-bedmar-etal-2013-semeval} follows up the DDIExtraction-2011 whose main goal was the detection of DDIs from texts. The DDIExtraction-2013 includes the classification of DDI types in addition to DDI detection.
Figure~\ref{fig:relate:semeval2013_brat} shows some examples of labeled sentences.
In the DDIExtraction-2013 task, it is necessary not only to detect DDIs but also to correctly classify DDIs into the type of the relations such as ``Effect'' and ``Mechanism''.
Additionally, while the data set used for the DDIExtraction-2011 task was composed by texts describing DDIs from the DrugBank~\cite{drugbank5}, the new data set for DDIExtraction-2013 also includes MEDLINE abstracts in order to deal with different types of texts and language styles.

The aim of the DDIExtraction-2013 task is to classify a given pair of drugs into the following four interaction types or no interaction: 
\begin{description}
\item [Mechanism]: A sentence describes pharmacokinetic mechanisms of a DDI, \\e.g., \textit{\textbf{Grepafloxacine} may inhibit the metabolism of \textbf{theobromine}.}
\item [Effect]: A sentence represents the effect of a DDI, e.g., \textit{\textbf{Methionine} may protect against the ototoxic effects of \textbf{gentamicin}.}
\item [Advice]: A sentence represents a recommendation or advice on the concomitant use of two drugs, e.g., \textit{\textbf{Alpha-blockers} should not be combined with \textbf{uroxatral}.}
\item [Int. (Interaction)]: A sentence simply represents the occurrence of a DDI without any information about the DDI, e.g., \textit{The interaction of \textbf{omeprazole} and \textbf{ketoconazole} has established.}
\end{description}

The DDIExtraction-2013 task relies on the DDI corpus, which is a semantically annotated corpus of documents describing DDIs from the DrugBank database and MEDLINE abstracts on the subject of DDIs.
The statistics of the DDIExtraction-2013 task data set are shown in Table~\ref{tab:sec2:stats_of_semeval2013}.
As shown in this table, the number of pairs that have no interaction (negative pairs) is larger than that of pairs that have interactions (positive pairs).

\begin{table*}[t!]
\centering
\begin{tabular}{lrrrr}\hline
 & \multicolumn{2}{c}{Train} & \multicolumn{2}{c}{Test}\\
 & DrugBank & MEDLINE & DrugBank & MEDLINE\\\hline
\#documents & 572 & 142 & 158 & 33 \\
\#sentences & 5,675 & 1,301 & 973 & 326 \\
\#pairs & 26,005 & 1,787 & 5,265 & 451 \\
\#positive DDIs & 3,789 & 232 & 884 & 95 \\
\#negative DDIs & 22,216 & 1,555 & 4,381 & 356 \\
\#Mechanism pairs & 1,257 & 62 & 278 & 24 \\
\#Effect pairs & 1,535 & 152 & 298 & 62 \\
\#Advice pairs & 818 & 8 & 214 & 7 \\
\#Int pairs & 179 & 10 & 94 & 2 \\\hline
\end{tabular}
\caption{Statistics for the DDIExtraction-2013 shared task data set}
\label{tab:sec2:stats_of_semeval2013}
\end{table*}

\subsubsection{Initializing Word Embeddings}

Skip-gram~\cite{mikolov2013distributed} was employed for the pre-training of word embeddings. 
The 2014 MEDLINE/PubMed baseline distribution is used, and the size of vocabulary was 1,630,978.
The embedding of the drugs, i.e., \textit{DRUG1}, \textit{DRUG2} and \textit{DRUGOTHER} are initialized with the pre-trained embedding of the word \textit{drug}.
The embeddings of training words that did not appear in the pre-trained embeddings, as well as the word position embeddings, are initialized with the random values drawn from a uniform distribution and normalized to unit vectors.
Words whose frequencies are one in the training data were replaced with an \textit{UNK} word during training, and the embedding of words in the test data set that did not appear in both training and pre-trained embeddings were set to the embedding of the \textit{UNK} word.

\subsubsection{Hyper-parameter Tuning}

The official training data set is split into two parts: training and development data sets. 
The hyper-parameters are tuned on the development data set on the softmax model without attentions. Table~\ref{tab:sec3:hyper_params} shows the best hyper-parameters on the softmax model without attentions. The same hyper-parameters are applied to the other models.
The statistics of the development data set is shown in Table~\ref{table:dev}. The sizes of the convolution windows is set to [3, 4, 5] that are the same as in Kim et al.~\cite{kim2014convolutional}.
The word position embedding size is chosen from \{10, 20, 30, 40, 50\}, the convolutional filter size from \{10, 50, 100, 200\}, the learning rate of Adam from \{0.01, 0.001, 0.0001\}, the mini-batch size from \{10, 20, 50, 100, 200\}, and the L2 regularization parameter $\lambda$ from \{0.01, 0.001, 0.0001, 0.00001\}.

\begin{table}[t!]
\centering
\begin{tabular}{lr}
\hline
Parameter & Value \\
\hline
Word embedding size & 200 \\
Word position embeddings size & 20 \\
Convolutional window size & [3, 4, 5] \\
Convolutional filter size & 100 \\
Initial learning rate & 0.001 \\
Mini-batch size & 100 \\
L2 regularization parameter & 0.0001 \\
\hline
\end{tabular}
\caption{Hyper-paramters}
\label{tab:sec3:hyper_params}
\end{table}

\begin{table}[t!]
\centering
\begin{tabular}{lr}
\hline
& Counts \\
\hline
Sentences & 1,404 \\
Pairs & 4,998 \\
\textit{Mechanism} pairs & 232 \\
\textit{Effect} pairs & 339 \\
\textit{Advice} pairs & 132 \\
\textit{Int} pairs & 48 \\\hline
\end{tabular}
\caption{Statistics of the development data set}
\label{table:dev}
\end{table}

\subsection{Results and Discussions}

This section first summarizes the performance of the  models and compares the performance with existing models. Then attention mechanisms are compared and finally some results for the analysis of the attentions are illustrated.

\begin{table*}[t!]
\centering
\begin{tabular}{lrrr} \hline
Type & $P$ (\%) & $R$ (\%) & $F$ (\%)\\\hline
 & \multicolumn{3}{c}{Softmax without attention}\\\hline
Mechanism  & 76.24 ($\pm$4.48) & 57.58 ($\pm$4.41) & 65.31 ($\pm$1.76)\\ 
Effect     & 67.84 ($\pm$3.56) & 63.61 ($\pm$4.95) & 65.39 ($\pm$1.38)\\ 
Advice     & 82.26 ($\pm$7.04) & 66.65 ($\pm$9.07) & 72.75 ($\pm$2.72)\\ 
Int        & {\bf 78.99} ($\pm$6.87) & 33.55 ($\pm$2.62) & {\bf 47.05} ($\pm$1.71)\\ \hline
All (micro) & 73.69 ($\pm$3.00) & 59.92 ($\pm$3.73) & 65.93 ($\pm$1.21)\\ \hline\\
 & \multicolumn{3}{c}{Softmax with attention}\\\hline
Mechanism  & 76.34 ($\pm$4.20) & {\bf 64.43} ($\pm$5.72) & 67.86 ($\pm$4.10)\\ 
Effect     & 66.84 ($\pm$3.12) & 65.98 ($\pm$2.63) & 65.58 ($\pm$2.09)\\ 
Advice     & 80.98 ($\pm$6.14) & 70.83 ($\pm$2.72) & 76.28 ($\pm$1.40)\\
Int        & 73.21 ($\pm$6.30) & {\bf 38.44} ($\pm$9.82) & 46.11 ($\pm$3.96)\\ \hline
All (micro)& 73.74 ($\pm$1.88) & 63.05 ($\pm$1.39) & 67.94 ($\pm$0.70) \\\hline\\
 & \multicolumn{3}{c}{Ranking without attention} \\\hline
Mechanism  & 78.41 ($\pm$3.99) & 58.17 ($\pm$5.10) & 66.51 ($\pm$2.61)\\
Effect     & 68.16 ($\pm$3.30) & 65.75 ($\pm$3.22) & 66.80 ($\pm$1.46)\\
Advice     & {\bf 84.49} ($\pm$3.55) & 67.14 ($\pm$4.68) & 74.61 ($\pm$1.82)\\
Int        & 73.95 ($\pm$7.09) & 33.43 ($\pm$1.18) & 45.91 ($\pm$1.23)\\\hline
All (micro) & 74.79 ($\pm$2.41) & 60.99 ($\pm$2.65) & 67.10 ($\pm$1.09)\\\hline\\
 & \multicolumn{3}{c}{Ranking with attention} \\\hline
Mechanism & {\bf 80.75} ($\pm$2.76) & 61.09 ($\pm$3.03) & {\bf 69.45} ($\pm$1.45)\\
Effect    & {\bf 69.73} ($\pm$2.64) & {\bf 66.63} ($\pm$2.93) & {\bf 68.05} ($\pm$1.29)\\
Advice    & 83.86 ($\pm$2.29) & {\bf 71.81} ($\pm$2.61) & {\bf 77.30} ($\pm$1.13)\\
Int       & 74.20 ($\pm$8.95) & 33.02 ($\pm$1.40) & 45.50 ($\pm$1.51)\\ \hline
All (micro)& {\bf 76.30} ($\pm$2.18) & {\bf 63.25} ($\pm$1.71) & {\bf 69.12} ($\pm$0.71)\\\hline
\end{tabular}
\caption[Performance of softmax/ranking CNN models with and without the attention mechanism]{Performance of softmax/ranking CNN models with and without the attention mechanism. The highest scores are shown in bold.}
\label{table:atthikaku}
\end{table*}

\begin{table*}[t!]
\centering
\begin{tabular}{llll} \hline
Methods & $P$ (\%) & $R$ (\%) & $F$ (\%)\\\hline
\multicolumn{4}{c}{No negative instance filtering}\\
\hline
CNN~\cite{liu2016drug} & 75.29 & 60.37 & 67.01\\
MCCNN~\cite{Quan2016} & - & - & 67.80\\
SCNN~\cite{zhao2016drug} & 68.5 & 61.0 & 64.5\\
Joint AB-LSTM~\cite{sahu2018drug} & 71.82 & 66.90 & 69.27\\
Attention model & 76.30 & 63.25 & 69.12\\
\hline
\multicolumn{4}{c}{With negative instance filtering}\\
\hline
FBK-irst~\cite{chowdhury2013fbk} & 64.6 & 65.6 & 65.1\\
Kim et al.~\cite{kim2015extracting} & - & - & 67.0\\
CNN~\cite{liu2016drug} &  75.72	& 64.66 & 69.75\\
MCCNN~\cite{Quan2016} & 75.99 & 65.25 & 70.21\\
SCNN~\cite{zhao2016drug} & 72.5 & 65.1 & 68.6 \\
Joint AB-LSTM~\cite{sahu2018drug} & 73.41 & 69.66 & 71.48 \\
\hline
\end{tabular}
\caption{Comparison with existing models}
\label{table:sonotahikaku}
\end{table*}

\begin{table*}[t!]
\centering
\begin{tabular}{lrrr} \hline
 & $P$ (\%) & $R$ (\%) & $F$ (\%)\\\hline
No attention & 74.79 ($\pm$2.41) & 60.99 ($\pm$2.65) & 67.10 ($\pm$1.09)\\
Input attention by Wang et al.~\cite{wang2016relation} &  73.48 ($\pm$1.96)& 59.58 ($\pm$1.51) & 65.77 ($\pm$0.80) \\
Attention & 76.30 ($\pm$2.66) & 63.25 ($\pm$2.59) & 69.12 ($\pm$0.71)\\
\hline
Attention w/o separate attentions $a_{drug}$ &  74.03 ($\pm$2.11)& 63.30 ($\pm$2.41) & 68.17 ($\pm$0.71) \\
Attention w/o the bias term $b_{\alpha}$ & 71.56 ($\pm$2.18)& 64.19 ($\pm$2.21) & 67.62 ($\pm$0.96) \\
\hline
\end{tabular}
\caption{Comparison of attention mechanisms on CNN models with ranking objective function}
\label{table:originhikaku}
\end{table*}

\subsubsection{Performance Analysis}

The performance of the base CNN models with two objective functions, as well as with or without the attention mechanism, is summarized in Table~\ref{table:atthikaku}. 
The incorporation of the attention mechanism improved the F-scores by about 2 percent points (pp) on models with both objective functions. Both improvements were statistically significant (p $<$ 0.01) with $t$-test.
This shows that the attention mechanism is effective for both models. 
The improvement of F-scores from the least performing model (softmax objective function without the attention mechanism) to the best performing model (ranking objective function
 with the attention mechanism) is 3.19 pp (69.12\% versus 65.93\%), and this shows both objective function and attention mechanism are key to improve the performance.
When looking into the individual types, ranking function with the attention mechanism archived the best F-scores on \textit{Mechanism}, \textit{Effect}, \textit{Advice}, while the base CNN model achieved the best F-score on \textit{Int}.

\subsubsection{Comparison with Existing Models}

A comparison with the existing state-of-the-art models is shown in Table~\ref{table:sonotahikaku}. 
The performance is mainly compared without negative instance filtering, which omits some apparent negative instance pairs with rules~\cite{chowdhury2013fbk}, since it is not incorporated.
The performance of the existing models with negative instance filtering is also shown for reference.

In the comparison without negative instance filtering, the model outperformed the existing CNN models~\cite{liu2016drug,Quan2016,zhao2016drug}. The model was competitive with Joint AB-LSTM model~\cite{sahu2018drug} that was composed of multiple RNN models.

\subsubsection{Comparison of Attention Mechanisms}

The attention mechanism is compared with the input attention of Wang et al.~\cite{wang2016relation} to show the effectiveness of the attention mechanism. Table~\ref{table:originhikaku} shows the comparison of the attention mechanisms. 
The base CNN-based model with ranking loss is also shown for reference, and the results of ablation tests.
As is shown in the table, the attention mechanism by Wang et al.~\cite{wang2016relation} did not work in DDI extraction. However, the attention improved the performance. This result shows that the extensions are crucial for modeling attentions in DDI extraction.
The ablation test results show that both extensions to the attention mechanism, i.e., separate attentions for entities and incorporation of the bias term, are effective for the task.

\subsubsection{Visual Analysis}

\begin{figure}[t]
\begin{center}
\includegraphics[width=1\linewidth]{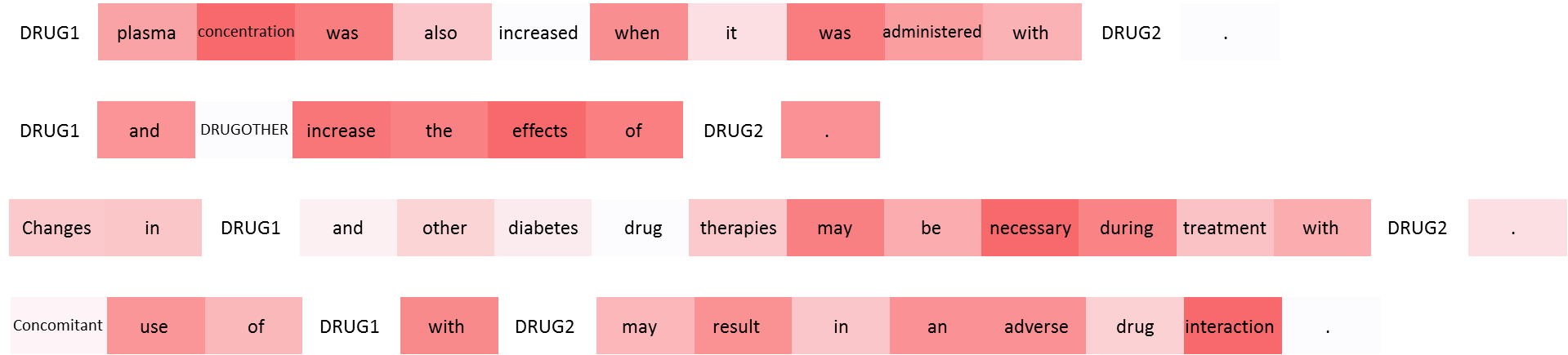}
\end{center}
\caption[The visualization of attention weights]{The visualization of attention weights. The dark part indicates that the attention value is large.}
\label{fig:sec3:attention_visualization}
\end{figure}

Figure~\ref{fig:sec3:attention_visualization} shows a visualization of attentions on some sentences with DDI pairs using the attention mechanism. In the first sentence, \textit{DRUG1} and \textit{DRUG2} have a \textsl{Mechanism} interaction. The attention mechanism successfully highlights the keyword \textit{concentration}. 
In the second sentence, which have an \textsl{Effect} interaction, the attention mechanism put high weights on \textit{increase} and \textsl{effects}. The word \textit{necessary} has a high weight on the third sentence with an \textsl{Advice} interaction.
For an \textsl{Int} interaction in the last sentence, the word \textit{interaction} is most highlighted.

\subsection{Summary}
In this chapter, an attention mechanism for relation extraction is provided as a preliminary.
Base CNN-based DDI extraction models is built with two different objective functions, softmax and ranking, and 
The attention mechanism is incorporated into the models.
The performance on the DDIExtraction-2013 data set is evaluated, and it is shown that both the attention mechanism and ranking-based objective function are effective for extracting DDIs.
The final model achieved an F-score of 69.12\%.
This model is a baseline, text-oriented relation extraction model and is not capable of domain information.

\section{Relation Extraction with a Single Kind of Domain Information}\label{sec:cnn_with_mol}
This chapter reviews a study on dealing with a single kind of domain information, which is the molecular structure of drugs as the preparatory step in the case study of using heterogeneous information on drugs.
A method that combines the input sentence information and molecular structure information is reviewed based on \cite{asada-etal-2018-enhancing}.
The summary of this chapter is three-fold: 
\begin{itemize}
\item{A neural method to extract DDIs from texts with the related molecular structure information is described.} 
\item{GCNs is applied to pairwise drug molecules and it is shown that GCNs can predict DDIs between drug molecular structures with high accuracy.} 
\item{It is shown that the molecular information is useful in extracting DDIs from texts.}
\end{itemize}

\subsection{Background}
In parallel to the progress in DDI extraction from texts, Graph Convolutional Networks (GCNs) have been proposed and applied to estimate physical and chemical properties of molecular graphs such as solubility and toxicity~\cite{NIPS2015_5954,li2015gated,gilmer2017neural}. 

It is a very challenging attempt to consider different kinds of items such as text information and molecular structure information at the same time as described in Chapters 5-7.
However, both the input sentence vector encoded by CNNs and the molecular structure vector encoded by GCNs are embedded in a low-dimensional real-valued vector space, so both information is effectively utilized for the DDI extraction task.

\subsection{Methods}
\subsubsection{Text-based DDI Extraction}

The proposed model for extracting DDIs from texts is based on the CNN model by Zeng et al.~\cite{zeng-etal-2014-relation}.
When an input sentence $S=(w_1, w_2, \cdots , w_N)$ is given, 
word embedding $\bm{w}^w_i$ of $w_i$ and word position embeddings $\bm{w}^p_{i, 1}$ and $\bm {w}^p_{i,2}$ that correspond to the relative positions from the first and second target entities are prepared, respectively. 
These embeddings are concatenated as in Equation~(\ref{eq:10}), and the resulting vector are used as the input to the subsequent convolution layer:
\begin{eqnarray}
\bm{w}_i=[\bm{w}^w_i; \bm{w}^p_{i,1}; \bm{w}^p_{i,2}],
\label{eq:10}
\end{eqnarray}
where $[;]$ denotes the concatenation.
The expression for each filter $j$ with the window size $k_l$ is calculated as:
\begin{eqnarray}
\bm{z}_{i,l}&=&[\bm{w}_{i-(k_l-1)/2},\cdots,\bm{w}_{i-(k_l+1)/2}],\\
m_{i,j,l}&=&\mathrm{relu}(\bm{W}^{conv}_j \odot \bm{z}_{i,l}+b^{conv}), \\
m_{j,l} &=&\max_i m_{i,j,l},
\end{eqnarray}
where $L$ is the number of windows, $\bm{W}^{conv}_j$ and $b^{conv}$ are the weight and bias of CNN, and $\max$ indicates max pooling~\cite{maxpooling}.

The output of the convolution layer is converted into a fixed-size vector that represents a textual pair as follows:
\begin{eqnarray}
\bm{m}_l&=&[m_{1,l}, \cdots, m_{J,l}],\\
\bm{h}_{t}&=&[\bm{m}_{1}; \ldots; \bm{m}_{L}],
\end{eqnarray}
where $J$ is the number of filters. 

Prediction $\hat{\bm{y}_t}$ is obtained by the following fully connected neural networks: 
\begin{eqnarray}
  \bm{h}_t^{(1)}&=&\mathrm{relu}(\bm{W}_{t}^{(1)} \bm{h}_{t}+\bm{b}_{t}^{(1)}), \label{eq:y}\\
  \hat{\bm{y}_t}&=&\mathrm{softmax}(\bm{W}_{t}^{(2)} \bm{h}^{(1)}_{t}+\bm{b}_{t}^{(2)}),
\end{eqnarray}
where $\bm{W}_t^{(1)}$ and $\bm{W}_t^{(2)}$ are weights and $\bm{b}_t^{(1)}$ and $\bm{b}_t^{(2)}$ are bias terms.

\subsubsection{Molecular Structure-based DDI Classification}

Drug pairs are represented in molecular graph structures using two GCN methods: CNNs for fingerprints (NFP)~\cite{NIPS2015_5954} and Gated Graph Neural Networks (GGNN)~\cite{li2015gated}. They both convert a drug molecule graph $G$ into a fixed size vector $\bm{h}_g$ by aggregating the representation $\bm{h}^T_v$ of an atom node $v$ in $G$. 
Atoms are represented as nodes and bonds are represented as edges in the graph.

\textbf{NFP} first obtains the representation $\bm{h}^t_v$ by the following equations~\cite{NIPS2015_5954}.
\begin{eqnarray}
  \bm{m}^{t+1}_v&=&\bm{h}^{t}_{v}+\sum_{w \in{N(v)}}\bm{h}^t_w,\\
  \bm{h}^{t+1}_v&=&\sigma(\bm{H}_t^{deg(v)}\bm{m}_v^{t+1}),
\end{eqnarray}
where $\bm{h}^t_v$ is the representation of $v$ in the $t$-th step, $N(v)$ is the neighbors of $v$, 
and $\bm{H}_t^{deg(v)}$ is a weight parameter.
$\bm{h}^0_v$ is initialized by the \textit{atom features} of $v$. 
$deg(v)$ is the degree of a node $v$ and $\sigma$ is a sigmoid function.
NFP then acquires the representation of the graph structure 
\begin{equation}
  \bm{h}_g=\sum_{v,t}\mathrm{softmax}(\bm{W}^t \bm{h}^t_v),
  \label{eq:gnfp}
\end{equation}
where $\bm{W}^t$ is a weight matrix.

\textbf{GGNN} first obtains the representation $\bm{h}^t_v$ by using Gated Recurrent Unit (GRU)-based recurrent neural networks~\cite{li2015gated} as follows:
\begin{eqnarray}
  \bm{m}^{t+1}_v&=&\sum_{w \in{N(v)}}\bm{A}_{e_{vw}}\bm{h}^t_w\\
  \bm{h}^{t+1}_v&=&\mathrm{GRU}([\bm{h}^t_v; \bm{m}_v^{t+1}]),
\end{eqnarray}
where $\bm{A}_{e_{vw}}$ is a weight for the \textit{bond type} of each edge $e_{vw}$.
GGNN then acquires the representation of the graph structure.
\begin{equation}
  \bm{h}_g=\sum_{v}\sigma(i([\bm{h}^T_v; \bm{h}^0_v]))\odot(j (\bm{h}^T_v)),
  \label{eq:gggnn}
\end{equation}
where $i$ and $j$ are linear layers and $\odot$ is the element-wise product.

The representation of a molecular pair is obtained by concatenating the molecular graph representations of drugs $g_1$ and $g_2$, i.e., $\bm{h}_{m}=[\bm{h}_{g_1}; \bm{h}_{g_2}]$. 

Prediction $\hat{\bm{y}}_m$ is obtained as follows:
\begin{eqnarray}
\bm{h}^{(1)}_{m} &=& \mathrm{relu}(\bm{W}^{(1)}_{m} \bm{h}_{m}+\bm{b}^{(1)}_{m}),\\
\hat{\bm{y}}_{m}&=&\mathrm{softmax}(\bm{W}^{(2)}_{m}\bm{h}^{(1)}_{m}+\bm{b}^{(2)}_{m}),
\end{eqnarray}
where $\bm{W}^{(1)}_{m}$ and $\bm{W}^{(2)}_{m}$ are weights and $\bm{b}^{(1)}_{m}$ and $\bm{b}^{(2)}_{m}$ are bias terms.

\subsubsection{DDI Extraction from Texts Using Molecular Structures}

\begin{figure}[t]
\begin{center}
\includegraphics[width=.95\linewidth]{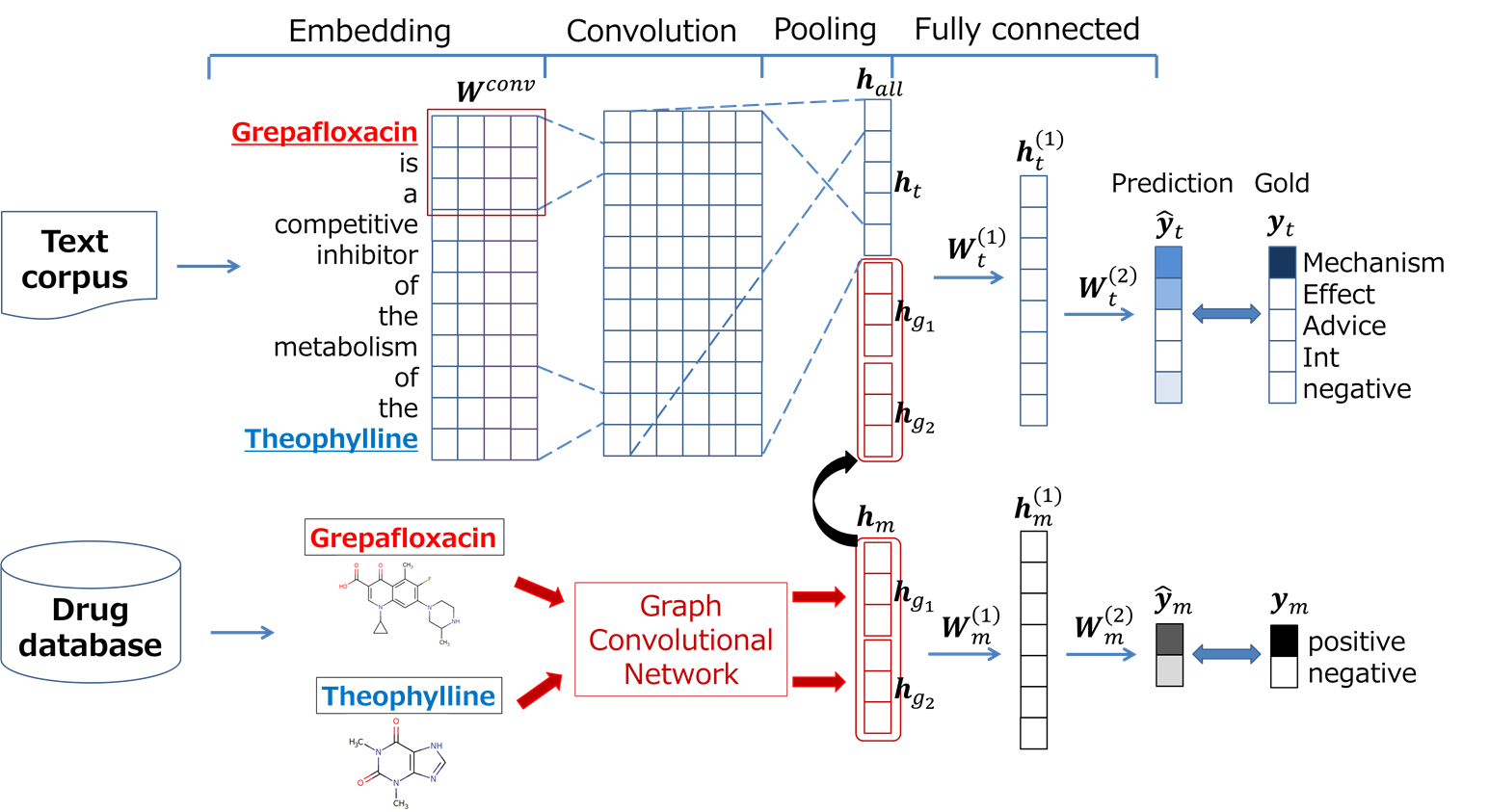}
\end{center}
\caption{Overview of the CNN-based DDI extraction model that use molecular structure information}
\label{fig:sec3:mol_cnn}
\end{figure}

The simultaneous use of textual and molecular information is achieved by concatenating the text-based and molecule-based vectors: 
$\bm{h}_{all}=[\bm{h}_{t}; \bm{h}_m]$. 
Molecule-based vectors are normalized. 
Then $\bm{h}_{all}$ is used instead of $\bm{h}_{t}$ in Equation~\ref{eq:y}. 

In training, the molecular-based DDI classification model is trained.
The molecular-based classification is conducted by minimizing the loss function 
$L_{m} = -\sum{\bm{y}_{m} \log\hat{\bm{y}}_{m}}$.
The parameters for GCNs are fixed and the text-based DDI extraction model is trained by minimizing the loss function.
$L_{t} = -\sum{\bm{y}_{t} \log\hat{\bm{y}}_{t}}$.

\subsection{Experimental Settings}

As preprocessing, sentences are split into words using the GENIA tagger~\cite{geniatagger}. 
The drug mentions of the target pair are replaced with \textsl{DRUG1} and \textsl{DRUG2} according to their order of appearance.   
Other drug mentions are also replaced with \textsl{DRUGOTHER}.
Negative instance filtering is not employed unlike other existing methods, e.g., Liu et al.~\cite{liu2016drug}, since the focus is to evaluate the effect of the molecular information on texts.

Mentions in texts are linked to DrugBank entries by string matching. The mentions and the names are lowercased in the entries and chose the entries with the most overlaps. 
As a result, 92.15\% and 93.09\% of drug mentions in train and test data set matched the DrugBank entries.

\subsubsection{Data and Task for Molecular Structures}

255,229 interacting (positive) pairs are extracted from DrugBank. 
Note that, unlike text-based interactions, DrugBank only contains the information about interacting pairs; 
there are no detailed labels and no information for non-interacting (negative) pairs.
Thus the same number of pseudo-negative pairs are generated by randomly pairing drugs and removing those in positive pairs. 
To avoid overestimation of the performance, drug pairs mentioned in the test set of the text corpus are deleted.
Positive and negative pairs are split into 4:1 for training and test data, and the molecular GCN-based model is evaluated on the classification accuracy. 

To obtain the graph of a drug molecule, the SMILES~\cite{weininger1988smiles} string encoding of the molecule is taken as input from DrugBank and then converted into the graph using RDKit~\cite{landrum2016rdkit}.
For the \textit{atom features}, randomly embedded vectors are used for each atom (i.e., C, O, N, ...). 4 \textit{bond types}, single, double, triple, and aromatic, are used.

\subsubsection{Training Settings}

Mini-batch training is employed using the Adam optimizer~\cite{kingma2014adam}. L2 regularization is used to avoid over-fitting. The bias term $\bm{b}^{(2)}_t$ is tuned for negative examples in the final softmax layer. 
Pre-trained word embeddings trained by using the word2vec tool~\cite{mikolov2013distributed} on the 2014 MEDLINE/PubMed baseline distribution are employed. The vocabulary size was 215,840.
The embedding of the drugs, i.e., \textsl{DRUG1} and \textsl{DRUG2} were initialized with the pre-trained embedding of the word \textsl{drug}.
The embeddings of training words that did not appear in the pre-trained embeddings were initialized with the average of all pre-trained word embeddings. Words that appeared only once in the training data were replaced with an \textsl{UNK} word during training, and the embedding of words in the test data set that did not appear in both training and pre-trained embeddings were set to the embedding of the \textsl{UNK} word. Word position embeddings are initialized with random values drawn from a uniform distribution.
The molecule-based vectors of unmatched entities are set to zero vectors. 

\subsection{Results and Discussions}

\begin{table}[t!]
\centering
\begin{tabular}{p{3.5cm}lll} \hline
Methods & P & R & F (\%)\\\hline
  Liu et al.~\cite{liu2016drug} & 75.29 & 60.37 & 67.01\\
  Zheng et al.~\cite{Zheng2017} & 75.9 & 68.7 & 71.5 \\
  Lim et al.~\cite{lim2018drug} & 74.4 & 69.3 & 71.7\\\hline
  Text-only  & 71.97 & 68.44 & 70.16\\
  + NFP & 72.62 & 71.81 & 72.21\\
  + GGNN & 73.31 & 71.81 & 72.55\\
  \hline
  \end{tabular}
  \caption{Evaluation on DDI extraction from texts}
  \label{tab:sec4:comparison}
\end{table}

\begin{table}[t!]
\centering
\begin{tabular}{lllll} \hline
DDI Type & \textsl{Mech.} & \textsl{Effect} & \textsl{Adv.} & \textsl{Int} (\%)\\\hline
  Text-only & 69.52 & 69.27 & 79.81 & 48.18 \\
  + NFP & 72.70 & 72.44 & 79.56 & 46.98 \\
  + GGNN & 73.83 & 71.03 & 81.62 & 45.83 \\
  \hline
  \end{tabular}
  \caption{F-scores of each DDI type}
  \label{tab:sec4:type}
\end{table}

\begin{table}[t!]
  \centering
  \begin{tabular}{lr} \hline
  Methods & Accuracy (\%) \\\hline
  NFP &  94.19\\
  GGNN & 98.00 \\\hline
  \end{tabular}
  \caption{Accuracy of binary classification on DrugBank pairs}
   \label{table:bin}
\end{table}
\begin{table}[t!]
\centering
\begin{tabular}{lllll} \hline
Methods & P & R  & F & Acc. (\%)\\\hline
  NFP & 15.56 & 48.93 & 23.61 & 45.78\\
  GGNN & 15.11 & 57.10 & 23.90 & 37.72\\
  \hline
  \end{tabular}
  \caption{Classification of DDIs in texts by molecular structure-based DDI classification model}
  \label{table:detection}
\end{table}

Table~\ref{tab:sec4:comparison} shows the performance of DDI extraction models. The performance without negative instance filtering or ensemble is shown for a fair comparison. The increase in recall and F-score are observed by using molecular information, which results in the state-of-the-art performance with GGNN.

Both GCNs improvements were statistically significant ($p<0.05$ for NFP and $p<0.005$ for GGNN) with randomized shuffled test~\cite{fisher1937design}. 

Table~\ref{tab:sec4:type} shows F-scores on individual DDI types. The molecular information improves F-scores especially on type \textsl{Mechanism} and \textsl{Effect}.

The accuracy of binary classification on DrugBank pairs is also evaluated by using only the molecular information in Table~\ref{table:bin}. 
The performance is high, although the accuracy is evaluated on automatically generated negative instances.

Finally, the molecular-based DDI classification model trained on DrugBank is applied to the DDIExtraction 2013 task data set. 
Since the DrugBank has no detailed labels, all four types of interactions are mapped to positive interactions and evaluated the classification performance. The results in Table~\ref{table:detection} show that GCNs produce higher recall than precision and the overall performance is low considering the high performance on DrugBank pairs. This might be because the interactions of drugs are not always mentioned in texts even if the drugs can interact with each other and because hedged DDI mentions are annotated as DDIs in the text data set. 
The DDI extraction model is trained only with molecular information by replacing $\bm{h}_{all}$ with $\bm{h}_m$, but the F-scores were quite low ($<$ 5\%). These results show that textual relations cannot be predicted only with molecular information.

\subsection{Summary}
This chapter reviewed a neural method for relation extraction using both textual information and molecular structures.
The model was evaluated on the DDI extraction task as a case study. The results show that DDIs can be predicted with using molecular structure information. 
Since this model cannot deal with heterogeneous domain information, the following chapters develop relation extraction models that can consider heterogeneous domain information.

\section{Relation Extraction with Multiple Domain Information} \label{sec:bert_mol_desc}
This chapter includes work from the published paper Asada et al. (2021a)~\cite{asada-bioinformatics} and Iinuma et al. (2021)~\cite{iinuma-etal-2021-tticoin}.
This chapter proposes a novel relation extraction method that utilizes two kinds of domain information. 
The DDI extraction task is chosen as a case study.
A method to utilize the description and the structure of the entity obtained from  drug database DrugBank as well as large-scale raw text information is proposed.
DrugBank is in focus because the DDIExtraction 2013 shared task data set is created based on the DrugBank database. 
Other databases are left for future work. 
Specifically, the description and molecular structure information of drugs in the database are utilized. 
The information from large-scale raw texts is incorporated by using a Bidirectional Encoder Representations from Transformers (BERT) model~\cite{devlin-etal-2019-bert} pre-trained on large-scale raw text. 

Experimental results show that SciBERT boosts the performance of the baseline model. As a result, the performance is already strong enough and better than the previously reported performance. 
It is shown that the drug database information is complementary to the large-scale pre-trained information, and the simultaneous use of drug description and drug molecular structure information can enhance the performance of DDI extraction from texts with SciBERT.  

\subsection{Background}

Since the annotation efforts are costly and time-consuming, it is unrealistic to prepare a sufficient amount of annotated data. 
In addition, it is difficult to learn how to extract DDIs from text only with the limited amount of annotated text because a deep understanding of DDI interaction descriptions often requires domain knowledge of drugs. 
Various drug information, such as detailed descriptions and molecular structure information on drugs, are registered in drug databases. 
Furthermore, models pre-trained on large-scale raw text show significant improvements in various natural language processing (NLP) tasks~\cite{devlin-etal-2019-bert}.
Effective use of such external information is necessary to reduce the reliance on annotated text.

This chapter extends the previous section at the following points.
\begin{itemize}
    \item The token representation changes from word2vec to contextualized vectors obtained by SciBERT. As a result, the performance of the baseline with the state-of-the-art performance is remarkably improved.
    \item The neural molecular GNN~\cite{tsubaki2018compound} that considers relatively large fragments of atoms and better represents molecular structures is employed.
    \item Drug descriptions registered in the drug database is used and it is shown that drug description information is useful for extracting DDIs from the corpus for some DDI types.
    \item It is found that the large-scale pre-training information, drug description, and drug molecular information are complementary and their effective combination can largely improve the DDI extraction performance.
\end{itemize}

\begin{figure*}[t]
  \centering
  \includegraphics[width=.95\linewidth]{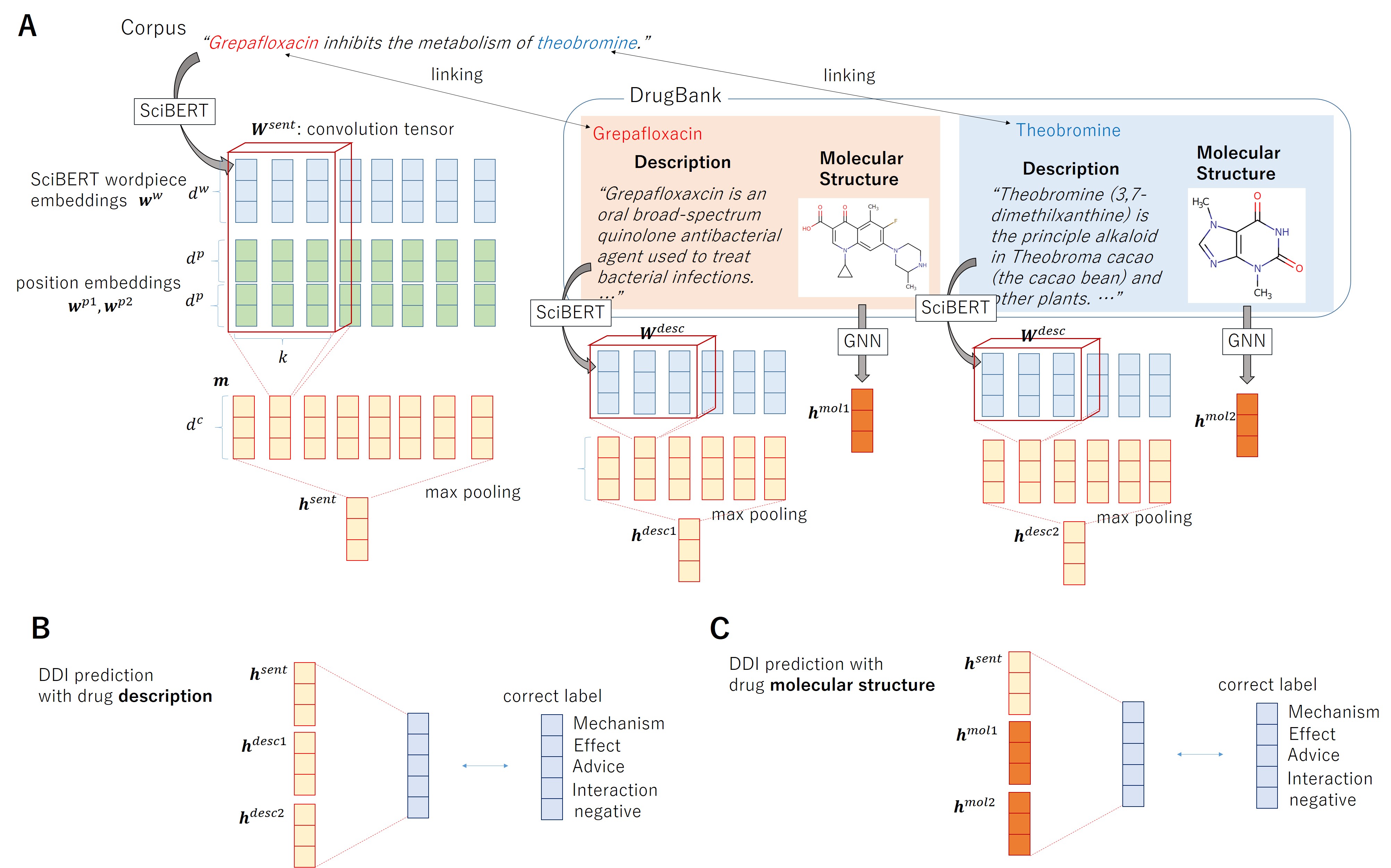}
  \caption[Overview of the DDI extraction model with drug descriptions and drug molecular structures]{Overview of the DDI extraction model with drug descriptions and drug molecular structures. (A) Illustrates how to encode input sentences, drug descriptions and drug molecular structures.
  (B) and (C) show the prediction layer when the drug description representation and the drug molecular structure representation are used.}
  \label{figure:overview}
\end{figure*} 

\subsection{Methods}

The overview of the proposed method is illustrated in Figure~\ref{figure:overview}.
For the baseline model, the convolutional neural network (CNN)-based DDI extraction model~\cite{asada-etal-2018-enhancing} that receives an input sentence with a target drug pair and classifies the pair into a specific DDI type is employed. 
The input sentence is enriched using SciBERT~\cite{beltagy2019scibert}, which is a BERT model trained on large-scale biomedical and computer science text.
The drug description representation of the target drugs is obtained using SciBERT and the molecular structure representation of the target drugs using molecular graph neural network (GNN) model proposed by Tsubaki et al.~\cite{tsubaki2018compound}.
These drug description and molecular structure representation are combined with the enriched input sentence representation and classify the target drug pair into a specific DDI type.

\subsubsection{Input Sentence Representation}

This section follows the previous section to preprocess the input sentences.
Then a preprocessed input sentence is converted into a real-valued fixed size vector by BERT and CNN-based model~\cite{devlin-etal-2019-bert,zeng-etal-2014-relation} and shows the model in the left part of Figure~\ref{figure:overview}A. 
Given an input sentence $S=(w_1, \cdots , w_n)$ with drug mentions $m_1$ and $m_2$, the sentence is first split into wordpieces (a.k.a., subwords) by the WordPiece algorithm~\cite{kudo-richardson-2018-sentencepiece}.
Each wordpiece $w_i$ is converted into a real-valued pre-trained contextualized embedding $\bm{w}^w_i\in\mathbb{R}^{d^w}$ by the BERT model (light blue vectors in Figure~\ref{figure:overview}A).
$d^p$-dimensional position embeddings $\bm{w}^{p1}_i$ and $\bm{w}^{p2}_i$ for each wordpiece, which correspond to the relative positions from the first and second target mentions are prepared, respectively. (green vectors in Figure~\ref{figure:overview}A)
The wordpiece embedding $\bm{w}^w_i$ and the position embeddings $\bm{w}^{p1}_i$ and $\bm{w}^{p2}_i$ are concatenated as in the following Equation (\ref{eq:cnn_input}):
\begin{equation}
  \bm{w}_i = [\bm{w}^w_i;\bm{w}^{p1}_i;\bm{w}^{p2}_i],
  \label{eq:cnn_input}
\end{equation}
where [;] denotes concatenation. the resulting embeddings are used to prepare the input to the convolution layer.

$\bm{z}_{i}$ that is the concatenation of $k$ input embeddings\footnote{Multiple windows can be employed instead of a single window with the size $k$, but there is no significant difference in the performance in the preliminary experiment.} around $w_i$ is introduced:
\begin{equation}
  \bm{z}_{i}=[\bm{w}^{\mathrm{T}}_{\lfloor{i-(k-1)/2}\rfloor}; \ldots; \bm{w}^{\mathrm{T}}_{\lfloor{i-(k+1)/2}\rfloor}]^{\mathrm{T}}.
  \label{eq:11}
\end{equation}
Convolution to the embeddings is applied as follows:
\begin{equation}\label{eq:m}
m_{i,j}=f(\bm{W}^{sent}_{j} \odot \bm{z}_{i}+b^{sent}),
\end{equation}
where $\odot$ is an element-wise product, $b^{sent}$ is a bias term, and $f(\cdot)$ is a GELU~\cite{hendrycks2016gaussian} function.\footnote{
The GELU activation function is chosen from ReLU, eLU, SeLU and GELU based on the results in the preliminary experiment.} A weight tensor for convolution is defined as $W^{sent}\in\mathbb{R}^{d^c\times (d^w+2d^p) \times k}$. The $j$-th column of $W^{sent}$ is represented as $W^{sent}_j$. $k$ is a window size. 
The tensor $W^{sent}$ is depicted as a red box in the left part of Figure~\ref{figure:overview}A.
Then, max-pooling to convert the output of each filter in the convolution layer into a fixed-size vector is employed as follows:
\begin{equation}
\bm{h}^{sent}=\max_i m_{i,j}. 
\end{equation}

\subsubsection{Drug Description Representation} 

Similarly to the input sentences, the description sentences of a drug mention are converted to the real-valued fixed size vector by BERT and CNN.
The wordpiece embeddings by BERT are directly used without word position embeddings to prepare the input to the convolution layer. A convolution weight tensor $W^{desc}\in\mathbb{R}^{d^c\times (d^w) \times k}$ and bias $b^{desc}$ for description are defined. 
Convolution and max-pooling are employed in the same way as the processing of the input sentences and the description representations $\bm{h}^{desc1}$ and $\bm{h}^{desc2}$ of drug mentions $m_1$ and $m_2$ are obtained respectively.

\subsubsection{Molecular Structure Representation}
\begin{figure}[t]
  \centering
  \includegraphics[width=.85\linewidth]{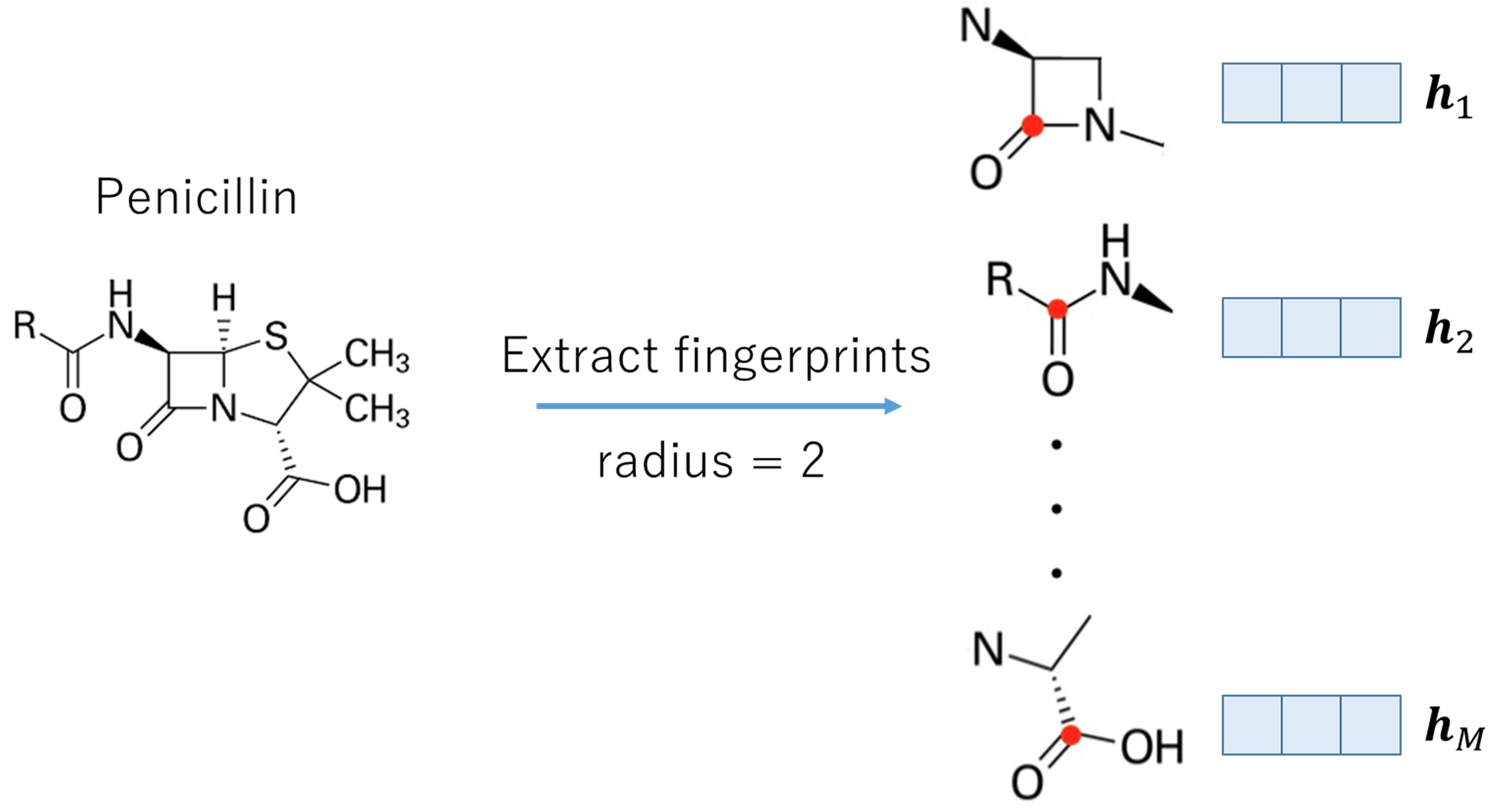}
  \caption{Illustration of molecular fingerprints. This figure shows the extraction of several fingerprint subgraphs from a molecular structure when radius is 2.}
  \label{figure:fingerprint_gnn}
\end{figure} 

The molecular graph structures of drugs are represented using GNNs.
GNNs convert a drug molecule graph $G$ into a fixed size vector $h^{g}$.
Atoms as represented as nodes and bonds as edges in the graph.
The neural molecular GNN method proposed by Tsubaki et al.~\cite{tsubaki2018compound} is employed.
The molecular GNN method uses relatively large fragments referred to as $r$-radius subgraphs or molecular fingerprints to represent atoms with their contexts in the graph.
The molecular GNN adopts fingerprint vectors as atom vectors, initializes the vectors randomly, and updates them considering the graph structure of a molecule.
The vector of the $i$-th atom in a drug molecule is defined as $\bm{h}_i$ and the set of its neighboring atoms as $N_i$.
The vector $\bm{h}_i$ is updated in the $\ell$-th step as follows:
\begin{equation}
    \bm{h}^{\ell}_{i}=\bm{h}^{\ell-1}_{i}+\sum_{j\in N_i} f(\bm{W}^{\ell-1}_{hidden}\bm{h}^{\ell-1}_{j}+\bm{b}_{hidden}^{l-1}),
\end{equation}
where $f(\cdot)$ denotes a ReLU function.
The drug molecular vector is obtained by summing up all the atom vectors and then the resulting vectors are fed into a linear layer.
\begin{equation}
    \bm{h}^{mol}=f(\bm{W}_{output}\sum_{i}^M \bm{h}^L_{i}+\bm{b}_{output}),
\end{equation}
where $M$ is the number of fingerprints. Figure~\ref{figure:fingerprint_gnn} shows how the molecular GNN model extracts fingerprints including $\beta$-lactam ($\bm{h}_1$) from penicillin drug ($r$=2) and update fingerprint vectors.

The molecular structure representations $\bm{h}^{mol1}$ and $\bm{h}^{mol2}$ of drug mentions $m_1$ and $m_2$ are obtained, respectively.

\subsubsection{DDI Extraction Using Database Information}

When the drug description information for DDI extraction is used, the input sentence representation and two description representations as in Equation~\ref{eq:all_h_desc}:
\begin{equation}
	\bm{h}=[\bm{h}^{sent};\bm{h}^{desc1};\bm{h}^{desc2}].
	\label{eq:all_h_desc}
\end{equation}

Similarly, two molecular structure representations are concatenated with the input sentence representation as in Equation~\ref{eq:all_h_mol}:
\begin{equation}
	\bm{h}=[\bm{h}^{sent};\bm{h}^{mol1};\bm{h}^{mol1}].
	\label{eq:all_h_mol}
\end{equation}

The resulting vector is used as the input to the prediction layer. 
$\bm{h}$ into prediction scores is converted using a weight matrix $\bm{W}^{pred}\in \mathbb{R}^{o{\times}d_p}$:
\begin{equation}\label{eq:score}
	\bm{s}=\bm{W}^{pred}\bm{h},
\end{equation}
where $\bm{s} = [s_1, \cdots, s_o]$ and $o$ is the number of DDI types. 
 $\bm{s}$ is converted into the probability of possible interactions $\bm{p}$ by a softmax function:
\begin{equation}\label{eq:pred}
	\bm{p}=[p_1,\cdots ,p_o],\ p_j=\frac{\exp{(s_j)}}{\sum^o_{l=1} \exp{(s_l)}}.
\end{equation}
The DDI extraction using drug description information and drug molecular structure information in Figure~\ref{figure:overview}B and C are illustrated, respectively.

\subsubsection{Training} 

The loss function $L$ is defined as in the Equation~(\ref{eq:loss}) using $\bm{p}$ in Equation~(\ref{eq:pred}) when the gold type distribution $\bm{y}$ is given. $\bm{y}$ is a one-hot vector where the probability of the gold label is 1 and the other probabilities are 0.
\begin{equation}\label{eq:loss}
	L=-\sum\bm{y}\log{\bm{p}}
\end{equation}

\subsubsection{Ensemble}

An ensemble technique is employed to combine the prediction from different models. 
Specifically, the prediction scores are simply summed up from different models for the ensemble after each of the models is trained separately.
For instance, when the prediction of the model with the description information and that with the molecular structure information are combined,  the prediction scores are summed up in  Equation~(\ref{eq:score}) as follows:
\begin{equation}
	\bm{s}=\bm{s}^{desc}+\bm{s}^{mol}.
\end{equation}

\subsection{Experimental Settings}

This section explains the DDI extraction task settings, drug database preprocessing, drug mention linking, and hyper-parameter settings.

\subsubsection{DrugBank Preprocessing}
DrugBank is a freely available drug database  containing more than 10,000 drugs.
Each drug is given sentences describing its characteristics and efficacy. The first sentence of the drug description of \textit{Salbutamo} is shown as an example: \textit{Salbutamol is a short-acting, selective beta2-adrenergic receptor agonist used in the treatment of asthma and COPD.}
DrugBank also contains drug molecular structure information.
Structure information is registered in SMILES string encoding.

To obtain the graph of a drug molecule, the SMILES string encoding of the molecule is obtained as input from DrugBank and then converted it into the graph structure using RDKit~\cite{Landrum2016RDKit2016_09_4}. 
Fingerprints are extracted from the graph using preprocessing scripts provided by Tsubaki et al.~\cite{tsubaki2018compound}.

\subsubsection{Drug Mention Linking}

\begin{figure}[t]
  \centering
  \includegraphics[width=.9\linewidth]{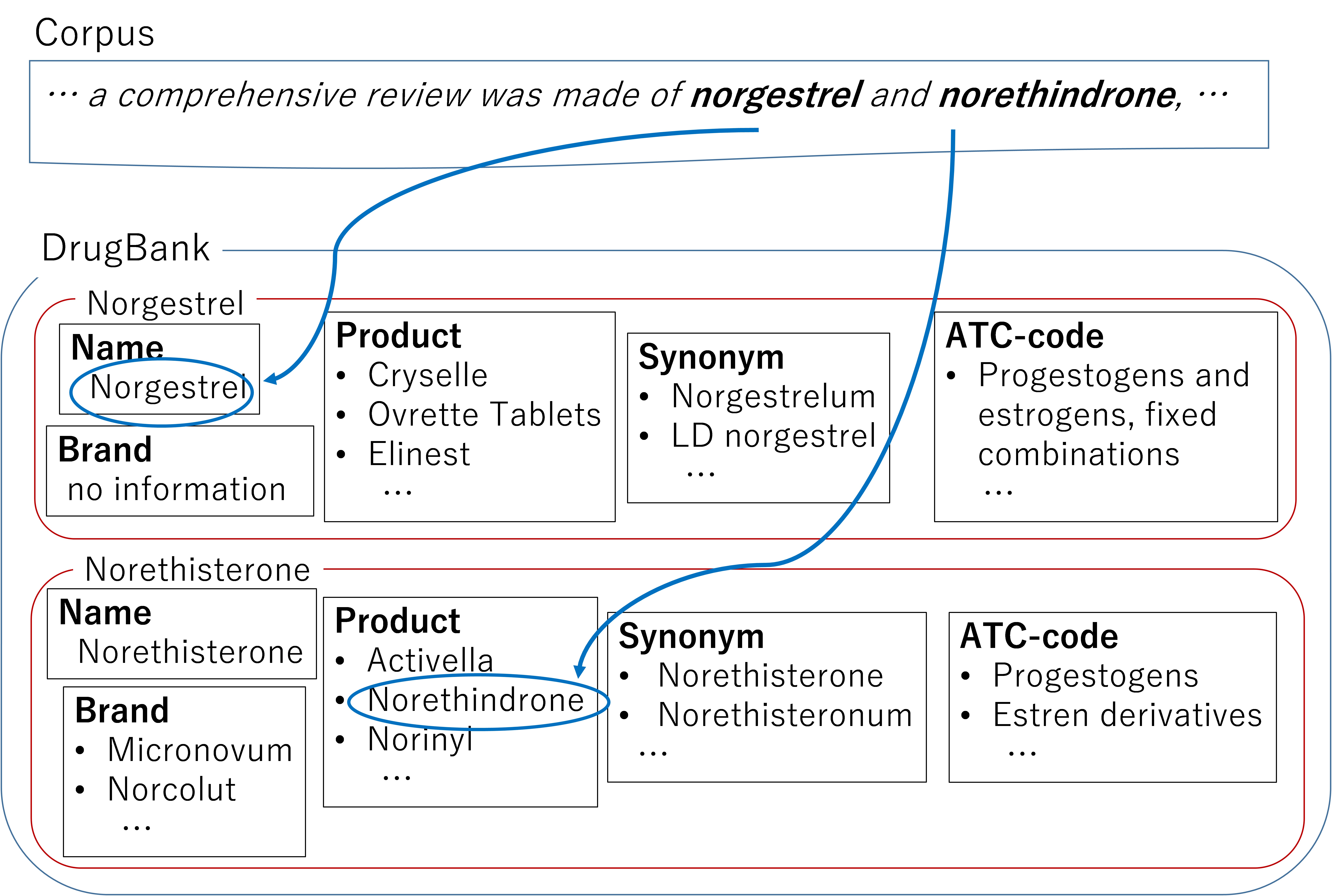}
  \caption{Linking between mentions and DrugBank entry}
  \label{figure:linking}
\end{figure}

Mentions in the corpus are linked to DrugBank entries by relaxed string matching. 
In particular, each mention and the following items in the DrugBank entries are lowercased, and the entry that includes an item showing the most overlap with the mention is chosen. 
\begin{itemize}
    \item Name: Headword of the drug entry
    \item Brand: Brand names from different manufactures
    \item Product: The final commercial preparation of the drug
    \item Synonym: Synonyms of the drug
    \item ATC codes: Codes for hierarchical drug classification
\end{itemize}
For the ATC code, the same code can be assigned to multiple drugs, so only the ATC codes that are assigned to single drugs is used for mention linking.
Also, for synonyms, mentions and synonyms are linked by exact string matching instead of relaxed string matching to avoid the matching with very short strings (e.g., abbreviations).
With this linking, 90.50\% and 91.10\% of drug mentions in DDIExtraction-2013 train and test data set matched the DrugBank entries.
Figure~\ref{figure:linking} shows how the linking is performed.
The input sentence contains two mentions ``norgestrel'' and ``norethindrone''.  String matching is conducted to link these mentions to DrugBank entries. As a result, the mention ``noregestrel'' matched the Name item and the mention ``norethindrone'' matched the Product item.

\subsubsection{Training Settings}
This section follows the training settings for the fine-tuning of BERT on the GLUE tasks~\cite{devlin-etal-2019-bert} except for the following two points.
First,the AdamW optimizer~\cite{loshchilov2018decoupled} is employed instead of Adam optimizer. Second, mixed-precision training~\cite{le2018mixed} is employed for the memory efficiency. 

Dropout~\cite{dropout2014srivastava} is employed to the input of the convolution layer for regularization.
Word position embeddings are initialized with random values drawn from a uniform distribution between $-10^{-3}$ and $10^{-3}$.
The description and molecular structure vectors of unmatched entities are set to zero vectors. 
Table~\ref{table:param} and \ref{table:GNN_params} shows hyper-parameters for CNNs and GNNs. The same hyper-parameters are used as the GLUE tasks in Devlin et al.~\cite{devlin-etal-2019-bert} for the BERT layer. 
In the DDIExtraction 2013 shared task, the official development data set is not provided; thus a development data set is prepared from the official training data set to choose the other hyper-parameters. 
In order to train the model on the same setting as other existing models~\cite{liu2016drug, asada-etal-2018-enhancing, peng-etal-2019-transfer}, the development data set is included in the entire training data set for training the model.
The entire training data set for training the model is used to evaluate the performance on the test set. 
For GNNs, the results with different radii 0, 1, and 2 for molecular fingerprints is shown. 
Note that GNNs with a radius of 0 mean no molecular fingerprints, which assign vectors to atoms.
\begin{table}[t]
\centering
  \begin{tabular}{lr}\hline
  Parameter & Value \\\hline
  Word embedding size $d^w$ & 768 \\
  Initial learning rate & 5e-5 \\
  Number of fine-tuning epochs & 3 \\
  L2 weight decay & 0.01 \\
  Dropout rate & 0.1 \\
  Mini-batch size & 32 \\
  Word position embedding size $d^p$ & 10 \\ 
  Convolution window size $k$ & 5 \\
  Convolution filter size $d^c$ & 768 \\
  Convolution window size for description & 3\\
  Convolution filter size for description & 20\\
  \hline
  \end{tabular}
  \caption{Hyper-parameters for CNNs}
  \label{table:param}
\end{table}

\begin{table}[t]
\centering
  \begin{tabular}{lr}\hline
  Parameter & Value \\\hline
  Molecular embedding size $d^g$ & 50 \\
  Number of hidden layer $L$ & 5 \\
  Radius & 1\\\hline
  \end{tabular}
  \caption{Hyper-parameters for GNNs}
  \label{table:GNN_params}
\end{table}

\subsection{Results and Discussions}
Table~\ref{table:comparison} shows the performance of DDI extraction models including the proposed models with different settings and the state-of-the-art models.
It can be seen that the baseline text-only model (SciBERT CNN) using SciBERT is powerful. SciBERT improved the performance of the model without SciBERT (word2vec CNN) by 11.04\% points in the micro F-score. 
With this improvement, the model with SciBERT has achieved the state-of-the-art performance when it is compared with the state-of-the-art models in the top rows of the table.
When the CNNs are ommited  from the baseline model (SciBERT Linear), the first special token [CLS] is used as the aggregated representation of the sentence and fed the embedding of [CLS] into the linear classifier layer. The performance slightly dropped with this omission but the difference is negligible. This indicates the BERT model is powerful enough to capture the similar information as CNNs.

Additional increase of the micro F-score is observed by using drug description and molecular structure information as shown in the bottom part of the table. This shows the large-scale raw text information from SciBERT and the database information are complementary, and they are both useful for extracting DDIs from text. 
For GNNs, GNNs with molecular fingerprints (radius=1 or 2) show better performance than GNNs without them (radius=0), and GNNs with the radius of 1 show the highest performance. 
When comparing the description and molecular structure information, the micro F-score with molecular structure information (radius=1) is slightly higher than one with the description information (+Desc), but their difference is not significant and the superiority depends on how to represent the molecular structure information, i.e., molecular fingerprints. The author leaves the search of the better representations for future work.
The improvement by the ensemble model of description and the molecular structure information is statistically significant when compared with the baseline model ($p<0.005$, McNemar test).
The scikit-learn~\cite{scikit-learn} Python library is used for evaluating the statistical significance.

Table~\ref{table:comparison_devel} shows the performance of DDI extraction models on the development data set. 
Consistently with the results on the test set in Table~\ref{table:comparison}, either of the description information and molecular structure information improves the performance and the combination of the two kinds of information showed the highest F-scores on the development data set. However, there are some inconsistencies in the results on development and test data sets; the model with molecular structure information showed a higher F-score than the model with description information on the development data set, while the model with molecular structure information showed a lower F-score on the test data set.

Table~\ref{table:type} shows the F-scores on individual DDI types.
The description information improves F-scores for \textsl{Mechanism}, \textsl{Effect}, and \textsl{Int.} types, but it degrades the F-scores for \textsl{Advice}.
The molecular structure information improves F-scores for \textsl{Effect} and \textsl{Advice}, but it degrades the F-scores for \textsl{Mechanism} and \textsl{Int.} for some radii.
This indicates the two information have different effects on extracting DDIs, and each kind of information is not enough to improve the entire DDI extraction performance.
When both the description and molecular structure information are used by the ensemble technique, the model shows higher performance than the baseline model on all types.
The training data set is cross-validated using 5-fold cross-validation and further analyzed the performance on individual DDI types.
Table~\ref{table:type_anal} shows the F-scores for folds of cross-validated training data set.
The micro-averaged F-score is used to calculate the average of the folds.
The models with individual information source show higher performance than the baseline model on \textsl{Mechanism} and \textsl{Int.}, while they show comparable or lower performance than the baseline model on other labels. 
Although the changes in performance are inconsistent for the DDI types and folds, the model with the ensemble technique shows higher performance than the models with individual information source on average. 
As a result, the model with the ensemble technique improves the F-scores on average for all the types except for \textsl{Int.}, where the model performs on par with the baseline model. These results show that the performance on each label is affected by data splitting, but overall, when both the description information and molecular structure information are used by the ensemble technique, the model is effective for improving the performance of DDI extraction.

Table~\ref{table:comparison_dataset} shows the comparison of F-scores on the two different subsets of the test set: MEDLINE and DrugBank.
The model with the description and one with molecular structure (radius=1) degrade the F-score for MEDLINE, whereas both the description and molecular structure information improved the F-scores for DrugBank. For both subsets, the ensemble model greatly improved the F-score.
These results also indicate the description and molecular structure information are complementary.

\begin{table}[t]
  \centering
  \begin{tabular}{llll} \hline
  Method & P & R & F (\%)\\\hline
  Liu et al.~\cite{liu2016drug} & 75.29 & 60.37 & 67.01\\
  BioBERT~\cite{peng-etal-2019-transfer} & - & - & 78.8\\\hline
  Text-only (word2vec CNN) \\\cite{asada-etal-2018-enhancing} & 71.97 & 68.44 & 70.16\\
  Text-only (SciBERT Linear) & 80.28& 81.92 & 81.09\\
  Text-only (SciBERT CNN) & 83.10 & 80.38 & 81.72\\
  + Desc & 84.05 & 81.81 & 82.91\\
  + Mol (radius=0) & 83.29 & 82.02 & 82.65\\
  + Mol (radius=1) & 83.57 & 82.12 & 82.84\\
  + Mol (radius=2) & 83.66 & 81.10 & 82.36\\
  + Desc + Mol (radius=1)  & \textbf{85.36} & \textbf{82.83} & \textbf{84.08}\\
  + Desc + Mol (radius=0,1,2)  & 84.51 & 82.53 & 83.51\\
  + Mol (radius=0,1,2)  & 84.69 & 82.53 & 83.60\\ \hline
  \end{tabular}
  \caption[Evaluation on DDI extraction from texts on the test set]{Evaluation on DDI extraction from texts on the test set. Text only (SciBERT CNN) model is defined as the baseline model.}
  \label{table:comparison}
\end{table}

\begin{table}[t]
  \centering
  \begin{tabular}{llll} \hline
  Method & P & R & F (\%)\\\hline
  Text-only (SciBERT CNN) & 83.55 & 80.19 & 81.84\\
  + Desc & 83.19 & 82.31 & 82.75\\
  + Mol (radius=0) & 83.73 & 81.25 & 82.47\\
  + Mol (radius=1) & 82.85 & 83.90 & 83.37\\
  + Mol (radius=2) & 82.88 & 83.58 & 83.23\\
  + Desc + Mol (radius=1) & \textbf{84.59} & \textbf{84.32} & \textbf{84.46}\\
  \hline
  \end{tabular}
  \caption{Evaluation on DDI extraction from texts on the development set}
  \label{table:comparison_devel}
\end{table}

\begin{table}[t]
  \centering
  \begin{tabular}{lllll} \hline
  & \multicolumn{4}{c}{DDI Type} \\
  Method & \textsl{Mech.} & \textsl{Effect} & \textsl{Adv.} & \textsl{Int.} (\%)\\\hline
  Text-only & 86.18 & 79.12 & 88.34 & 55.94 \\
  + Desc & \textbf{87.62} & 81.08 & \underline{87.05} & \textbf{60.27}\\
  + Mol (radius=0) & \underline{84.65} & 81.20 & 90.67 & \underline{55.71} \\
  + Mol (radius=1) & 86.33 & 80.48 & \textbf{92.07} & \underline{49.25} \\
  + Mol (radius=2) & \underline{84.02} & \textbf{82.24} & 88.58 & 57.34 \\
  + Desc + Mol (radius=1) & 87.61 & 82.05 & 90.79 & 58.74 \\\hline
  \end{tabular}
  \caption[Performance on individual DDI types in F-scores]{Performance on individual DDI types in F-scores. The best score for each type is shown in bold and the scores lower than the baseline model are shown with underlines.}
  \label{table:type}
\end{table}

\begin{table}[t]
  \centering
  \begin{tabular}{llllll} \hline
  & & \multicolumn{4}{c}{DDI Type} \\
  & Method & \textsl{Mech.} & \textsl{Effect} & \textsl{Adv.} & \textsl{Int.} (\%)\\\hline
Fold 1 & Text-only & 84.60 & \textbf{86.38} & 85.80 & 68.29 \\
& + Desc & \underline{82.55} & \underline{81.82} & \underline{85.23} & \underline{64.37} \\
& + Mol (radius=1) & \underline{84.55} & \underline{84.62} & \underline{84.53} & \textbf{71.05} \\
& + Desc + Mol (radius=1) & \textbf{86.13} & \underline{85.46} & \textbf{86.69} & \underline{67.47} \\\hline
Fold 2 & Text-only & 83.46 & 83.26 & 78.80 & \textbf{81.48} \\
& + Desc & 84.15 & \underline{82.52} & 81.99 & \underline{79.01} \\
& + Mol (radius=1) & \underline{82.26} & \textbf{83.45} & 81.64 & \underline{76.54} \\
& + Desc + Mol (radius=1) & \textbf{84.29} & 83.38 & \textbf{82.64} & \underline{79.01} \\\hline
Fold 3 & Text-only & 84.91 & 59.21 & \textbf{76.54} & 91.43 \\
& + Desc & \underline{83.40} & \textbf{88.31} & \underline{73.53} & 91.67 \\
& + Mol (radius=1) & \underline{84.43} & 86.24 & \underline{75.24} & \textbf{94.44} \\
& + Desc + Mol (radius=1) & \textbf{86.09} & 87.25 & \underline{76.22} & 92.96 \\\hline
Fold 4 & Text-only & 76.81 & 81.56 & 78.01 & 79.45 \\
& + Desc & 77.54 & 82.47 & \textbf{79.65} & 81.16 \\
& + Mol (radius=1) & \textbf{78.17} & 84.03 & \underline{77.34} & \underline{76.92} \\
& + Desc + Mol (radius=1) & 77.35 & \textbf{85.15} & 79.40 & \textbf{83.33} \\\hline
Fold 5 & Text-only & 81.97 & 81.76 & \textbf{89.51} & 76.54 \\
& + Desc & 84.95 & 83.02 & \underline{87.73} & \textbf{81.48} \\
& + Mol (radius=1) & 86.09 & 83.74 & \underline{87.23} & \underline{73.33} \\
& + Desc + Mol (radius=1) & \textbf{86.26} & \textbf{84.91} & \underline{88.34} & \underline{75.00} \\\hline
Average & Text-only & 82.34 & 76.99 & 81.67 & \textbf{79.07} \\
& + Desc & 83.09 & 84.39 & \underline{81.27} & \underline{78.09} \\
& + Mol (radius=1) & 82.47 & 83.57 & \underline{81.60} & \underline{78.97} \\
& + Desc + Mol (radius=1) & \textbf{84.01} & \textbf{85.20} & \textbf{82.70} & \underline{78.99} \\\hline

  \end{tabular}
  \caption[Individual F-scores on 5-fold cross-validated training data set]{Individual F-scores on 5-fold cross-validated training data set. The micro-averaged F-score is used to calculate the average of the folds. The best score for each type is shown in bold and the scores lower than the baseline model are shown with underlines.}
  \label{table:type_anal}
\end{table}

\begin{table}[t]
  \centering
  \begin{tabular}{llll} \hline
  Method & MEDLINE & DrugBank & Overall (\%)\\\hline
  Text-only (SciBERT CNN) & 74.57 & 82.44 & 81.72\\
  + Desc & 74.41 & 83.75 & 82.91\\
  + Mol (radius=0) & 75.00 & 83.41 & 82.65\\
  + Mol (radius=1) & 73.98 & 83.71 & 82.84\\
  + Mol (radius=2) & 74.57 & 83.15 & 82.36\\
  + Desc + Mol (radius=1) & \textbf{78.16} & \textbf{84.67} & \textbf{84.08}\\\hline
  \end{tabular}
  \caption{Comparisons of F-scores on different parts of the test set}
  \label{table:comparison_dataset}
\end{table}

\subsubsection{Pre-training of GNNs and CNNs on DrugBank}

To investigate the further use of DrugBank information, it is verified if the DrugBank DDI labels can improve the DDI extraction performance.
Specifically, GNNs are pretrained  for molecular structure information and CNNs for description information on DrugBank DDI labels.
Many drug pairs have information of interactions, so this pre-training needs no additional annotations.

50,000 interacting (positive) pairs are extracted from DrugBank. Note that, unlike the DDIExtraction 2013 shared task data set, DrugBank only contains the information of interacting pairs; there are no detailed labels and no information for non-interacting (negative) pairs. 
Thus, the same number of pseudo negative pairs are generated by randomly pairing drugs and removing those in positive pairs.
To avoid overestimation of the performance, drug pairs mentioned in the test set of the text corpus are deleted in preparing the pairs. Positive and negative pairs are split into 4:1 for train and test data, and evaluated the classification accuracy using only the molecular information or only the description.

First, the performances of the accuracy of binary classification on DrugBank DDI pairs are shown in Table~\ref{table:acc}.
The performance is surprisingly high, although the accuracy is evaluated on automatically generated negative instances.
Overall, both drug description and molecular structure information can capture DDI information in DrugBank.
In detail, the accuracy with drug description information is higher than that with molecular structure information. 
For molecular structure information, GNN with the radius of 2 shows the best performance.
The difference in accuracy between radius 0 and 2 is 21.78\% points, and this large difference shows the importance of capturing molecular fingerprints for DDI.

\begin{table}[t]
  \centering
  \begin{tabular}{llr} \hline
   &  & Accuracy (\%) \\\hline
  Description & SciBERT & 91.05 \\
  Molecular Structure & GNN (radius=0) & 67.58\\
   & GNN (radius=1) & 82.21 \\
   & GNN (radius=2) & 89.36 \\
  \hline
  \end{tabular}
  \caption{Accuracy of binary classification on the DrugBank pairs}
  \label{table:acc}
\end{table}

CNNs and GNNs are pre-trained using the DrugBank interaction labels including the pseudo negative labels and fine-tuned on the DDIExtraction 2013 data set.
Table~\ref{table:pretraining} shows the comparison of the F-scores with or without pre-training.
Unfortunately, for all the settings, the models with pre-training show lower performance than those without pre-training.
This may be because the labels in the DDI extraction tasks are annotated depending on the context of the pairs and the labels can be inconsistent with labels in DrugBank and because the pseudo negative examples are used in training instead of the real negative examples.

\begin{table}[t]
  \centering
  \begin{tabular}{llllll} \hline
  & Methods & P & R & F (\%)\\\hline
  & SciBERT & 83.10 & 80.38 & 81.72\\\hline
   w/ pre-training& + Desc & \textbf{84.62} & 79.26 &  81.85\\
  & + Mol (radius=0) & 82.69 & 81.00 & 81.83\\
  & + Mol (radius=1) & 84.51 & 80.28 & 82.34\\
  & + Mol (radius=2) & 82.36 & 80.28 & 81.74\\\hline
  w/o pre-training & + Desc & 84.05 & 81.81 & \textbf{82.91}\\
  & + Mol (radius=0) & 83.29 & 82.02 & 82.65\\
  & + Mol (radius=1) & 83.57 & \textbf{82.12} & 82.84\\
  & + Mol (radius=2) & 83.66 & 81.10 & 82.36\\
  \hline
  \end{tabular}
  \caption{Evaluation on DDI extraction from texts with or without pre-training of GNNs for the molecular structure and CNNs for the description}
  \label{table:pretraining}
\end{table}

\subsubsection{Can DrugBank Information Alone Extract DDIs from Texts?}

To further investigate how contextual information is important in the DDI task,
it is verified whether the textual DDI can be extracted only from the drug information in DrugBank without using the input sentence.
The input sentence representation $\bm{h}^{sent}$ is simply omitted from Equation~\ref{eq:all_h_desc} and \ref{eq:all_h_mol} and the DDI extraction models are trained, but the F-scores were quite low (<5\%) for both models.
This result shows that DDI relations cannot be extracted from texts only with the description and molecular structure information.
This indicates that DDI extraction from text greatly depends on the context information around drug mention pairs and the models on the database information serve as a supplement to the textual CNN model.

\begin{figure}[t]
  \centering
  \includegraphics[width=.9\linewidth]{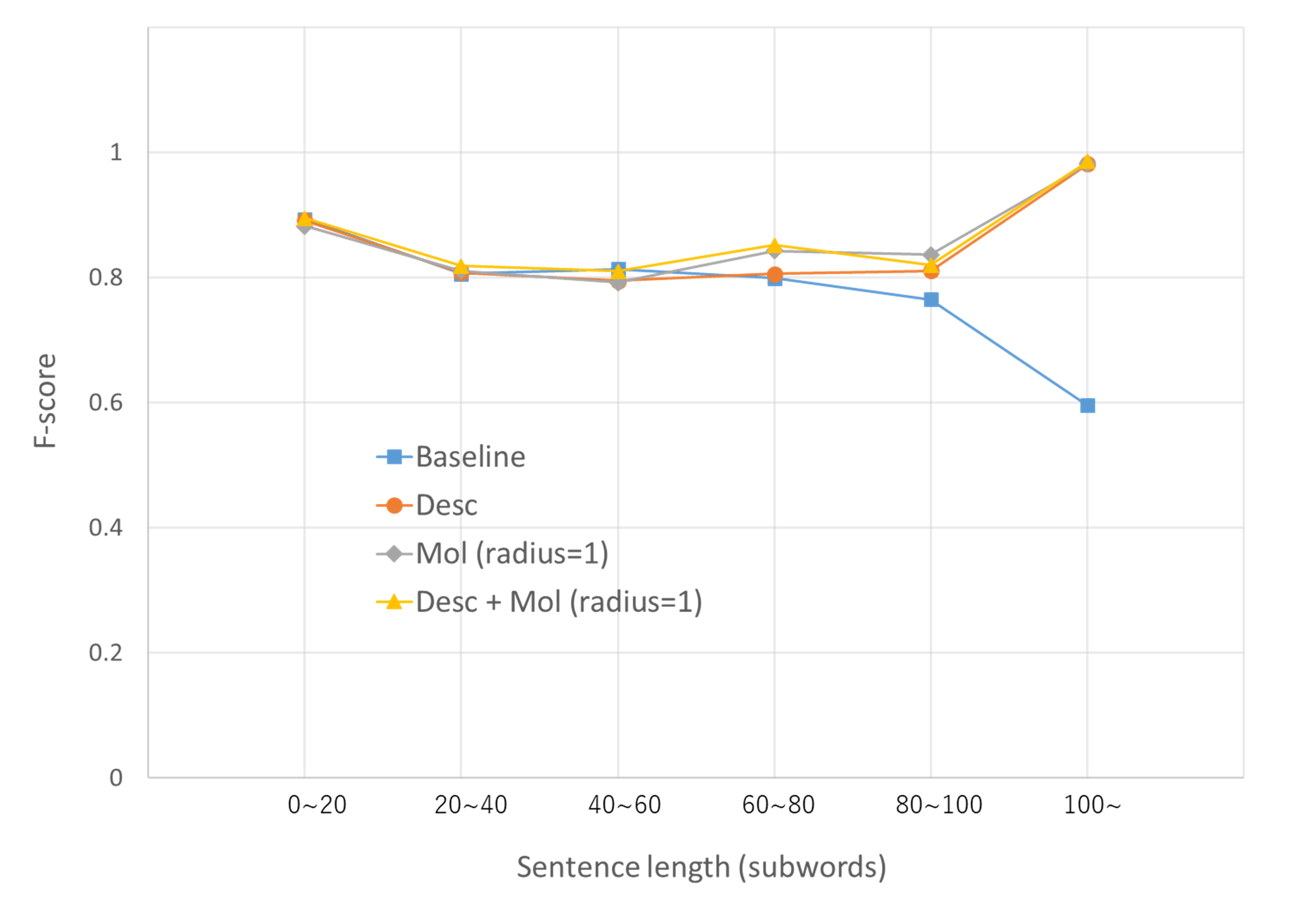}
  \caption[F-scores for different sentence lengths on the 5-fold cross-validated training data set]{F-scores for different sentence lengths on the 5-fold cross-validated training data set. The micro-averaged F-score is used to calculate the average of the folds.}
  \label{figure:sent_len}
\end{figure} 

\subsubsection{Error Analysis}
Figure~\ref{figure:sent_len} shows F-scores for different sentence lengths on the validation data set. 
Since the instances with longer sentence lengths are relatively few, 5-fold cross-validation is used on the official training data set.
Here, the sentence length is defined to be the number of subwords divided by the SciBERT vocabulary.
In the previous work, Quan et al.~\cite{Quan2016} analyzed the F-scores for the sentence length and pointed out that the performance is low for very long sentences with 60 or more words. Wang et al.~\cite{Wang2017} also analyzed the F-scores for the sentence length and showed that F-scores tend to drop when the lengths of the instances are in the range from 71 to 100. The baseline model shows lower performance for long sentences with 80 or more subwords, and this result shows the same tendency as the previous analyses.
The model shows higher performance than the baseline model, especially for sentences with more than 100 subwords. 
This shows that the DrugBank information is helpful to predict DDIs when the input sentences are long and complex and it is difficult to consider the whole contexts.

\subsection{BioCreativeVII Track-1 DrugProt}
\subsubsection{Introduction}

The DrugProt task of the BioCreative VII Track 1 is tackled with neural models that employ external knowledge. The models are based on BERT, which shows the state-of-the-art performance on several NLP tasks and can be considered as external knowledge from other texts. 
In addition, distant supervision data~\cite{iinuma-etal-2022-improving} is utilized and information of structure of drugs and proteins is utilized as external knowledge from knowledge bases.

\subsubsection{Task Definition}
The DrugProt data set is a corpus that is exhaustively annotated by domain experts, and all drug and protein mentions in PubMed articles are labeled. 
In addition, for all possible drug-protein pairs, binary relationships corresponding to the 13 types of drug-protein interactions are annotated. In other words, when some binary relations are true, the drug-protein pair has multiple interactions and when all binary relations are false, it indicates that there is no interaction between the pair.
The goal of the task is to correctly predict the interaction between drug-protein pairs given the input sentence and the mentions of drug and protein.

\subsubsection{Methods}

The model that utilizes the description and structure of the protein and drug entities is proposed.

\begin{figure}[t]
    \begin{center}   
        \includegraphics[width=.9\linewidth]{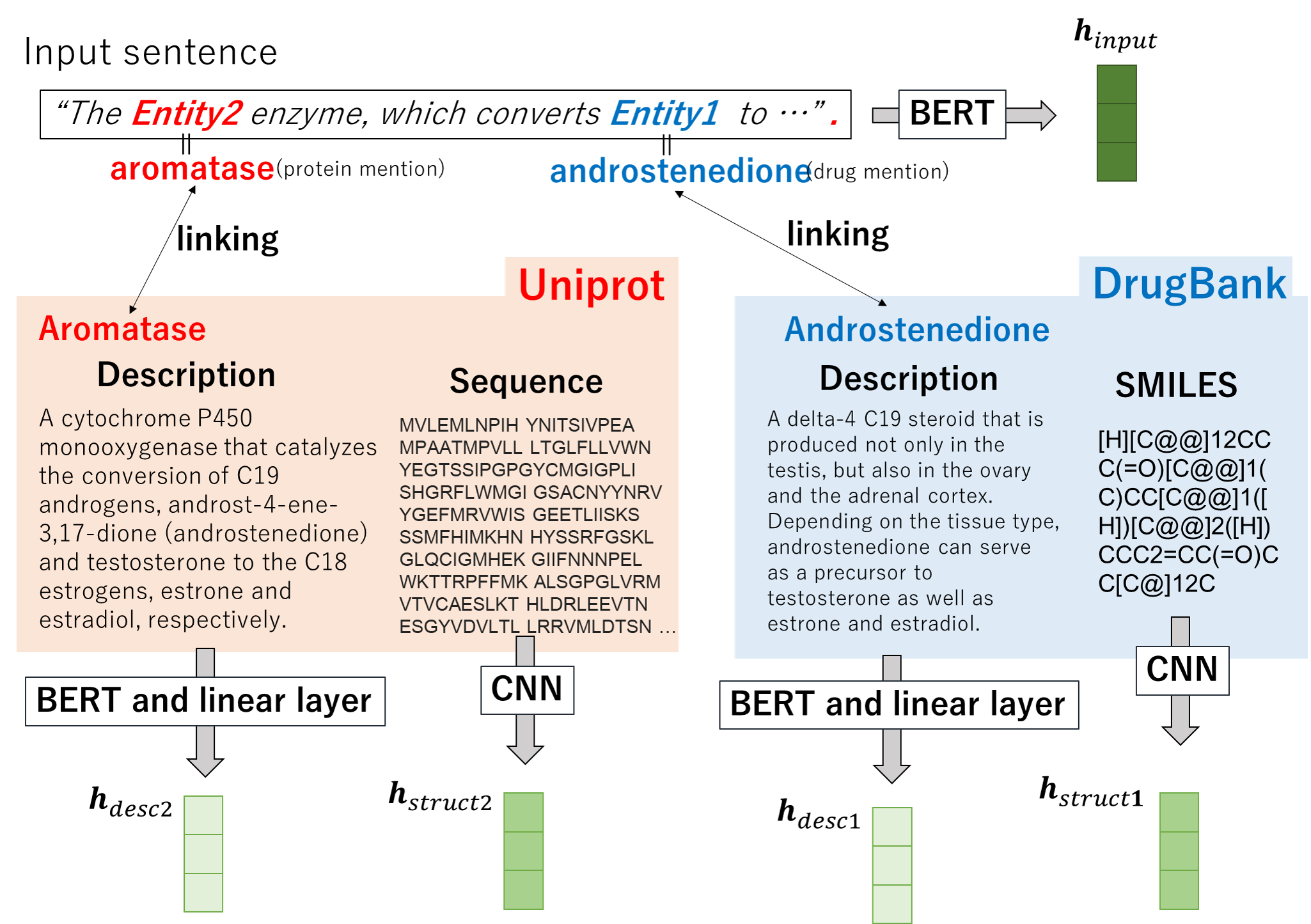}
        \caption{The model with description and structure information}
        \label{fig:Asada_model}
    \end{center}
\end{figure}

\paragraph{Utilizing descriptions and structures of entities}
\label{subsec:desc_struct}
The first model utilizes the descriptions and structures of the protein and drug entities. This model is based on the drug-drug interaction extraction method of Asada et al.~\cite{asada-bioinformatics} and the model for drug-protein interaction extraction has been extended.
The drug and protein mentions in an input sentence are linked to the databases DrugBank~\cite{drugbank5} and Uniprot~\cite{uniprot}, respectively. The textual information and information of structure registered in the database are then used for relation extraction.
Figure~\ref{fig:Asada_model} shows the overview of the model.

The mentions in the sentence and database entries are linked by relaxed string matching. For each drug, the ``description'' item registered in DrugBank is used as the description information and the ``SMILES''~\cite{weininger1988smiles} item is used as information of structures. For each protein, the ``function'' item registered in Uniprot is used as the description information and the ``sequence'' item is used as the information of structures.

\paragraph{Encoding input sentences}
Each preprocessed input sentence is fed into a BERT encoder and the embedding of [CLS] token is used as the input sentence representation vector $\textbf{h}_{input}$. $\textbf{h}_{input}$ is then taken as the input of the fully connected layer and obtains the $d_r$-dimensional vector $\textbf{h}^{fc}_{input}$, where $d_r$ is the number of relation labels including the negative label. 
It should be noted here that although DrugProt data set contains multiple labels for an instance, the approach cannot predict the multiple labels correctly. However, since it is known that the number of instances of multiple labels is extremely small in the exploratory experiment, such a standard classification approach is conducted.

$\textbf{h}^{fc}_{input}$ is converted into the probability of possible relations by a softmax function $\textbf{p}_{input}=\textrm{softmax}(\textbf{h}^{fc}_{input})$.
The cross-entropy loss $L=-\sum\textbf{y} \textrm{log}\textbf{p}_{input}$ is used as the loss function, where $\textbf{y}$ is the gold type distribution. $\textbf{y}$ is a one-hot vector where the probability is 1 for the correct label and 0 otherwise.

\paragraph{Encoding description information}
The descriptions registered in the database are also encoded by BERT in the same way as the input sentence of the corpus. A separate BERT is prepared for database descriptions. 
The vector $\textbf{h}_{desc\_CLS}$ of the BERT [CLS] token is converted into a $d_d$-dimensional vector $\textbf{h}_{desc}$ as follows:
\begin{equation}
    \textbf{h}_{desc}=\textrm{GELU}(\textbf{Wh}_{desc\_CLS}+\textbf{b}),
\end{equation}
where the $\textrm{GELU}$ is an activation function and $\textbf{W}$ and $\textbf{b}$ are weights and bias of the linear layer, respectively.
The representations of the entity 1 description $\textbf{h}_{desc1}$ and the entity 2 description $\textbf{h}_{desc2}$  and the input sentence $\textbf{h}_{input}$ are concatenated. Then, the resulting vector is used as the input to the fully connected layer:
\begin{equation}
    \textbf{h}^{fc}_{desc} = \textrm{FC} ( [\textbf{h}_{input};\textbf{h}_{desc1};\textbf{h}_{desc2}] ),
\end{equation}
where FC is a fully-connected layer and $[;]$ denotes the vector concatenation.
$\textbf{h}^{fc}_{desc}$ is converted into the probability $\textbf{p}_{desc}$ by a softmax function, and the model parameters are updated by minimizing the loss function $L=-\sum\textbf{y} \textrm{log}\textbf{p}_{desc}$. 

\paragraph{Encoding information of structures}
For drugs, the SMILES strings are used as the information of structures. For proteins, the amino acid sequences are used.
Both SMILES strings and amino acid sequences are encoded by character-based CNNs.

First, the $d_c$-dimensional character embedding is assigned to each character of the sequence; specifically, atoms of drugs such as `C' and `N', or bonds of drugs such as `=' and `\#', amino acid symbols of proteins such as `A', `R', and `N'.
After each character of the sequence is converted to the corresponding embedding, all character embeddings are encoded as the inputs to CNNs with multiple convolutional window sizes~\cite{nguyen-grishman-2015-relation}, and max pooling is employed to obtain the whole sequence representation.
The representation of the entity 1 structure representation $\textbf{h}_{struct1}$, the entity 2 structure representation $\textbf{h}_{struct2}$, and the input sentence representation $\textbf{h}_{input}$ are concatenated to make the input of the fully connected layer:
\begin{equation}
    \textbf{h}^{fc}_{struct} = \textrm{FC} ( [ \textbf{h}_{input};\textbf{h}_{struct1};\textbf{h}_{struct2}] ),
\end{equation}
$\textbf{h}^{fc}_{struct}$ is converted into the probability $\textbf{p}_{struct}$ by a softmax function, and update the model parameters by minimizing the cross-entropy loss.

\paragraph{Inference}
Finally, description and structure information are confined using an ensemble technique when predicting the drug-protein relation label.
The final prediction is obtained by averaging the prediction probabilities of the three models described in the previous section as follows:
\begin{equation}
    \textbf{p}_{all} = \frac{1}{3}(\textbf{p}_{input}+\textbf{p}_{desc}+\textbf{p}_{struct}).
\end{equation}
The relation label prediction is calculated as $\textrm{argmax} \textbf{p}_{all}$.

\begin{table*}[h]
\centering
\small
\begin{tabular}{l|l|rrr|rrr}
\hline
Method type & Method & \multicolumn{3}{l|}{Development} & \multicolumn{3}{l}{Test} \\ 
& & P & R & F1 & P & R & F1\\\hline
with database information & BioBERT-Large & 0.788 & 0.746 & 0.766 & - & - & -\\
& +desc & 0.770 & 0.773 & 0.772 & - & - & -\\
& +struct & 0.759 & 0.784 & 0.771 & - & - & -\\
& \textbf{1-desc\_struct} & 0.772 & 0.778 & 0.775 & 0.749 & 0.777 & \textbf{0.763}\\\hline
with distant supervised data & PubMedBERT & 0.776& 0.751& 0.763& - & - & -\\
& \textbf{4-ds\_pretrain} & 0.766 & 0.726 & 0.746 & 0.752 & 0.739 & 0.746\\
& \textbf{5-ds\_pretrain\_init} & 0.789 & 0.739 & 0.763 & 0.720 & 0.721 & 0.721 \\\hline
ensemble & \textbf{2-ds\_desc\_struct} & 0.791 & 0.761 & \textbf{0.776} & 0.767 & 0.755 & 0.761\\
& \textbf{3-ds\_init\_desc\_struct}& 0.780 & 0.752 & 0.766 & 0.765 & 0.746 & 0.756\\
\hline
  \end{tabular}
  \caption[Micro-averaged F-scores on DrugProt development set and test set]{Micro-averaged F-scores on DrugProt development set and test set. The results on the test set are shown only for the five submitted models. Bold is the best F-score.}
  \label{tab:main_results}
\end{table*}

\begin{table*}[t]
    \centering
    \small
    \begin{tabular}{lrrrrr}
    \hline
    Development &\textbf{1-de.\_st.}&\textbf{2-ds\_de.\_st.}&\textbf{3-in.\_de.\_st.}&\textbf{4-ds\_pr.}&\textbf{5-ds\_pr.\_in.}\\
    \hline\hline
    ACTIVATOR             &0.754&0.766&0.748&0.728&0.748\\
    AGONIST               &0.770&0.783&0.785&0.769&0.780\\
    AGONIST-A.     &0.000&0.000&0.000&0.571&0.000\\
    AGONIST-I.     &0.000&0.000&0.000&0.667&0.000\\
    ANTAGONIST            &0.915&0.931&0.916&0.916&0.925\\
    DIRECT-REGULATOR      &0.658&0.638&0.613&0.583&0.620\\
    INDIRECT-D.  &0.758&0.772&0.742&0.747&0.778\\
    INDIRECT-U.  &0.775&0.780&0.741&0.691&0.761\\
    INHIBITOR             &0.859&0.850&0.854&0.844&0.843\\
    PART-OF               &0.733&0.730&0.748&0.681&0.703\\
    PRODUCT-OF            &0.602&0.637&0.603&0.549&0.611\\
    SUBSTRATE             &0.728&0.726&0.713&0.690&0.703\\
    SUBSTRATE\_P. &0.000&0.000&0.000&0.000&0.000\\\hline
    macro-average         &0.580&0.585&0.574&0.649&0.575\\
    micro-average         &0.775&0.776&0.766&0.746&0.763\\
    \hline
    \end{tabular}
    \caption[F-scores per class on DrugProt development set]{F-scores per class on DrugProt development set. AGONIST-A., AGONIST-I., INDIRECT-D., INDIRECT-U. and SUBSTRATE\_P. stand for AGONIST-ACTIVATOR, AGONIST-INHIBITOR, INDIRECT-DOWNREGULATOR, INDIRECT-UPREGULATOR and SUBSTRATE\_PRODUCT-OF, respectively.}
    \label{tab:BCVII_results-class}
\end{table*}

\subsubsection{Experiments}

\paragraph{Models}

The five models have been submitted for BioCreative VII competition:
\begin{description}
    \item[1-desc\_struct]A model using the description and structure information of protein/drug entity.
    \item[2-ds\_desc\_struct]The ensemble of model 1 and 4
    \item[3-ds\_init\_desc\_struct]The ensemble of model 1 and 5
    \item[4-ds\_pretrain]A model using distant supervised data. All parameters are pre-trained on distant supervised data. 
    \item[5-ds\_pretrain\_init]A model using distant supervised data. Layers other than the fully connected layer are initialized with parameters pre-trained on distant supervised data.
\end{description}

\paragraph{Experimental settings}
For the model \textbf{1-desc\_struct}, BioBERT-Large~\cite{Jinhyuk2019biobert} is used as the text encoder. 
The BioBERT-Large was prepared separately for the input sentence and the entity description, and they are fine-tuned during training. The maximum sentence length was set to 128 for both the input sentence and the entity description. The dimension size $d_d$ was set to 32.

For drugs, string matching was performed for the entry names, synonyms, product names and brand names in DrugBank. For proteins, string matching was performed for the entry names, recommended names, alternative names, and gene names in Uniprot.
As a result, 94\% and 99\% of drug and protein mentions in the train data set matched the DrugBank and Uniprot entries.
When the entity could not be linked to database entries or the description and structure information is not registered in the database, an empty string is used as the input of BERT and CNNs, that is, all tokens are replaced with the padding token.

In the information of structures encoding using character CNNs, the maximum sequence length of SMILES was set to 200 and that of amino acid sequences was set to 1,500.
For both drugs and proteins, the character embedding dimension size $d_c$ was set to 100, the convolution output vector dimension size was set to 16, and the convolution window size was set to [3,5,7]. Since three convolution windows are used, the dimension sizes of the structure vectors $\textbf{h}_{struct1}$ and $\textbf{h}_{struct2}$ are both $16\times 3=48$.

\paragraph{Results}
The performance of proposed models is evaluated on the development set and test set in Table~\ref{tab:main_results}.
Regarding the method using the description and structure information of the database, the F-score is improved in the development data set in both the cases where the description information and the structure information is used individually, compared with the baseline BioBERT-Large model.
The model \textbf{1-desc\_struct}, which uses both description and structure information, further improved the F-score from baseline model.

Table~\ref{tab:BCVII_results-class} shows the F-score for each interaction type. For most of the classes of F-scores and a micro average, \textbf{5-ds\_pretrain\_init} showed better performance than \textbf{4-ds\_pretrain}. On the other hand, for AGONIST-ACTIVATOR and AGONIST-INHIBITOR with less training data, \textbf{4-ds\_pretrain}, which initializes all weights with weights pre-trained on distantly supervised data, showed higher performance.

%
%

\subsection{Summary}
A novel neural method for relation extraction from text using large-scale raw text information and drug database information, especially the drug descriptions and the drug molecular structure information, is proposed. The results show that the large-scale raw text information with SciBERT greatly improves the performance of DDI extraction from texts on the data set of the DDIExtraction-2013 shared task.
In addition, either the drug descriptions and the molecular structures can further improve the performance for specific DDI types, and their simultaneous use can improve the performance on all the DDI types.

\def\OurKG{PharmaHKG}

\section[Representing Heterogeneous Knowledge Graph]{Representing Heterogeneous Knowledge \newline Graph} \label{sec:hetero_kg}

This chapter includes work from the published paper Asada et al. (2021)~\cite{Asada2021-uv}.

\subsection{Background: An Overview of Knowledge Graph Representation}
Recently, obtaining the representation of Knowledge Graph (KG) elements in a dense vector space has attracted a lot of research attention.  Major advances in the KG representation learning model, which expresses entities and relationships as elements of a continuous vector space, are witnessed. 
The vector space embedding of all elements in KGs has received considerable attention because it is used to create a statistical model of the whole KGs, i.e., to easily calculate the semantic distance between all elements and to predict the probabilities of possible relational events (i.e., edges) on the graph. Such models can be used to infer new knowledge from known facts (i.e., link prediction), to clarify entities (i.e., entity resolution), to classify triples (i.e., triple classification), and to answer the probability question and answering \cite{nickel2011three,bordes2011learning,socher2013reasoning,nickel2016holographic}. They can enhance knowledge learning capabilities from the perspectives of knowledge reasoning, knowledge fusion, and knowledge completion~\cite{pham2018learning,xie2016representation,lin2016knowledge,ji2020joint}.

Applications of the KG are often severely affected by data sparseness; however, a typical large-scale KG is usually far from perfect. The task of completing the KG aims to enrich the KGs with new facts. Many graph-based methods have been proposed to find new facts between entities based on the network structure of KG~\cite{lao2011random}. Much effort has also been put into extracting relevant facts from plain text~\cite{zeng-etal-2014-relation}. However, these approaches do not utilize KG information.  
Neural-based knowledge representations have been proposed to encode both entities and relationships in a low-dimensional space where new facts can be found in \cite{lin2015learning,bordes2013translating}. While traditional methods often deal with KGs without node types, in many real world data, entities have different semantic types. Recent methods deal with heterogeneous KGs with different types of nodes~\cite{schlichtkrull2018modeling,wang2019heterogeneous}.  More importantly, neural models can be used to perform learning of text and knowledge within a unified semantic space to more accurately complete the KG~\cite{han2016joint}. 

Nowadays, there has been a lot of interest in jointly learning KG and embedding textual information. However, traditional KG models based on representation learning only use the information of molecular structures embedded in a particular KG. Plain text textual information, on the other hand, provides a wealth of semantic and contextual information that can contribute to the clarity and completion of entity representations and relationship representations of a given KG. Therefore, textual information can be seen as an effective supplement to the completed task of the KG. To explore the informative semantic signals of plain text, there has recently been a great deal of interest in learning together the embeddings of KG and text information in \cite{toutanova2015representing}. Moreover, the researchers provided a text-enhanced KG representation model that utilized textual information to enhance the knowledge representations~\cite{wang2020model}.

KGs have attracted great attention from both academia and industry as a means of representing structured human knowledge. Various kinds of KGs have been proposed such as Freebase~\cite{bollacker2008freebase}, YAGO~\cite{suchanek2007yago}, and WordNet~\cite{miller1995wordnet}.
A KG is a structured representation of facts that consists of entities, relations, and semantic descriptions. Entities are real-world objects and abstract concepts, relations represent relationships between entities, and semantic descriptions of entities, and these relationships include types and properties that have well-defined meanings. The KG usually consists of a set of triples $\{(h, r, t)\}$, where $h, r$, and $t$ represent the head entity, relationship, and tail entity, respectively.

Based on the above motivation, this chapter investigates a heterogeneous pharmaceutical knowledge graph containing textual information constructed from several databases. The heterogeneous entity items consisting of drug, protein, category, pathway, and Anatomical Therapeutic Chemical (ATC) code, and relations among them, which include category, ATC, pathway, interact, target, enzyme, carrier, and transporter, are constructed. Three methods are compared to incorporate text information in KG embedding training with representing text with BERT. The resulting node and edge embeddings are evaluated by the link prediction task and the usefulness of using text information in KG embedding training is verified.  
The study of KG completion is roughly divided into two types: a study in which the link prediction task is performed by using score functions such as TransE~\cite{bordes2013translating}, DistMult~\cite{yang2014embedding}, and a study in which Graph Convolutional Networks~\cite{DBLP:conf/iclr/KipfW17} etc, are applied to the whole KG, and the node classification task is performed. In this study, the link prediction task is in focus and the usefulness of text information in scoring function-based link prediction tasks is investigated.

The contributions are summarized as follows:
\begin{itemize}
    \item A heterogeneous KG with textual information (called {\em \OurKG{}}) in the drug domain is proposed. This can be used to develop and evaluate heterogeneous knowledge embedding methods.
    \item Three methods to incorporate text information into KG embedding models are proposed.
    \item The combinations of four KG embeddings models and three methods are evaluated and compared on the link prediction task in the proposed KG, and it is shown that there is no single method that can perform best for different relations and the best combination depends on the relation type.
\end{itemize}

\subsection{Heterogeneous Pharmaceutical Knowledge Graph with Textual Information}

In this section, a heterogeneous pharmaceutical KG \OurKG{} that is constructed in this paper is first introduced. 
Then, the definition of KG and the learning method of embeddings in the KG are explained. 
Finally, the proposed method that effectively uses text information for KG representation learning is explained.

\begin{figure*}[t!]
\begin{center}
\includegraphics[width=\linewidth]{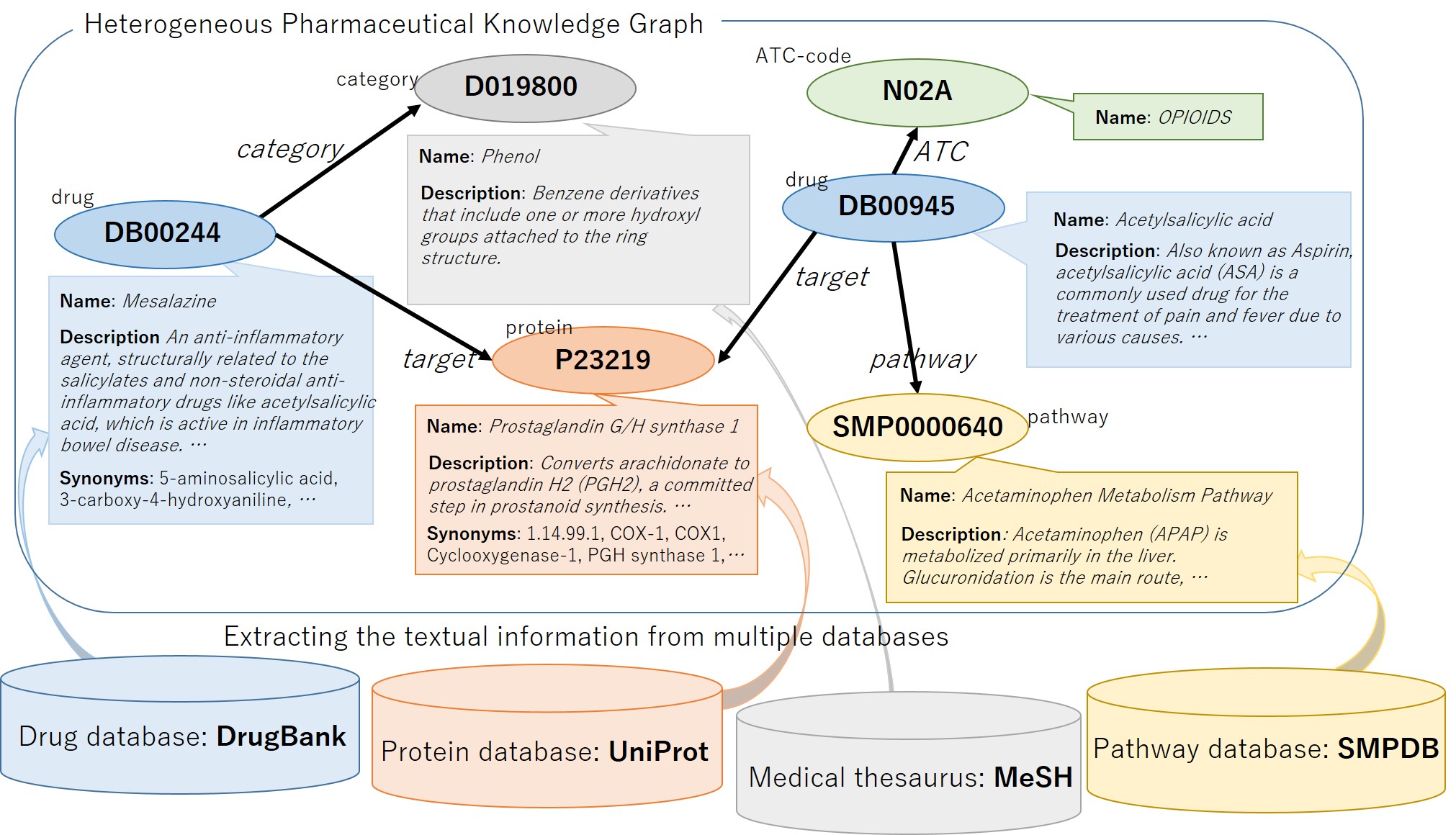}
\end{center}
\caption{Illustration of the Heterogeneous Pharmaceutical Knowledge Graph}\label{fig:consruct_kg}
\end{figure*}

A heterogeneous pharmaceutical KG with textual information is constructed from DrugBank~\cite{drugbank5} and its related data sources. 
DrugBank is one of the rich drug databases. It contains several different types of nodes, which can be a good source for a heterogeneous knowledge graph. The nodes are related to several textual information in DrugBank and their linked entries in several other data sources such as UniProtKG~\cite{uniprot}, Small Molecule Pathway Database (SMPDB)~\cite{smpdb2.0} and medical thesaurus Medical Subject Headings (MeSH). The existence of such textual information fits the objective to evaluate the utility of textual information in KG representation.
The KG and the related data sources are illustrated in Figure~\ref{fig:consruct_kg}. This section first explains the nodes and relations in the KG and then explains the textual information.

\subsubsection{Constructing Heterogeneous Pharmaceutical Knowledge Graph}

A KG consisting of five different types of heterogeneous items, i.e., drug, protein, pathway, category, and ATC code, is constructed from different databases and thesaurus. The number of nodes is shown in Table~\ref{table:kg_ent_stats}.

\begin{table}[t]
  \centering
  \begin{tabular}{lr} \\\hline
  Entity type & \# \\\hline
  Drug (DrugBank-ID) & 11,516 \\
  Protein (Uniprot-ID) & 5,339 \\
  Pathway (SMPDB-ID) & 874 \\
  Category (MESH-ID) & 2,166 \\
  ATC (ATC-code) & 1,093 \\\hline
  Total	& 20,988 \\\hline
  \end{tabular}
  \caption{Statistics of heterogeneous pharmaceutical KG entities\label{table:kg_ent_stats}}
\end{table}

\begin{itemize}
    \item \textbf{Drug}: information of drugs is extracted from DrugBank~\cite{drugbank5}. More than 10,000 drugs are registered in DrugBank, and various types of information such as drug names, descriptions, molecular structures and experimental properties are registered.
    \item \textbf{Protein}: The information of proteins is extracted from UniProtKG~\cite{uniprot}. UniProtKG consists of the Swiss-Prot which is manually annotated and reviewed and TrEMBL which is automatically annotated and not reviewed, and the Swiss-Prot knowledge base is used.
    \item \textbf{Pathway}: Information of pathways from 
Small Molecule Pathway Database (SMPDB)~\cite{smpdb2.0} is extracted.
SMPDB is an interactive, visual database containing more than 30,000 small molecule pathways found in humans. 
 \item \textbf{Category}: Information of drug categories is extracted from medical thesaurus Medical Subject Headings (MeSH)~\cite{mesh}. Each drug recorded in DrugBank has several hypernymy categorical classes and these classes have MeSH term ID. As an example, a drug \textit{Morphine} has categories such as \textit{Alkaloids} (MeSH ID:D000470), \textit{Anesthetics} (MeSH ID:D018681), and the detailed information can be obtained by referring to MeSH.
\item \textbf{ATC}: Anatomical Therapeutic Chemical (ATC) classification system also has categorical information of drugs. In the ATC classification system, drugs are divided into different groups according to the organ or system on which they act and their therapeutic, pharmacological, and chemical properties. Drugs are classified in groups at five different levels. The drugs are divided into fourteen main groups (first level), with pharmacological or therapeutic subgroups (second level). The third and fourth levels are chemical/pharmacological/theraperutic subgroups and the fifth level is the chemical substance. For example, the drug ``Metformin'' is classified into ``A: Alimentary tract and metabolism'' (first level), ``A10: Drugs used in diabetes'' (second level), ``A10B: Blood glucose lowering drugs, excl. insulins'' (third level), ``A10BA: Biguanides'' (fourth level) and ``A10BA02: metformin'' (fifth level).
\end{itemize}

\begin{table}[t]
  \centering
  \begin{tabular}{lrrrrr} \hline
  Relation type & ALL & train & valid & test \\\hline
  category & 60,459 & 54,419 & 3,020 & 3,020\\
  ATC & 16,341 & 14,711 & 815 & 815\\
  pathway & 18,707 & 16,847 & 930 & 930\\
  interact & 2,682,142 & 2,413,932 & 134,105 & 134,105\\
  target & 18,467 & 16,627 & 920 & 920\\
  enzyme & 5,206 & 4,686 & 260 & 260\\
  carrier & 815 & 735 & 40 & 40\\
  transporter & 3,093 & 2,793 & 150 & 150\\
  \hline
  Total & 2,750,228 & 2,525,829 & 140,240 & 140,240\\\hline
  \end{tabular}
  \caption{Statistics of heterogeneous pharmaceutical KG edges for each relation type
  \label{table:kg_rel_stats}}
\end{table}

Five different types of nodes are connected by the following eight types of relations: \textit{category}, \textit{ATC},  \textit{pathway}, \textit{interact}, \textit{target}, \textit{enzyme}, \textit{carrier}, and \textit{transporter}. The statistics of the KG edges for each relation type are shown in Table~\ref{table:kg_rel_stats}.
The relation triples from DrugBank are extracted.

Drug nodes and MeSH categorical terms are linked by \textit{category} relation. 
\begin{itemize}
\item \textit{category}: This relation type indicates the MeSH category of drugs. This relationship indicates that the drug is classified into the therapeutic category or the general category (anti-convulsant, antibacterial, etc.) defined by MeSH. These relationships are registered by the manual search of DrugBank developers.

\end{itemize}

Drug nodes and ATC classification system codes are linked by \textit{ATC} relation. 
In order to incorporate hierarchical information into the KG, ATC codes are linked to ATC codes by \textit{ATChypernym} relation. ATC codes are linked to the next higher level codes with this relation. 
Relational triples such as A10BA - ATChypernym - A10B, N02 - ATChypernym - N by linking the ATC code of the next higher level are created. Since this relation is apparent from the surface strings of ATC codes, this relation for link prediction is not considered. 
\begin{itemize}
    \item \textit{ATC}: Drugs are linked to any level of ATC codes with this relation.
    In DrugBank, drug elements may have one or more ATC-code elements, e.g., drug \textit{Morphine} has four ATC codes (A07DA52, N02AA51, N02AA01 and N02AG01), and each ATC-code element has child elements. All these child entities and the drug entity are connected by the \textit{ATC} relation. 
\end{itemize}

Drug nodes and protein nodes are also connected with pathways.
\begin{itemize}
\item \textit{pathway}: This relation type indicates a drug or protein is included in a pathway.
When the drug is involved in metabolic, disease, and biological pathways as identified by the SMPDB, the drug entity and the pathway entity is connected by the \textit{pathway} relation.
Also, when the enzyme protein is involved in the same pathways, the protein entity and the pathway entity are connected by the \textit{pathway} relation.
\end{itemize}

Drug nodes can be connected by a relation \textit{interact}.
\begin{itemize}
\item \textit{interact}: A triple of this relation type indicates that the drug pair has a DDI.
When concomitant use of the pair of drugs will affect its activity or result in adverse effects, these two drug entities are connected by \textit{interact} relation. These interactions may be synergistic or antagonistic depending on the physiological effects and mechanism of action of each drug.
\end{itemize}

Drug nodes and protein nodes can be linked by \textit{target}, \textit{enzyme}, \textit{carrier}, or \textit{transporter} relation~\cite{drugbank5}.
\begin{itemize}
\item \textit{target}: A protein, macromolecule, nucleic acid, or small molecule to which a given drug binds, resulting in an alteration of the normal function of the bound molecule and a desirable therapeutic effect. Drug targets are most commonly proteins such as enzymes, ion channels, and receptors.
\item \textit{enzyme}: A protein that catalyzes chemical reactions involving a given drug (substrate). Most drugs are metabolized by the Cytochrome P450 enzymes.
\item \textit{carrier}: A secreted protein that binds to drugs, carrying them to cell transporters, where they are moved into the cell. Drug carriers may be used in drug design to increase the effectiveness of drug delivery to the target sites of pharmacological actions.
\item \textit{transporter}: A membrane bound protein that shuttles ions, small molecules, or macromolecules across membranes, into cells or out of cells.
\end{itemize}

\subsubsection{Textual Information of Knowledge Graph}

Here, the text information relating to each type of node is explained. 
\begin{itemize}
    \item \textbf{Drug}: Drugs are assigned a unique DrugBank-id. Various text information contained in the DrugBank xml file is used.
    ``Name'', to the heading of the drug and standard name of the drug as provided by the drug manufacturer, ``Description'', which describes the general facts, composition and/or preparation of the drug, ``Indication'' is a description or common names of diseases that the drug is used to treat,  ``Pharmacodynamics'' is a description of how the drug works at a clinical or physiological level, ``Mechanism of Action'' is a description of how the drug works or what it binds to at a molecular level, ``Metabolism'' is a mechanism by which or organ location where the drug is neutralized, and ``Synonyms'' indicates alternate drug names.
    \item \textbf{Protein}: Protein targets of drug actions, enzymes that are inhibited/induced or involved in metabolism, and carrier or transporter proteins involved in the movement of the drug across biological membranes. Each of \textit{targets}, \textit{enzymes}, \textit{carriers}, \textit{transporters} have unique UniProt-id. 
    UniProt-id is referred to and the following types of textual information, functions, miscellaneous description, short name, alternative names, and gene names are obtained.
    \item \textbf{Pathway}: Pathway relations are extracted from DrugBank. Each pathway has a unique ID defined by The Small Molecule Pathway Database (SMPDB)~\cite{smpdb2.0}. ``Name'' and ``Description'' of the pathway are registered in SMPDB.
    \item \textbf{Category}: Drug categories are classified according to the medical thesaurus MeSH. This textual information is registered in MeSH: ``Name'' is a definition word, ``ScopeNote'' is a term description, ``Entry terms'' is a synonym.
    \item \textbf{ATC}: Drugs are classified in a hierarchy with five different levels by WHO drug classification system (ATC) identifiers. Each level of ATC classification code has a name, which is defined as the international nonproprietary name (INN) or the name of the ATC level. These names given to ATC codes as textual information are used.
\end{itemize}

\subsection{Learning Knowledge Graph Embeddings}

\subsubsection{Knowledge Graph Definition}

A heterogeneous KG is treated as a directed graph whose nodes and edges have semantic types. The semantic types are assigned to different types of nodes (drug, protein, pathway, etc.) and relations (target, carrier, etc.) to represent detailed information about nodes and relations. 
A KG is defined as a directed graph $\mathcal{G}=(E,R,F)$, where the nodes $E$ denotes the set of typed entities, $R$ refers to the set of typed relations and $F$ represents the set of facts (i.e., directed edges). The nodes are often called entities. The facts or directed edges are often called triplets and are represented as a $(h, r, t)$ tuple, when $h$ is the head entity, $t$ is the tail entity and $r$ is the relation from the head entity to the tail entity.

\subsubsection{Scoring Functions}\label{sec:scoring_functions}
The methods that represent the KG by using embeddings of entities and relations can catch the structure information of the KG and provide structure-based embeddings. Entities and relations are directly represented as real-valued vectors, matrices or complex-valued vectors.
Scoring function $f(h,r,t)$ is defined on each triple $(h,r,t)$ to access the validity of triples. Triples observed in the KG tend to have higher scores than those that have not been observed. The following four scoring functions are employed.

\paragraph*{TransE}
TransE~\cite{bordes2013translating} is a representative translational distance model that represents entities and relations as vectors in the same semantic space of dimension $\mathbb{R}^d$ where $d$ is the dimension of the target space with reduced dimension. A fact in the source space is represented as a triplet $(h,r,t)$. The relationship is interpreted as a translation vector so that the embedded entities are connected by relation $r$ and have a short distance. The norm is set to $2$, and the scoring function is computed as:
\begin{equation}
    f(h,r,t)=-|\mathbf{h}+\mathbf{r}-\mathbf{t}|_2.
\end{equation}

\paragraph*{DistMult}
DistMult~\cite{yang2014embedding} is a method that speeds up the RESCAL model~\cite{nickel2011three} by considering only symmetric relations and restricting $M_r$ from a general asymmetric $r\times r$ matrix to a diagonal square matrix, thus reducing the number of parameters per relation to $O(d)$. DistMult scoring function is computed as:
\begin{equation}
   f(h,r,t)=\mathbf{h}^{\mathsf{T}}\mathrm{diag}(\mathbf{r})\mathbf{t}=\sum^{d-1}_{i=0}[\mathbf{h}]_i[\mathbf{r}]_i[\mathbf{t}]_i.
\end{equation}

\paragraph*{ComplEx}
ComplEx~\cite{trouillon2016complex} uses complex vector operations to consider both symmetric and asymmetric relation.
The scoring function for complex entity and relation vectors $\mathbf{h}$, $\mathbf{r}$, and $mathbf{t} \in \mathbb{C}^d$ is computed as:
\begin{equation}
    f(h,r,t)=\mathrm{Real}\big(\mathbf{h}^{\mathsf{T}}\mathrm{diag}(\mathbf{r})\mathbf{t}\big),
\end{equation}
where $\mathrm{Real}$ extracts the real part of the complex vectors.

\paragraph*{SimplE}
SimplE~\cite{kazemi2018simple} considers two vectors $\mathbf{h}, \mathbf{t}\in \mathbb{R}^d$ as the head and tail embeddings for each entity and two vectors $\mathbf{v}_r, \mathbf{v}_{r^{-1}}\in \mathbb{R}^d$ for each relation $r$. The similarity function of SimplE for a triple $(h, r, t)$ is defined as:
\begin{equation}
    f(h,r,t)=\frac{1}{2}\big(\langle \mathbf{h}_h, \mathbf{v}_r, \mathbf{t}_t \rangle + \langle \mathbf{h}_t, \mathbf{v}_{r^{-1}}, \mathbf{t}_h \rangle\big).
\end{equation}

The above four score functions are chosen because these are widely used and cover the standard ideas for scoring relational triples: distance-based, bilinear-based and complex number-based.

\subsubsection{Negative Sampling and Loss Functions}\label{sec:negative_sampling_and_loss_functions}
Generally, to train a KG embedding, the models apply a variety of negative sampling by corrupting triplets $(h,r,t)$. They corrupt either $h$, or $t$ by sampling from the set of head or tail entities for heads and tails, respectively. The corrupted triples can be either of $(h',r,t)$ or $(h,r,t')$, where $h'$ and $t'$ are the negative examples. 
The author acknowledges that due to the incompleteness of the current KG, the unregistered and potentially positive relational triples can be negative examples: this problem is common in most studies that tackle with the link prediction task.
To avoid easy negative examples and utilize the entity type information, the node types of negative examples is restricted depending on $r$.
The logistic loss and the margin-based pairwise ranking loss are commonly used for training.
The logistic loss returns $-1$ for negative examples and $+1$ for the positive examples. $\mathbb{D}^+$ and $\mathbb{D}^-$ are negative and positive data, $y=\pm 1$ is the label for positive and negative triples, and $f(\cdot)$ is the scoring function. Model parameters are trained by minimizing the negative log-likelihood of the logistic model with $L2$ regularization on the parameters $\Theta$ of the model;
\begin{equation}
    L_{KG}\sum_{(h,r,t)\in \mathbb{D}^+ \cup \mathbb{D}^-}\log \big(1+\exp (y\times f(h,r,t))\big)+\lambda||\Theta||^2_2.
\end{equation}
The margin-based pairwise ranking loss minimizes the rank for positive triples. Ranking loss is given by:
   \begin{equation}
    L_{KG}=\sum_{(h,r,t)\in \mathbb{D}^+}\sum_{(h',r',t')\in \mathbb{D}^-}\max(0,\gamma-f(h,r,t)+f(h',r',t'))+\lambda||\Theta||^2_2.
\end{equation}

\subsection{Methods}

\begin{figure*}[t!]
\begin{center}
\includegraphics[width=.9\linewidth]{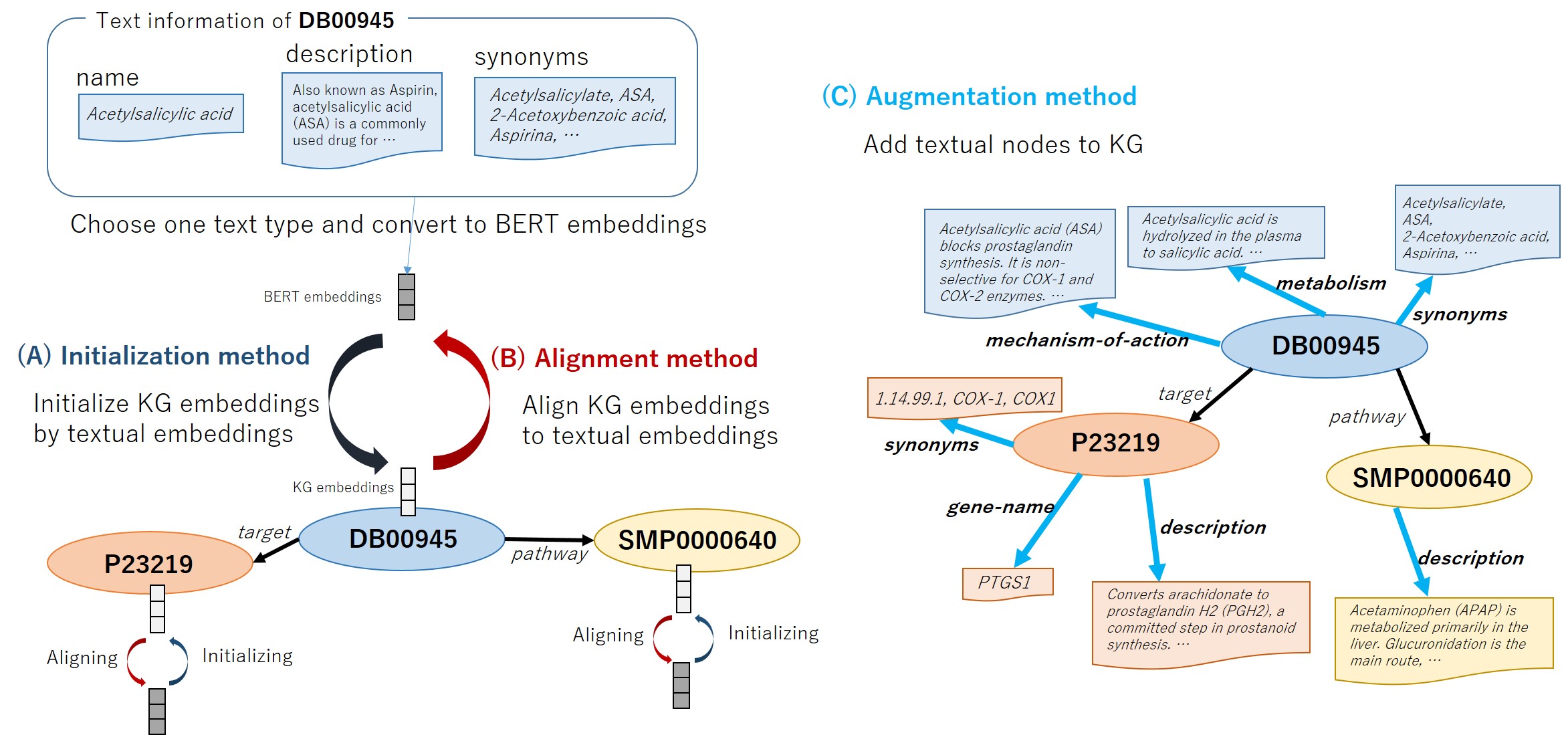}
\end{center}
\caption{Overview of methods: (A) Initializing node embeddings (Initialization), (B) Aligning entity embeddings and textual embeddings (Alignment), and (C) Augmenting KG embeddings (Augmentation)}\label{fig:methods}
\end{figure*}

This section verifies the usefulness of using text information in KG embedding training by three methods explained below.
Figure~\ref{fig:methods} shows the overview of the three methods that utilize text information for KG embedding representation.
BERT~\cite{devlin-etal-2019-bert}, which is an extremely high-performance contextual language representation model, is employed in encoding text. BERT is pre-trained with the masked language model objective and next sentence prediction task objective on large unlabeled corpora, and fine-tuned BERT towards the target task achieved the state-of-the-art performance.

\subsubsection{Initializing Node Embeddings}
Usually, the initial value of embedding for each node in the KG is given randomly in the existing methods. As shown in Figure~\ref{fig:methods} (A), first, the type of text to use, e.g., drug nodes have text types such as Name, Description and Synonyms, is selected.
Then, the selected text is taken as the input of the text encoder model BERT and the \texttt{<CLS>} embeddings of the BERT as the initial value of the node embeddings.
For the methods that use two embeddings for an entity, i.e., ComplEx (real and imaginary embeddings) and SimplE (head and tail embeddings), both vectors with the \texttt{<CLS>} embeddings are initialized.
When multiple text items are registered (e.g., the drug Acetaminophen has multiple synonyms, ``Acenol'', ``APAP'', ``Paracetamol''), these terms are connected with a comma and taken as an input for BERT. 
This method is called the \textit{Initialization} method.
The motivation of the \textit{Initialization} method is to help represent node embeddings by using the BERT embeddings that pre-trained on a large amount of biomedical literature.
Correct relational triples are aimed to predict from textual information by BERT even if the information of structures in the graph is insufficient.

\subsubsection{Aligning Entity Embeddings and Textual Embeddings}

The aligning method aims to gradually project KG embeddings into textual embeddings space by adding the regularization term to the loss function.
\begin{equation}
    L_{a} = \lambda_{a}|| V_{KG}-V_{text} ||_2,
\end{equation}
\begin{equation}
    L = L_{KG} + L_{a},
\end{equation}
where $\lambda_{a}$ is a regularization coefficient of alignment, $V_{KG}$ and $V_{text}$ are vector lookup table matrices of KG and textual embeddings respectively. 
Similar to the initialization method, the textual embeddings are obtained from BERT and when there are two embeddings for an entity, both vectors are regularized.
This method is called the \textit{Alignment} method.
The motivation of the \textit{Alignment} method is that as the updating the node representation progresses, the two spaces of the text embeddings and the graph structure embeddings are projected into the same space, and finally more suitable node representations are obtained.

\subsubsection{Augmenting KG Embeddings}
In this method, as shown in Figure~\ref{fig:methods} (C), the KG structure is augmented by adding relation triples based on the text information of the node.
The node's own embedding is initialized with textual embeddings of Name. The embedding value of linked nodes is initialized with the BERT output.
Moreover, since ATC classification codes have a hierarchical structure as shown in Figure~\ref{fig:methods} (C), after extending the link from the drug node to create new categorical nodes, further linking is made between the categorical nodes. 
A graph that can consider both text information and the hierarchical information is constructed.
This method is called the \textit{Augmentation} method.
The motivation of the \textit{Augmentation} method is to consider multiple text information of one entity at once.

\subsection{Experimental Settings}
\subsubsection{Constructing Heterogeneous KG with Textual Information}
The overview of constructing a heterogeneous KG with textual information is shown in Figure~\ref{fig:consruct_kg}.
Four publicly available databases, DrugBank, UniProt, MeSH term descriptions and SMPDB are downloaded, and first DrugBank is processed and relations between the drug and other heterogeneous items are extracted. Here, the text information on each drug is also extracted and associated with the entity ID in the KG.
Next, for entities other than drugs, the link ID of DrugBank is used to refer to other databases and associated the text information with the entity ID in the KG.
As a result, five types of entities (i.e., drug, protein, pathway, category, and ATC) are included in the constructed KG. Between entities, there are relation links: category, ATC, pathway, interact,  target, enzyme, carrier, and transporter. 
The total number of relational triples is about 2.7M, and as shown in Table~\ref{table:kg_rel_stats}, the number of drug-interact-drug triples is large and accounts for the majority of them.
Note that only the relation drug-interact-drug is symmetric, and the other relations are asymmetric, that is, when there is a DrugA-interact-DrugB relation triple in the KG, there is also a DrugB-interact-DrugA triple.

\subsubsection{Encoding Text Information}

PubMedBERT~\cite{pubmedbert} is employed to encode textual information into fixed-length real-valued embeddings. 
PubMedBERT is a model that uses 21B words of the PubMed corpus for pre-training, and shows high performance in several NLP tasks in the biomedical domain.
In this paper, texts such as names and descriptions are used as inputs for pre-trained PubMedBERT, and the output \texttt{<CLS>} token embedding is used as a textual representation.
The maximum length of the input subword is set to 512.

\subsubsection{KG Embedding Training Settings}
Four KG embedding scoring functions are employed as explained in Section~\ref{sec:scoring_functions}. 
For each of the scoring functions, three methods are applied to train embeddings using textual information; the initialization, aligning and augmenting methods.

\begin{table}[t]
  \centering
  \begin{tabular}{lrrrrr} \hline
   & Name (\%) & Description (\%) & Synonyms (\%)  \\\hline
  drug & 100 & 53.72 & 48.50  \\
  protein & 100 & 96.17 & 100 \\
  category & 100 & 94.01 & 91.42 \\
  ATC & 100 & - & - \\
  pathway & 100 & 100 & - \\\hline
  \end{tabular}
  \caption[The percentage of nodes that have each type of text]{The percentage of nodes that have each type of text. Nodes in all databases have Name text information. While many proteins and categories have Description information and Synonyms information, the percentage of drugs that have this information is low.}
  \label{table:coverage_of_text_item}
\end{table}

The ratio of nodes that have each textual information is shown in Table~\ref{table:coverage_of_text_item}.
The node has the text information of Name in any database. In UniProt, most proteins have Description and Synonyms texts information, and many categorical terms in MeSH also have Description and Synonyms. On the other hand, some drugs in DrugBank do not have some text information.
In the Initialization method and Alignment method, one text type is selected and the embeddings of textual information are used\footnote{the combination of different text information in these methods is left for future work}. When the node does not have text information, the text of Name is used instead.

Drugs and proteins have textual information that other nodes do not have, and their coverage is as follows: 32.61\% of drugs have Indication information, 24.60\% of drugs have Pharmacodynamics information, 18.40\% of drugs have Metabolism information, 30.52\% of drugs have Mechanism-of-action information and 96.05\% of proteins have Gene-name. These text items are linked to the KG nodes in the Augmentation method, so the Augmentation method can utilize all text information.

The random initialization method is prepared without textual information (\textit{No Text}) as the baseline.
In this setting, embeddings of entities and relations are initialized with the random values drawn from a uniform distribution between $\pm (\gamma + \frac{\epsilon}{d})$, where $\gamma=12$, $\epsilon=2$ and $d$ is a dimension of KG embeddings. 

\subsubsection{Task Setting}
\label{sec:task_setting}
The quality of node and edge embeddings are evaluated by the link prediction task. 
Link prediction is a task to search for an entity that probably constructs a new fact with another given entity and a specific relation. For KGs are always imperfect, link prediction aims to discover and add missing knowledge into it. With the existing relations and entity, candidate entities are selected to form a new fact. The head or tail of the triples in the validation or test data set are replaced with other entities that have the same entity types and calculate the scores of all created negative triples in the KG. The calculated positive triple score and the scores of all negative triples are sorted and the rank of the positive triple score is evaluated.
Mean reciprocal rank (MRR) is used as an evaluation metric.
When negative example triples are created, if there are correct triples that exist in the KG, such triples from the ranking are excluded.
This evaluation setting has been adopted in many existing studies as a \textbf{filtered} setting~\cite{bordes2013translating, trouillon2016complex, kazemi2018simple}.
In addition, similar to the negative sampling setting during training, given the relational edge label, the node types of head or tail are trivial, so triples with inappropriate combinations of edge and node types are also excluded.

The extracted approximately 2.7M relational triples are divided into 90:5:5 as the train, valid and test data sets.
In the augmentation method, relational triples created from textual nodes are added to the train data set.

\subsubsection{Hyper-parameter Settings}
Hyper-parameters are tuned by evaluating the MRR score on the validation set for each model. Hyper-parameters are chosen with following values: regularization coefficient $\lambda\in\{10^{-3}, 10^{-6}, 10^{-9}, 10^{-12}, 0\}$, alignment regularization coefficient  $\lambda_a \in \{10^{-3}, 10^{-6}, 10^{-9}, 10^{-12}\}$, initial learning rate $\alpha_0 \in \{0.5, 0.25, 0.1, 0.05, 0.025, 0.01\}$,
For the loss function, the pair-wise hinge loss function is adopted for TransE and DistMult and the logistic loss function for ComplEx and SimplE according to the setting of the original papers.
The KG embedding dimension is set to 768 in order to match the dimension of the output of BERT embedding.
For all models, the batch size is set to 4096 and the number of epochs to 100.

\subsubsection{Implementation Details}
\label{sec:implementation_details}

All the models are implemented by using the PyTorch library~\cite{NEURIPS2019_9015}, the DGL-KE library~\cite{DGL-KE} for KG embeddings, and the transformers library~\cite{wolf-etal-2020-transformers} for BERT. 
The original DGL-KE implementation is modified in the following point. While DGL-KE samples negative examples from all combinations of entity pairs, the proposed model excludes impossible negative instances by restricting the types of entities by the relations (e.g., a drug-interact-category triple is not created for negative examples) as explained in Section~\ref{sec:negative_sampling_and_loss_functions}.

\subsection{Results and Discussions}

\begin{table*}[h!]
  \small
  \centering
  \begin{tabular}{l|r|rrr|rrr|r} \hline
   \multicolumn{9}{c}{TransE} \\\hline
   & No Text & \multicolumn{3}{c|}{Initialization} & \multicolumn{3}{c|}{Alignment} & Augmentation \\
   & & Name & Desc. & Syn. & Name & Desc. & Syn. & \\\hline
   category    & 0.1978 & 0.2117 & 0.2120 & 0.2231 & 0.2136 & 0.2059 & 0.1913 & \textbf{0.2239}\\
   ATC         & 0.2929 & 0.2695 & 0.2571 & 0.2495 & \textbf{0.3000} & 0.2973 & 0.2872 & 0.2571\\
   pathway     & \textbf{0.6741} & 0.5674 & 0.5792 & 0.5793 & 0.6694 & 0.6711 & 0.6713 & 0.5473\\
   interact    & \textbf{0.3109} & 0.2867 & 0.2845 & 0.2843 & 0.3103 & 0.3106 & 0.3108 & 0.2644\\
   target      & 0.0802 & 0.0748 & 0.0811 & 0.0808 & 0.0808 & 0.0780 & 0.0821 & \textbf{0.0889}\\
   enzyme      & 0.3262 & 0.3067 & 0.3314 & 0.3090 & 0.3474 & 0.3564 & \textbf{0.3590} & 0.2926\\
   carrier     & \textbf{0.4155} & 0.3078 & 0.2843 & 0.3679 & 0.4037 & 0.4044 & 0.4010 & 0.3459\\
   transporter & \textbf{0.3576} & 0.3194 & 0.3104 & 0.3182 & 0.3444 & 0.3308 & 0.3391 & 0.2866\\\hline
   Avg. (Macro)& 0.3319 & 0.2930 & 0.2950 & 0.3015 & \textbf{0.3337} & 0.3318 & 0.3302 & 0.2883\\

   \hline\hline
   \multicolumn{9}{c}{DistMult} \\\hline
   & No Text & \multicolumn{3}{c|}{Initialization} & \multicolumn{3}{c|}{Alignment} & Augmentation \\
   & & Name & Desc. & Syn. & Name & Desc. & Syn. & \\\hline
   category    & 0.2539 & \textbf{0.2876} & 0.2797 & 0.2753 & 0.2679 & 0.2661 & 0.2586 & 0.2649\\
   ATC         & 0.2428 & 0.2674 & \textbf{0.2899} & 0.2639 & 0.2612 & 0.2617 & 0.2698 & 0.2904\\
   pathway     & \textbf{0.6792} & 0.5424 & 0.5542 & 0.6002 & 0.6524 & 0.6711 & 0.6615 & 0.4997\\
   interact    & 0.7730 & 0.6338 & 0.5895 & 0.6302 & 0.7911 & 0.7868 & \textbf{0.7990} & 0.6113\\
   target      & 0.0738 & 0.0866 & 0.0864 & 0.0947 & 0.0778 & 0.0758 & 0.0734 & \textbf{0.1049}\\
   enzyme      & 0.2501 & 0.2516 & 0.2358 & \textbf{0.2847} & 0.2247 & 0.2334 & 0.2140 & 0.2183\\
   carrier     & 0.2023 & \textbf{0.2254} & 0.1640 & 0.1622 & 0.1369 & 0.1311 & 0.1649 & 0.2134\\
   transporter & 0.2293 & \textbf{0.2703} & 0.1840 & 0.2190 & 0.1969 & 0.1939 & 0.1708 & 0.2062\\\hline
   Avg. (Macro)& \textbf{0.3380} & 0.3206 & 0.2979 & 0.3162 & 0.3261 & 0.3274 & 0.3265 & 0.3011\\

  \hline\hline
  \multicolumn{9}{c}{ComplEx} \\\hline
   & No Text & \multicolumn{3}{c|}{Initialization} & \multicolumn{3}{c|}{Alignment} & Augmentation \\
   & & Name & Desc. & Syn. & Name & Desc. & Syn. & \\\hline
   category    & 0.0905 & 0.3455 & \textbf{0.3495} & 0.3386 & 0.3302 & 0.0577 & 0.0611 & 0.3420\\
   ATC         & 0.3326 & 0.3463 & 0.3623 & 0.3485 & 0.3271 & 0.3425 & 0.3407 & \textbf{0.3652}\\
   pathway     & 0.6956 & 0.6856 & 0.7051 & 0.7157 & 0.7220 & 0.6963 & \textbf{0.7323} & 0.6820\\
   interact    & \textbf{0.8678} & 0.7632 & 0.7166 & 0.7802 & 0.8578 & 0.8189 & 0.8497 & 0.8230\\
   target      & 0.0496 & 0.1116 & 0.1093 & 0.1153 & 0.0859 & 0.0640 & 0.0740 & \textbf{0.1243}\\
   enzyme      & 0.2103 & 0.2256 & 0.2512 & \textbf{0.2538} & 0.2245 & 0.1907 & 0.2097 & 0.2073\\
   carrier     & 0.1533 & 0.1557 & 0.1817 & 0.1423 & 0.0994 & 0.1750 & 0.1462 & \textbf{0.1934}\\
   transporter & 0.1942 & \textbf{0.3119} & 0.2667 & 0.2593 & 0.2151 & 0.2076 & 0.2362 & 0.2801\\\hline
   Avg. (Macro)& 0.3242	& 0.3681 & 0.3678 & 0.3692 & 0.3577 & 0.3190 & 0.3312 & \textbf{0.3771}\\

  \hline\hline
  \multicolumn{9}{c}{SimplE} \\\hline
   & No Text & \multicolumn{3}{c|}{Initialization} & \multicolumn{3}{c|}{Alignment} & Augmentation \\
   & & Name & Desc. & Syn. & Name & Desc. & Syn. & \\\hline
   category    & 0.0461 & 0.3591 & 0.3536 & \textbf{0.3668} & 0.0520 & 0.3263 & 0.2619 & 0.3367 \\
   ATC         & 0.3278 & \textbf{0.3820} & 0.3617 & 0.3732 & 0.3644 & 0.3410 & 0.3425 & 0.3475 \\
   pathway     & \textbf{0.7513} & 0.7164 & 0.7299 & 0.7180 & 0.7336 & 0.7428 & 0.7448 & 0.7189 \\
   interact    & 0.6215 & 0.7229 & \textbf{0.7253} & 0.7338 & 0.6488 & 0.6242 & 0.6602 & 0.7230 \\
   target      & 0.0815 & 0.1128 & 0.1169 & \textbf{0.1171} & 0.0971 & 0.0918 & 0.0873 & 0.1163 \\
   enzyme      & 0.1903 & 0.2442 & 0.2143 & \textbf{0.2555} & 0.2499 & 0.2031 & 0.1977 & 0.2304 \\
   carrier     & 0.1358 & \textbf{0.2544} & 0.2441 & 0.2526 & 0.1881 & 0.1766 & 0.1266 & 0.1493 \\
   transporter & 0.2242 & \textbf{0.2718} & 0.2189 & 0.2543 & 0.2396 & 0.2042 & 0.2417 & 0.2173 \\\hline
   Avg. (Macro)& 0.2973 & 0.3829 & 0.3705 & \textbf{0.3839} & 0.3216 & 0.3387 & 0.3328 & 0.3549 \\
  \hline
  \end{tabular}
  \caption[Comparison of MRR performance for each method]{Comparison of MRR performance for each method. The MRR for each relational triple and calculated the macro-averaged MRR are summarized. The highest score for each node row is shown in bold.}
  \label{table:link_prediction_result_rel}
\end{table*}

Table~\ref{table:link_prediction_result_rel} shows the comparison of link prediction MRR for each relation edge type, the macro-averaged MRR. 
While a micro-average MRR is calculated by directly calculating the MRR for all instances in the KG without considering the types, a macro-averaged MRR is calculated by first calculating the MRR for each type and then taking the average of the MRR scores. Since the constructed triples are highly imbalanced and the proportion of interact triples is large, models with high prediction performance of relation \textit{interact}  can result in high micro-averaged MRR. The macro-averaged MRR is reported to avoid the effect of this imbalance.
For each scoring function, the comparison of performance between the models with and without text information is shown. 

When the TransE algorithm is used, in the \textit{category} types, the textual models improved MRR but in other relation triple types, the MRR decreased and the averaged MRR also decreased.
Of the three methods that used text, the Initialization by synonyms embeddings method showed the highest macro-averaged MRR.

When the DistMult scoring function is used, the MRR decreased in \textit{interact} and \textit{pathway}, but on the categorical relation \textit{category} and \textit{ATC}, the MRR was improved when the Initialization method is adopted. Initialization methods that use Name information improved the MRR of \textit{target}, \textit{enzyme}, \textit{carrier} and \textit{transporter}, which are the relations between drugs and proteins. The averaged MRR was lower than that of the models without textual information.

When the ComplEx scoring function is used, the MRR decreased in the \textit{interact} and \textit{pathway} relation, while the MRR increased on the categorical relations and relations between drugs and proteins, these are the same tendency as the DistMult algorithm. Especially in the \textit{category} relation, the ComplEx scoring function model without text information has a much lower MRR than TransE or DistMult-based models, but the performance was improved by using text information.
The Initialization and Augmentation methods show higher macro-averaged MRR than the model without text information. 

When the SimplE scoring function is used, 
the model without text information showed the lowest macro-averaged MRR; however,  
the Initialization model that used the Synonyms information showed a higher MRR than the model without text information for all relation types except \textit{pathway}, and showed the highest macro-averaged MRR in all models.
These results showed that it is effective to utilize text information during updating KG embeddings under the SimplE scoring function.

These results show that the utility of textual information for learning KG embeddings depends on the scoring functions and relation types. 
The textual information is always useful in predicting categorical relations such as \textit{category} and \textit{ATC}, while the text information can be harmful for other relations and the utility depends on the scoring functions.
The best settings for each relation type are summarized in Table~\ref{table:summary}. This shows that there is no best single embedding method. The best method to incorporate text information including No Text and the most useful text type also depends on the relation types.

\begin{table}
\centering
\begin{tabular}{llll}
\hline
   &  MRR & Method & Text Information \\\hline
   category  & 0.3668 & SimplE & Initialization+Synonyms \\
   ATC & 0.3820 & SimplE & Initialization+Name \\
   pathway   & 0.7513 & SimplE & No Text \\
   interact & 0.8678 & ComplEx & No Text \\
   target & 0.1243 & ComplEx & Augmentation \\
   enzyme & 0.3590 & TransE & Alignment+Synonyms \\
   carrier & 0.4155 & TransE & No Text \\
   transporter & 0.3576 & TransE & No Text \\
  \hline
  Avg. (Macro)& 0.3839 & SimplE & Initialization+Synonyms\\
  \hline
  \end{tabular}
  \caption{Summary of the best settings for each relation}
  \label{table:summary}
\end{table}

\subsubsection{Analysis of the Data Ombalance of the Constructed KG}
\begin{figure*}[t!]
\begin{center}
\includegraphics[width=\linewidth]{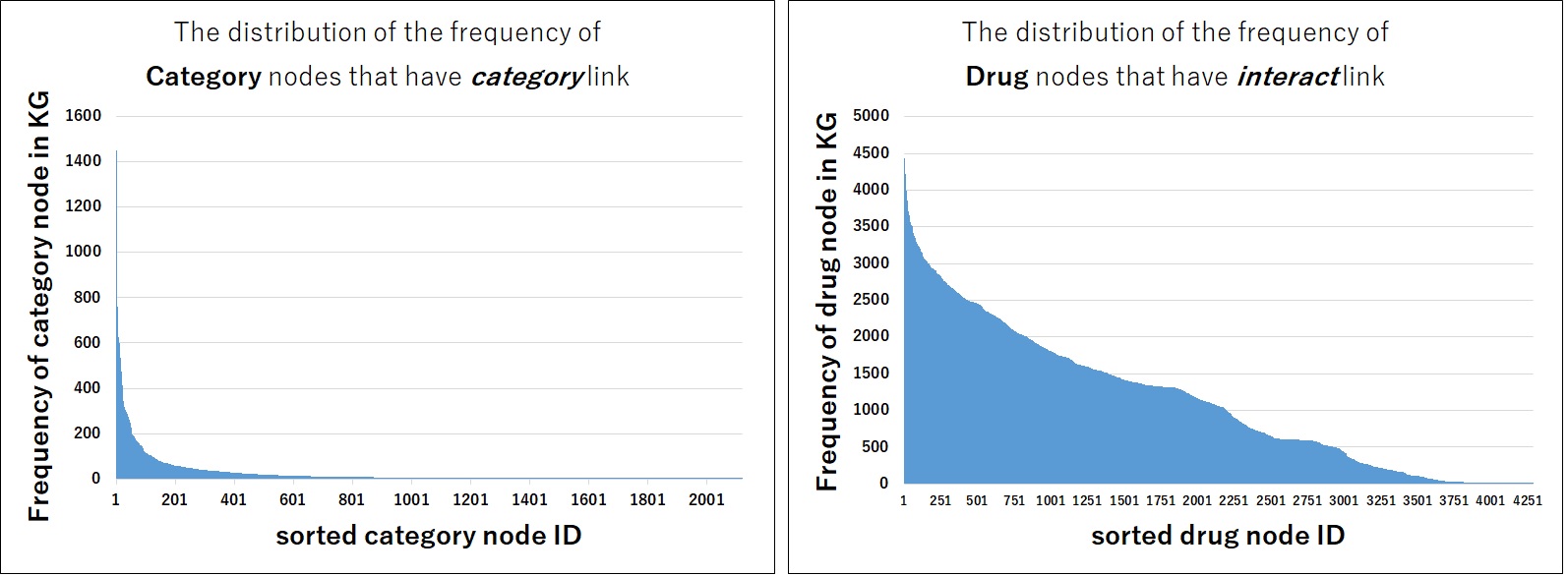}
\end{center}
\caption[The distribution of the frequency of nodes in the train data set]{The distribution of the frequency of nodes in the train data set. The frequencies of category nodes linked by \textit{category} relation are highly imbalanced, while the frequencies of drug nodes linked by \textit{interact} relation are less imbalanced. }\label{fig:freq}
\end{figure*}

Why some models that use text information show lower performance in \textit{interact} and \textit{pathway} relation and show higher performance in categorical relation and drug-protein relation?
In order to analyze these tendencies, The frequency of nodes in the constructed KG is investigated.
Figure~\ref{fig:freq} shows the distribution of the frequencies of category nodes that have the \textit{category} link and drug nodes that have the \textit{interact} link in train triples.
Compared with the distribution of drug nodes frequency, the frequency distribution of category nodes is extremely imbalanced. The distribution shows that a small part of category nodes have a large number of triples between drugs, and many other category nodes have few triples, it could be difficult to predict triples that contain these nodes.
Even if it is difficult to train the representation of nodes from the information of KG structure, it may be possible to predict the correct triples by utilizing the textual embeddings encoded by pre-trained BERT.

\subsubsection{Ablation Study of Augmentation Method}
In the Augmentation method, multiple text items can be considered at the same time. Table~\ref{table:ablation} shows the results of removing each text item. Here, description and synonyms are text items that heterogeneous entities have in common, indication, pharmacodynamics, mechanism-of-action and metabolism are text items that only drug entities have, and gene-name is that only protein entities have.
From Table~\ref{table:ablation}, it can be seen that the averaged MRR becomes lower regardless of which text items are removed, and these results show that all text items are effective for the link prediction task.
In addition, the averaged MRR drops greatly when description or synonyms is excluded, these are the text items that many entities have. The averaged MRR also drops greatly when text information with high coverage is excluded, such as metabolism-of-action.

\begin{table}
\centering
\begin{tabular}{lr}
\hline
   &  Averaged MRR \\\hline
   Full text nodes & \textbf{0.3771} \\
   - description & 0.3626 \\
   - synonyms    & 0.3655 \\
   - indication (drug) & 0.3761 \\
   - pharmacodynamics (drug) & 0.3727 \\
   - mechanism-of-action (drug) & 0.3646 \\
   - metabolism (drug) & 0.3689 \\
   - gene-name (protein) & 0.3754 \\
  \hline
  \end{tabular}
  \caption{Ablation study of text information on Augmentation method (ComplEx score function)}
  \label{table:ablation}
\end{table}

\subsubsection{Effect of Node Type Filtering}
As explained in Sections~\ref{sec:negative_sampling_and_loss_functions}, \ref{sec:task_setting}, \ref{sec:implementation_details}, the model filters out impossible negative instances by restricting the types of entities in the relations.
Table~\ref{table:entity_type_filtering} shows the effect of the entity type filtering. Overall, by performing entity type filtering, the averaged MRR is improved. In particular, in the Augmentation method, entity type filtering is very effective; this is because the Augmentation method adds textual nodes to the graph and is more likely to create an inappropriate negative example during negative sampling.

\begin{table*}
  \small
  \centering
  \begin{tabular}{l|l|r|rrr|rrr|r}
  \hline
   & & No Text & \multicolumn{3}{c|}{Initialization} & \multicolumn{3}{c|}{Alignment} & Aug. \\
   & & & Name & Desc. & Syn. & Name & Desc. & Syno. & \\\hline
   TransE & w/ type filtering & 0.3319 & 0.2930 & 0.2950 & 0.3015 & 0.3337 & 0.3318 & 0.3302 & 0.2883\\
   & w/o type filtering   & 0.2759 & 0.2373 & 0.2265 & 0.2429 & 0.2738 & 0.2722 & 0.2753 & 0.1932 \\\hline
   DistMult & w/ type filtering  & 0.3380 & 0.3206 & 0.2979 & 0.3162 & 0.3261 & 0.3274 & 0.3265 & 0.3011\\
   & w/o type filtering & 0.2217 & 0.2285 & 0.2423 & 0.2547 & 0.2462 & 0.2219 & 0.2464 & 0.1449 \\\hline
   ComplEx & w/ type filtering  & 0.3242 & 0.3681 & 0.3678 & 0.3692 & 0.3577 & 0.3190 & 0.3312 & 0.3771 \\ 
   & w/o type filtering  & 0.2848 & 0.2906 & 0.2887 & 0.2981 & 0.2931 & 0.3052 & 0.3095 & 0.2373 \\\hline
   SimplE & w/ type filtering & 0.2973 & 0.3829 & 0.3705 & 0.3839 & 0.3216 & 0.3387 & 0.3328 & 0.3549\\
   & w/o type filtering & 0.2848 & 0.2906 & 0.2887 & 0.2981 & 0.2931 & 0.3052 & 0.3095 & 0.2373 \\\hline
  \end{tabular}
  \caption{Comparison of averaged MRR performance for w/ (with) and w/o (without) entity type filtering}
  \label{table:entity_type_filtering}
\end{table*}

\begin{table*}[t!]
\centering
\small
\begin{tabular}{ll|l|p{11.8cm}}
\hline
\multicolumn{4}{l}{Examples where textual information is \textbf{helpful}} \\\hline
(a) & \multicolumn{3}{l}{Relation:~\textit{category},~~~textual model rank:1,~~~non-textual model rank:65} \\\hline
&Head & ID & DB13746 (drug entity) \\
&& Name & \textit{Bioallethrin}\\
&& Desc. & \textit{Bioallethrin refers to a mixture of two of the allethrin isomers (1R,trans;1R and 1R,trans;1S) in an approximate ratio of 1:1, where both isomers are active ingredients. A mixture of the two same \underline{stereoisomers}, but in an approximate ratio of R:S in 1:3, is called esbiothrin.} \\
&& Syn. & \textit{Depalléthrine} \\\hline
&Tail & ID & D013237 (category entity) \\
&& Name & \textit{Stereoisomerism}\\
&& Desc. & \textit{The phenomenon whereby compounds whose molecules have the same number and kind of atoms and the same atomic arrangement, but differ in their spatial relationships.}\\
&& Syn. & \textit{Molecular Stereochemistry, Stereochemistry, Molecular, \underline{Stereoisomers}, \underline{Stereoisomer}}\\
\hline\hline
(b)&\multicolumn{3}{l}{Relation:~\textit{ATC},~~~textual model rank:1,~~~non-textual model rank:25} \\\hline
&Head & ID & DB00369 (drug entity) \\
&& Name & \textit{Cidofovir}\\
&& Desc. & \textit{Cidofovir is an injectable \underline{antiviral} medication employed in the treatment of cytomegalovirus (CMV) retinitis in patients diagnosed with AIDS.} \\
&& Syn. & \textit{CDV, Cidofovir anhydrous, Cidofovirum} \\\hline
&Tail & ID & J05A (ATC entity) \\
&& Name & \textit{DIRECT ACTING \underline{ANTIVIRALS}}\\
&& Desc. & ATC entity has no description\\
&& Syn. & ATC entity has no synonyms\\
\hline\hline
\multicolumn{4}{l}{Examples where textual information is \textbf{harmful}} \\\hline
(c) &\multicolumn{3}{l}{Relation:~\textit{pathway},~~~textual model rank:25,~~~non-textual model rank:1} \\\hline
&Head & ID & P51589 (protein entity) \\
&& Name & \textit{Cytochrome P450 2J2}\\
&& Desc. & \textit{This enzyme metabolizes arachidonic acid predominantly via a NADPH-dependent olefin epoxidation to all four regioisomeric cis-epoxyeicosatrienoic acids.} \\
&& Syn. & \textit{1.14.14.1, Arachidonic acid epoxygenase, CYPIIJ2} \\\hline
&Tail & ID & SMP0000695 (pathway entity) \\
&& Name & \textit{Etoricoxib Action Pathway}\\
&& Desc. & \textit{Etoricoxib (also named as Arcoxia) is a COX-2 selective inhibitor. It can be used to treat fever, pain, swelling, inflammation, and platelet aggregation.}\\
&& Syn. & pathway entity has no synonyms\\
\hline\hline
(d) &\multicolumn{3}{l}{Relation:~\textit{interact},~~~textual model rank:4,119,~~~non-textual model rank:1} \\\hline
&Head & ID & DB08893 (drug entity) \\
&& Name & \textit{Mirabegron}\\
&& Desc. & \textit{Mirabegron is a beta-3 adrenergic receptor agonist for the management of overactive bladder. It is an alternative to antimuscarinic drugs for this indication.} \\
&& Syn. & \textit{Mirabegron} \\\hline
&Tail & ID & DB00937 (drug entity) \\
&& Name & \textit{Diethylpropion}\\
&& Desc. & \textit{A appetite depressant considered to produce less central nervous system disturbance than most drugs in this therapeutic category. It is also considered to be among the safest for patients with hypertension.}\\
&& Syn. & \textit{alpha-Benzoyltriethylamine, alpha-Diethylaminopropiophenone, Amfepramone}\\
\hline
\end{tabular}
\caption[The content of the text in the examples where the difference between the rank of textual model and the rank of non-textual model is largest for each relation type]{The content of the text in the examples where the difference between the rank of textual model and the rank of non-textual model is largest for each relation type.
The score function SimplE was used for \textit{category}, \textit{ATC} and \textit{pathway} relation and ComplEx was used for \textit{interact} relation. The Augmentation model was selected as the model with textual information. The highlighted part is the mention common to head and tail entities. The Description and Synonyms are partly excerpted due to space limitations.}
\label{table:content_analysis}
\end{table*}

\subsubsection{Case Study}
As can be seen from Table~\ref{table:summary}, textual information acts harmfully in some relations. In this section,  the content of the text is analyzed to investigate when the text information is harmful or helpful.
Table~\ref{table:content_analysis} shows examples of improved or worsened score ranks on the link prediction task. The examples are where the difference between the rank of the textual model and the rank of the non-textual model is largest, that is, examples where textual information is most useful or harmful for each relation type. In addition, the cases are narrowed down where the better rank is 1. 
In example (a), the highlighted ``\textit{stereoisomers}'' in the description of the drug entity appears in the synonyms of the category entity.
Similarly, in example (b), ``\textit{antiviral}'' in the description of the drug entity appears in the name of the ATC entity. 
The description of the drug entity directly mentions the category in which the drug is included, which is thought to have helped to predict the link of the categorical relation type.

On the other hand, for the examples where the textual information is most harmful, in example (c), the description of protein ``\textit{Cytochrome P450 2J2}'' does not directly mention the ``\textit{Etoricoxib Action Pathway}'' pathway.
In example (d), the description of each drug entity mainly describes the indication of the drug, not the relationship to other drugs.
It is difficult to tell the cause of the poor rank because multiple factors may be involved, but the description of the head entity mainly explains the function and role of the head entity itself, and there is no description that mentions the relationship with the tail entity. This point is considered to be one of the causes of the textual information becoming noise.

\subsection{Summary}

A new heterogeneous knowledge graph containing textual information \OurKG{} from several databases is constructed. The combinations of three methods to use textual information and four scoring functions on the link prediction task are compared.
The utility of text information and the best combination for the link prediction depend on the target relation types.
In addition, when  the averaged MRR is focused on all relation types, a method that combines SimplE and text information achieved the highest MRR, and this result showed the usefulness of text information in the link prediction task in the drug domain.

\section{Integrating Heterogeneous Domain Information for Relation Extraction}

This chapter proposes a novel method that utilizes the heterogeneous KG information for relation extraction.
This chapter includes work from Asada et al. (2022)~\cite{asada-bioinformatics2}.

\subsection{Background}
Chapter 6 integrated multiple databases into a heterogeneous KG and conducted a link prediction task on the heterogeneous KG to obtain representation vectors of the drugs.
This chapter  proposes a novel method that effectively combines the input sentence information and the heterogeneous KG information to extract DDIs from texts.
The model on the data set of the DDIExtraction-2013 task is evaluated to demonstrate the usefulness of heterogeneous KG information.

In Chapter 5, the approach to combining the input sentence representation vector by BERT with the description and molecular structure representation vectors was simply concatenating the respective representation vectors.
This approach is considered insufficient to capture the correlation between the context around the drug mentions in the input sentence and the external knowledge information.

\begin{figure}[t]
\begin{center}
\includegraphics[width=\linewidth]{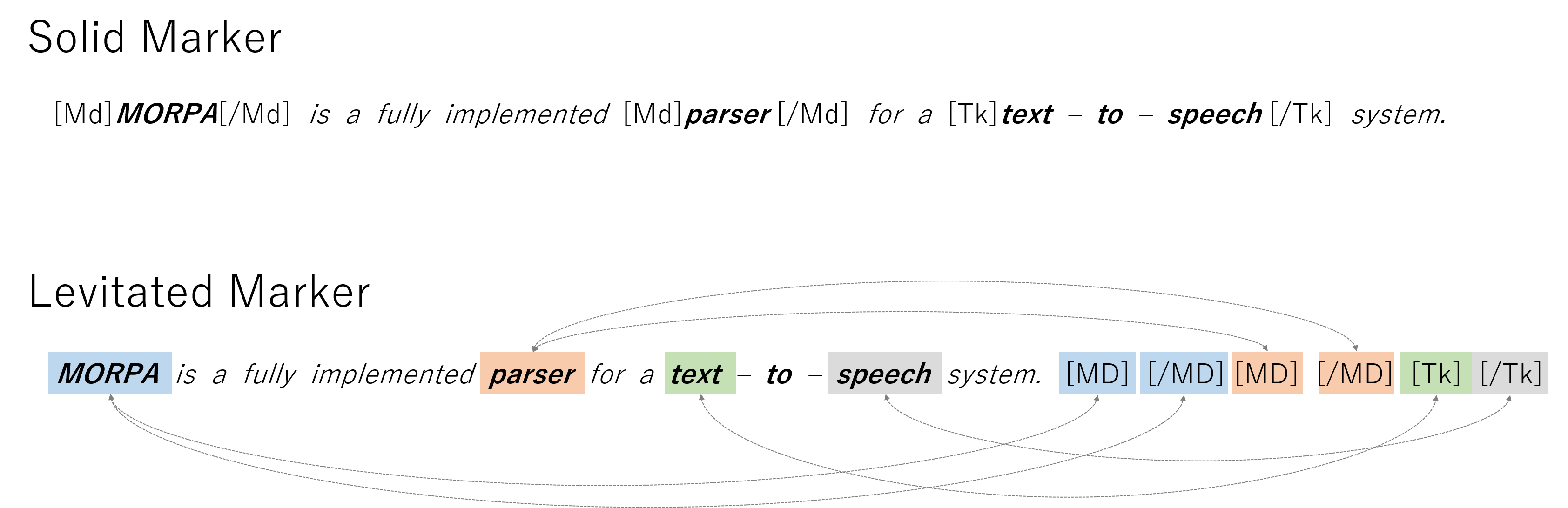}
\end{center}
\caption[Overview of the levitated marker]{Overview of the levitated marker. Tokens of the same color share the position IDs in the levitated marker.}\label{fig:sec7:levitated_marker}
\end{figure}

This chapter uses the idea of ``entity marker'' to devise a model that incorporates the heterogeneous KG embeddings from the lowest layer of BERT and integrally considers the correlation of input sentence information and heterogeneous KG information.
Many relation extraction methods based on the entity marker idea have been studied~\cite{zhang-etal-2019-ernie, peters-etal-2019-knowledge}. The levitated marker model~\cite{zhong-chen-2021-frustratingly} that most inspired the proposed method is described.
Figure~\ref{fig:sec7:levitated_marker} shows the overview of solid marker~\cite{baldini-soares-etal-2019-matching, xiao-etal-2020-denoising} and levitated marker~\cite{zhong-chen-2021-frustratingly}.
Solid marker explicitly inserts two solid markers before and after the span to highlight the span in the input sentence.
Here, [Md] and [TK] stand for Method and Task, and these markers indicate the type of mentions in the sentence. 
The levitated marker first sets the pair of markers to share the same position with the target tokens and then ties a pair of markers by a directional attention.
The levitated marker can identify the mentions without collapsing the original input sentence. The mention and the marker are tied by an attention mechanism of the Transformer model.
The levitated marker embeddings are replaced with the KG embedding to which the target drug mention corresponds, and consider the relationship between the word embeddings and the KG embeddings for DDI extraction.

\subsection{Method}

\subsubsection{Obtaining Heterogeneous KG Embeddings}
In constructing the heterogeneous KG, the augmentation method described in Chapter 6 is used.
In the Augmentation method, textual nodes as well as entity nodes on the heterogeneous KG are placed, and the relational triples in the train data set are augmented.

\begin{figure}[t]
\begin{center}
\includegraphics[width=1.\linewidth]{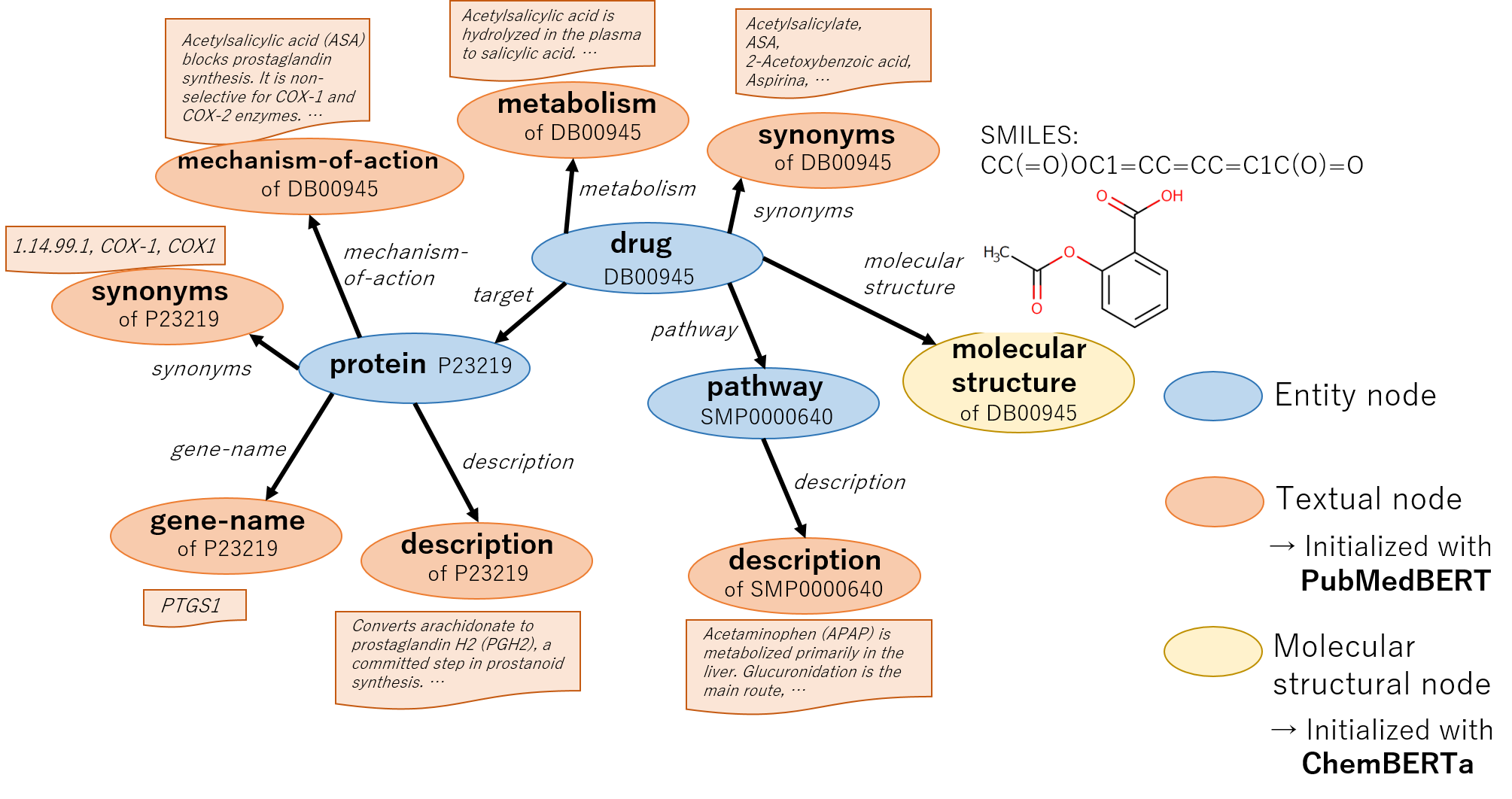}
\end{center}
\caption{Heterogeneous KG with additional drug molecular structure information\label{fig:sec7:mol_hetero_kg}}
\end{figure}

One extension is made from the heterogeneous KG of Chapter 6, that is, the molecular structural nodes of the drugs are added to the heterogeneous KG.
An overview of the newly constructed KG is shown in Figure~\ref{fig:sec7:mol_hetero_kg}.
Similar to initializing textual nodes with embeddings by a pre-trained BERT model, molecular structural nodes with a pre-trained model of the SMILES string coding representation embeddings~\cite{chithrananda2020chemberta} are also initialized.

The constructed pharmaceutical heterogeneous KG enables us to obtain Drug representation vectors that take into account various information such as hierarchical categorical information, interacted protein information, related pathway information, drug textual information, and drug molecular structural information.
In the next section, a novel method for DDI extraction from the literature using the obtained heterogeneous KG representations of drugs is described.

\subsubsection{DDI Extraction from Texts with Heterogeneous KG Embeddings}

\begin{figure}[t]
\begin{center}
\includegraphics[width=1.\linewidth]{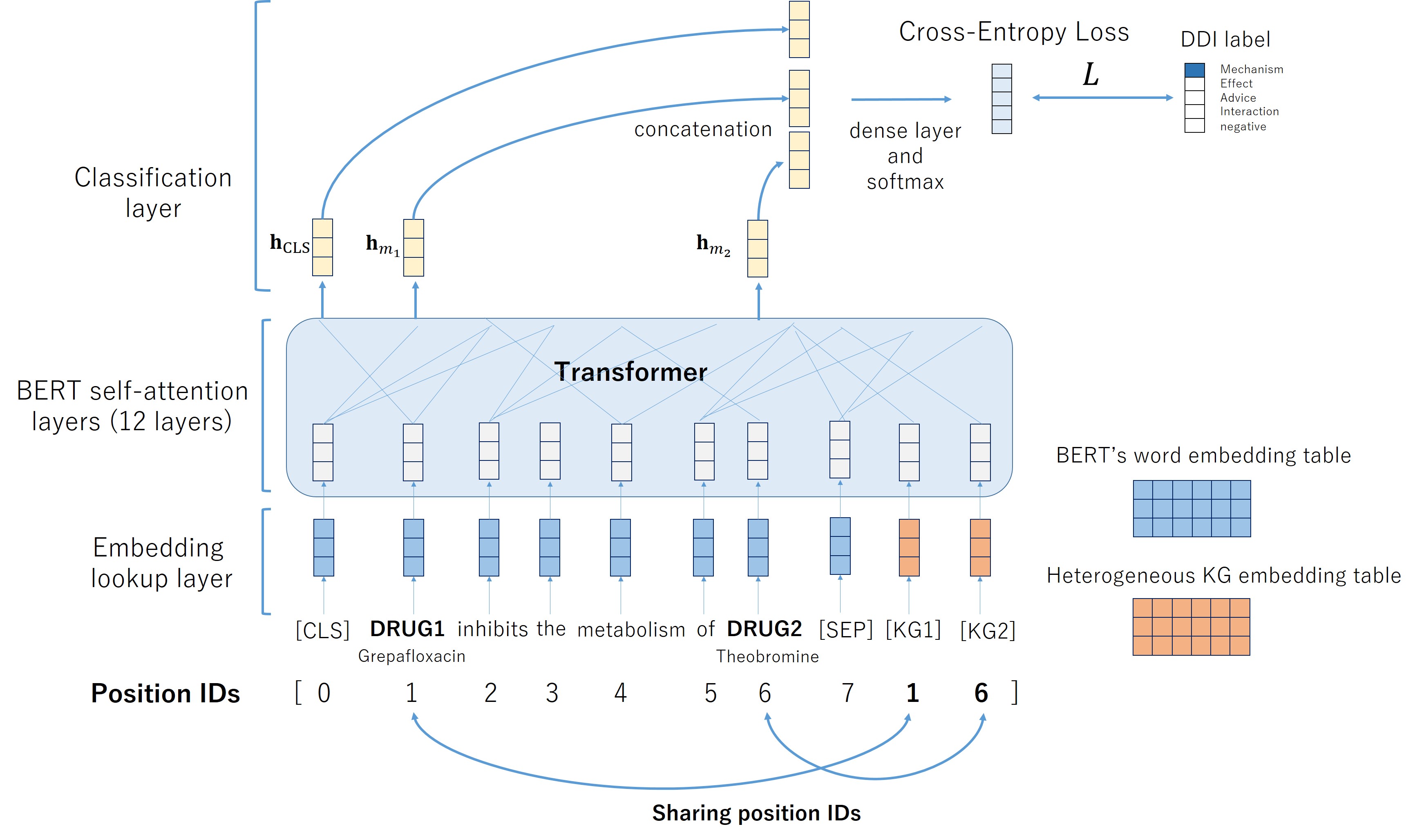}
\end{center}
\caption{The DDI extraction model which utilize heterogeneous information of drugs\label{fig:sec7:our_model}}
\end{figure}

An overview of the DDI extraction model is shown in Figure~\ref{fig:sec7:our_model}.
the proposed model adopts the idea of a levitated entity marker, and two drug mention markers are placed at the end of a sentence.

\paragraph{Embedding Layer}
The input sentence $S$ is tokenized to sub-word units by the BERT tokenizer and converted into the format shown below:
\begin{equation}
    S=\{\texttt{[CLS]}, w_1, w_2, \cdots, w_{m_1}, \cdots, w_{m_2}, \cdots, \texttt{[SEP]}, \texttt{[KG1]}, \texttt{[KG2]} \},
\end{equation}
where $w_i$ is the $i$-th sub-word, and \texttt{[CLS]}, \texttt{[SEP]} are the special tokens of BERT, $m_1$ is the drug mention 1 (\textit{DRUG1}), and $m_2$ is the drug mention 2 (\textit{DRUG2}), and \texttt{[KG1]}, \texttt{[KG2]} are markers for mapping mentions and KG entries.

Then, in the lowest embedding layer of the BERT model, the sub-word $w_i$ and special tokens are looked up from the pre-trained BERT embedding table and converted to embedding vectors.
In addition, the marker embeddings are replaced with the heterogeneous KG embeddings.
All tokens are converted to embedding vectors and the embedding matrix $\bm{W}^0$ of the input sentence is shown as follows:
\begin{equation}
    \bm{W}^0=\{\mathbf{w}_{\texttt{CLS}}, \mathbf{w}_1, \mathbf{w}_2, \cdots, \mathbf{w}_{m_1}, \cdots, \mathbf{w}_{m_2}, \cdots, \mathbf{w}_\texttt{SEP}, \mathbf{w}_\texttt{KG1}, \mathbf{w}_\texttt{KG2} \},
\end{equation}
Here, $\mathbf{w}_i$, $\mathbf{w}_\texttt{CLS}$ and $\mathbf{w}_\texttt{SEP}$ are looked up from the BERT embedding table $\bm{V}_{BERT} \in \mathbb{R}^{N_{v} \times d}$ and $\mathbf{w}_\texttt{KG1}$, $\mathbf{w}_\texttt{KG2}$ are looked up from the heterogeneous KG embedding table $\bm{V}_{KG} \in \mathbb{R}^{N_{e} \times d}$.
$d$ is the dimension of the embedding vector, and $N_v$ is the number of vocabularies of the BERT tokenizer, and $N_e$ is the number of entities in the heterogeneous KG.

\paragraph{Self-Attention Layer}

The embedding matrix $\bm{W}^0$ is the input to the $L$-layers of BERT self-attention module:
\begin{equation}
    \bm{W}^{l+1} = \mathrm{SelfAttention}^{l}(\bm{W}^{l}),
\end{equation}
where $l=0,1,2,\cdots, L-1$.
The output of the final attention layer $\bm{W}^L$ is shown as follows:
\begin{equation}
    \bm{W}^{L}=\{ \textbf{h}_\texttt{CLS}, \textbf{h}_1, \textbf{h}_2, \cdots, \textbf{h}_{m_1}, \cdots, \textbf{h}_{m_2}, \cdots, \textbf{h}_\texttt{SEP}, \textbf{h}_\texttt{KG1}, \textbf{h}_\texttt{KG2} \},
\end{equation}
where $\textbf{h}_i \in \mathbb{R}^d$ is the hidden state vector of $i$-th token.

As shown in Figure~\ref{fig:sec7:our_model}, mention 1 and its KG entity, and mention 2 and its KG entity share the position ID, which ties mention and marker by directional attention.

\paragraph{Prediction Layer}
The loss function is calculated from the hidden representation vectors of the final layer of BERT architecture.
First, the hidden representation of the \texttt{[CLS]} token and two drug mention tokens are concatenated as follows:
\begin{equation}
    \textbf{h}_{all} = [\textbf{h}_\texttt{CLS};\textbf{h}_{m_1};\textbf{h}_{m_2}].
\end{equation}

The concatenated representation vector $\textbf{h}_{all}$ is passed through a dense layer and middle layer representation is obtained,
\begin{equation}
    \textbf{h}_{mid} = \bm{W}^{}_{mid}\textbf{h}_{all}+\textbf{b}^{}_{mid},
\end{equation}
where $\bm{W}_{mid}\in \mathbb{R}^{3d\times d_m}$, $\textbf{b}_{mid} \in \mathbb{R}^{d_m}$ are the trainable weight and bias, and $d_m$ is the dimension of middle layer vector.
Then the middle layer representation is converted into fully-connected representation as follows:
\begin{equation}
    \textbf{h}_{fc} = \bm{W}^{}_{fc}\textbf{h}_{mid}+\textbf{b}^{}_{fc},
\end{equation}
where $\bm{W}_{fc}\in \mathbb{R}^{d_m\times c}$, $\textbf{b}_{fc} \in \mathbb{R}^{c}$ are the trainable weight and bias, and $c$ is the number of label types.
The fully-connected representation vector $\textbf{h}_{fc}$ is converted to probability form by the softmax function:
\begin{equation}
    \hat{\bm{y}}=\mathrm{softmax}(\textbf{h}_{fc}).
\end{equation}
The cross-entropy loss between the prediction probability $\hat{\bm{y}}$ and the gold label $\bm{y}$ is employed,
\begin{equation}
    L=-\sum \bm{y} \log \hat{\bm{y}}.
\end{equation}
The model parameters are updated to minimize the loss $L$.

\subsection{Experimental Settings}

\subsubsection{Mention Linking}
In contrast to the previous chapters, in this chapter, not only with DrugBank drug entities but also with MeSH categorical terms, ATC code categorical terms, and UniProt protein terms are linked.
The DDIExtraction-2013 shared task data set consists of four types of entities, \textsl{DRUG}, \textsl{DRUG\_N}, \textsl{BRAND}, and \textsl{GROUP}. The \textsl{GROUP} type drug mentions may be linked to categorical terms, so the linking coverage is better than in the previous chapters.

As a result, 97.05\% of the unique mentions in train data set were linked to heterogeneous KG entries, and 97.89\% of the unique mentions in test data set were linked.
As for the coverage on instances where mention 1 and mention 2 are both linked, the coverage of train data instances was 91.90\% (25,540~/~27,792), and the coverage of test instances was 90.75\% (5,187~/~5,716).

When the drug mention is not linked to the KG entries, the special tokens (\texttt{KG1} or \texttt{KG2})  are masked and these embeddings are excluded from the attention calculation. 
In this way, the proposed model takes into account KG embedding information for linked examples, and for unlinked examples, the model behaves like the baseline model.

\subsubsection{Link Prediction Settings}
The same train/validation/test split triples is used as the data sets created in Chapter 6 and the link prediction task is conducted.
Mini-batch training using the Adagrad~\cite{john2011adagrad} optimizer is employed.

Hyper-parameter tuning is conducted on the validation data set. Hyper-parameters include initial learning rate and mini-batch size.
As in Chapter 6, TransE, DistMult, ComplEx and SimplE were used for score functions.
The employed hyper-parameters is shown in Table~\ref{tab:sec3:hyper_params}.

SMILES strings are extracted from DrugBank database. The 9,859 drug entities in the heterogeneous KG data set have SMILES strings. Relation triples (drug, \textit{structure}, SMILES) are added to the train data set and molecular structural nodes (SMILES nodes) are initialized by the embedding vectors of the pre-trained SMILES representation language model.

The \texttt{[CLS]} token representation of PubMedBERT~\cite{pubmedbert} is used as the initial value of the textual nodes, and ChemBERTa~\cite{chithrananda2020chemberta} was used as the initial value for the molecular structural nodes.
Chrithranada et al. pre-trained ChemBERTa on 77M unique SMILES from PubChem~\cite{pubchem2021kim}, the world's largest open-source collection of chemical structures. The SMILES were canonicalized and globally shuffled to facilitate large-scale pre-training.
ChemBERTa is based on the RoBERTAa~\cite{roberta2019liu} transformer model. In pre-training, ChemBERTa model masks 15\% of the tokens in each SMILES string. 

\subsubsection{DDI Extraction Settings}

The AdamW optimizer~\cite{loshchilov2018decoupled} is employed, and mixed-precision training~\cite{le2018mixed} is employed for memory efficiency. 
The weight averaging~\cite{ruppert1988efficient, boris1992acceleration} technique is employed, where all model parameters are saved at each update and the model predicts the DDI label from the average of all stored parameters.

PubMedBERT is employed as the textual representation model for the DDI extraction task.
The word embeddings of PubMedBERT and heterogeneous KG embeddings are frozen during training.

Stratified 5-fold cross-validation is conducted on the DDIExtraction-2013 train data set for hyper-parameter tuning.
Hyper-parameters include learning rate, weight decay coefficient, dropout probability and mini-batch size.
The employed hyper-parameters are shown in Table~\ref{tab:sec7:hyper-params}.
The significance results are based on the Randomization test~\cite{fisher1937design}.

\begin{table}[t]
\centering
  \begin{tabular}{lr}\hline
  Parameter & Value \\\hline
  \multicolumn{2}{c}{Link Predction}\\\hline
  Entity embedding size & 768 \\
  Learning rate & 0.25 \\
  Regularization parameter & 9e-9 \\
  Mini-batch size & 8,192 \\\hline
  \multicolumn{2}{c}{DDI Extraction}\\\hline
  Embedding size & 768 \\
  Middle layer size & 768 \\
  Maximum sequence length & 256 \\
  Learning rate & 5e-5 \\
  Regularization parameter & 8e-0 \\
  Dropout probability & 0.5 \\
  Mini-batch size & 128 \\
  \hline
  \end{tabular}
  \caption{Hyper-parameters for link prediction and DDI extraction model
  \label{tab:sec7:hyper-params}}
\end{table}

\subsection{Results and Discussions}
First, the results of link prediction task on the heterogeneous KG are shown in Table~\ref{tab:sec7:LP_results}.
For each of the four score functions, TransE, DistMult, ComplEx and SimplE, the four methods listed below are evaluated:
\begin{description}
    \item [entity nodes only] This is a method that trains only from the nodes contained in the heterogeneous KG. In Figure~\ref{fig:sec7:mol_hetero_kg}, only the actual nodes (the blue one) is included in the train data set. 
    \item [with textual nodes] In this method, in addition to the actual nodes, pseudo nodes that hold textual information such as synonyms and descriptions of entity items are added to the heterogeneous KG. In Figure~\ref{fig:sec7:mol_hetero_kg}, textual nodes (the red one) are added to the actual nodes. 
    \item [with molecular structural nodes] In this method, pseudo molecular structure modes are added to the heterogeneous KG in addition to the actual nodes. As shown in Figure~\ref{fig:sec7:mol_hetero_kg}, molecular structural nodes (the yellow ones) are added.
    \item [with textual nodes and molecular structural nodes] In this method, both textual nodes and molecular structural nodes are added. This approach can consider a wide variety of heterogeneous information about drugs.
\end{description}
Table~\ref{tab:sec7:LP_results} showed that the TransE model performs poorly for both MRR and Hits@k. This should be due to the inability of the TransE model to capture the symmetrical relational triples.
The TransE model showed low performance of MRR and Hits@k because the heterogeneous KG 
contained a large proportion of the (drug, \textit{interact}, drug) triples, which is a symmetric relationship. Furthermore, the TransE model showed the highest MRR and Hits@k when using the ``entity nodes only'' method, meaning that adding textual or molecular structural nodes did not improve link prediction performance.

On the other hand, DistMult, ComplEx, and SimplE, which can consider symmetric relationships, showed higher performance than TransE. These models successfully improved the performance of link prediction task by adding textual nodes and molecular structural nodes, respectively. And when both textual nodes and molecular structural nodes are added, the further performance improvement was achieved.
As shown by the underlined values in Table~\ref{tab:sec7:LP_results}, the highest performance for all MRR and Hits@k metrics was achieved by the method using both textual and molecular structural nodes.
These results show that rich embedding representations are obtained by considering various heterogeneous domain information.

\begin{table}[t]
    \centering
    \begin{tabular}{lrrrr} \hline
          & TransE & DistMult & ComplEx & SimplE \\\hline
         \multicolumn{5}{c}{entity nodes only}\\\hline
         MRR    & \textbf{0.3114} & 0.6732 & 0.6627 & 0.6228\\
         Hits@1 & \textbf{0.0108} & 0.5416 & 0.5436 & 0.3954\\
         Hits@3 & \textbf{0.5590} & 0.7680 & 0.7377 & 0.8287\\
         Hits@10& \textbf{0.7364} & 0.9138 & 0.8892 & 0.9481\\\hline
         \multicolumn{5}{c}{with textual nodes}\\\hline
         MRR    & 0.2894 & 0.7702 & 0.7874 & 0.7175\\
         Hits@1 & 0.0092 & 0.6199 & 0.6424 & 0.5019\\
         Hits@3 & 0.5102 & 0.9104 & 0.9258 & 0.9303\\
         Hits@10& 0.6987 & 0.9703 & 0.9722 & 0.9744\\\hline
          \multicolumn{5}{c}{with molecular structural nodes}\\\hline
         MRR    & 0.3003 & 0.7677 & 0.7313 & 0.7156\\
         Hits@1 & 0.0094 & 0.6171 & 0.5383 & 0.4987\\
         Hits@3 & 0.5352 & 0.9092 & 0.9180 & 0.9307\\
         Hits@10& 0.7195 & 0.9700 & 0.9717 & 0.9746\\\hline
          \multicolumn{5}{c}{with textual nodes and molecular structural nodes}\\\hline
         MRR    & 0.2877 & \underline{\textbf{0.7933}} & \textbf{0.7923} & \textbf{0.7235}\\
         Hits@1 & 0.0091 & \underline{\textbf{0.6610}} & \textbf{0.6509} & \textbf{0.5086}\\
         Hits@3 & 0.5051 & \textbf{0.9166} & \textbf{0.9279} & \underline{\textbf{0.9386}}\\
         Hits@10& 0.6995 & \textbf{0.9711} & \textbf{0.9729} & \underline{\textbf{0.9753}}\\\hline
    \end{tabular}
    \caption[The comparison of link prediction performance on heterogeneous KG]{The comparison of link prediction performance on heterogeneous KG. Figures marked in bold indicate the highest performance when limited to each individual score function. The underlines indicate the highest performance for all score functions.\label{tab:sec7:LP_results}}
\end{table}

\vskip\baselineskip
Then, the performance of DDI extraction models that leverage these heterogeneous KG embeddings is described.
The score functions and hyper-parameters are chosen by the results of 5-fold cross-validation, as described in later sections.
Table~\ref{tab:sec7:DDIE_test_result} shows the performances evaluated on the DDIExtraction-2013 task test set.
The proposed model PubMedBERT+HKG achieved the micro-averaged F-score of 85.40\%, showing the current state-of-the-art performance. The proposed model achieved a significant F-score improvement of 1.70 percent points over the baseline model by using heterogeneous information about drugs.
Compared to other existing models, the PubMedBERT+HKG model showed a higher F-score.
The SciBERT+Mol.+Desc. model is an ensemble of SciBERT+Mol. and SciBERT+Desc.
The proposed model showed higher performance than the ensemble of multiple models.

\begin{table}[t]
    \centering
    \begin{tabular}{llrrr}\hline
         & Method & P & R & F (\%) \\\hline
         Reported scores 
         & CNN~\cite{liu2016drug} & 75.29 & 60.37 & 67.01\\
         & BiLSTM~\cite{sahu2018drug} & 67.77 & 66.80 & 67.28 \\
         & PubMedBERT~\cite{pubmedbert} & - & - & 82.42 \\
         & SciFive-Large~\cite{phan2021scifive} & - & - & 83.67 \\\hline
         The author's implementation & CNN + Mol.~\cite{asada-etal-2018-enhancing} & 73.31 & 71.81 & 72.55\\ 
         & SciBERT + Mol.~\cite{asada-bioinformatics} & 83.57 & 82.12 & 82.84 \\
         & SciBERT + Desc.~\cite{asada-bioinformatics} & 84.05 & 81.81 & 82.91 \\
         & SciBERT + Mol. + Desc.~\cite{asada-bioinformatics} & 85.36 & 82.83 & 84.08\\
         & PubMedBERT (baseline) & 83.45 & 83.96 & 83.70 \\
         & \textbf{PubMedBERT + HKG} & 85.32 & 85.49 & \textbf{85.40}* \\
         \hline 
    \end{tabular}
    \caption[The comparison of DDI extraction performance on DDIExtraction-2013 test data set]{The comparison of DDI extraction performance on DDIExtraction-2013 test data set. * indicates performance improvement from PubMedBERT (baseline) at a significance level of $p<0.001$.\label{tab:sec7:DDIE_test_result}}
\end{table}

Then, the F-scores for each of the four DDI labels of DDIExtraction-2013 task data set are shown in Table~\ref{tab:sec7:DDIE_test_result_every_type}.
As shown in Table~\ref{tab:sec7:DDIE_test_result_every_type}, the model with heterogeneous KG information improves F-scores for all DDI types.
In particular, the F-score greatly improves for \textsl{Mechanism} relation. HKG model improved F-score by 3.80 percent points from the baseline model.

\begin{table}[t]
\centering
\begin{tabular}{lrrrrrr}\hline
 Method & & \textsl{Mech.} & \textsl{Effect} & \textsl{Adv.} & \textsl{Int.} & Avg. (\%) \\\hline
PubMedBERT (baseline)           & P & 87.12 & 87.18 & 78.71 & 82.69 & 83.93 \\
                                & R & 85.10 & 92.31 & 88.33 & 44.79 & 77.63 \\
                                & F & 86.10 & 89.67 & 83.25 & \textbf{58.11} & 79.28 \\\hline
\textbf{PubMedBERT + HKG}  & P & 88.96 & 88.84 & 81.06 & 81.13 & 84.99\\
                                & R & 88.08 & 93.67 & 89.17 & 44.79 & 78.92\\
                                & F & \textbf{88.52} & \textbf{91.19} & \textbf{84.92} & 57.72 & \textbf{80.58}\\\hline 
\end{tabular}
\caption{The comparison of F-scores for individual DDI types and macro-averaged F-score on DDIExtraction-2013 test data set\label{tab:sec7:DDIE_test_result_every_type}}
\end{table}

\subsubsection{Selecting Score Functions}

\begin{table}[t]
    \centering
    \begin{tabular}{lrrr}\hline
         Method & P & R & F (\%) \\\hline
         baseline               & 82.19 & 83.23 & 82.54\\
         + HKG (TransE)    & 83.46 & 84.37 & 83.82\\
         + HKG (DistMult)  & 83.68 & 85.42 & \textbf{84.48}\\
         + HKG (ComplEx)   & 83.68 & 84.32 & 83.90\\
         + HKG (SimplE)    & 83.46 & 84.50 & 83.86\\
         \hline
    \end{tabular}
    \caption[The comparison of DDI extraction performances with different score functions for training KG embeddings]{The comparison of DDI extraction performances with different score functions for training KG embeddings. the author performs 5-fold cross-validation and show the average of F-scores over the five validation data sets.\label{tab:sec7:cv5_scorefunc}}
\end{table}

In this section, the author will discuss which score function was effective for DDI extraction.
Table~\ref{tab:sec7:DDIE_test_result} shows the average F-scores for the five validation data sets for each score function.
F-score is higher than the baseline model when using heterogeneous KG embeddings trained by any of score functions.
The improvement of F-score points from the baseline model is 1.28 for the TransE model and 1.94 for the DistMult model and 1.36 for the ComplEx and 1.32 for the SimplE model.
From these results, the DistMult score function is adopted for the DDI extraction model.
The DistMult model preformed best on the link prediction task on MRR, Hits@1, and also showed the best performance on the DDI extraction task.

\subsubsection{Ablation Study on Model Architecture}

\begin{table}[t]
    \centering
    \begin{tabular}{lrrrr}\hline
         Method & P & R & F (\%) & $\Delta$ (pp) \\\hline
         Full model                     & 83.68 & 85.42 & 84.48 & -\\
         w/o sharing position ids       & 82.49 & 86.04 & 84.18 & 0.30\\
         w/o freezing KG embeddings     & 83.69 & 84.32 & 83.90 & 0.58\\
         w/o \texttt{CLS} representation& 82.63 & 85.17 & 83.81 & 0.67\\
         w/o mention representation     & 82.07 & 85.69 & 83.78 & 0.70\\
         w/o KG embeddings (baseline)   & 82.19 & 83.23 & 82.54 & 1.94 \\
         \hline 
    \end{tabular}
    \caption[The ablation study on model architecture]{The ablation study on the model architecture. The performance with 5-fold cross-validation on the training set is showed.\label{tab:sec7:ablation}}
\end{table}

This section provides ablation studies. 
Table~\ref{tab:sec7:ablation} shows the ablation study results on 5-fold cross-validation data sets.

\paragraph{w/o Sharing Position IDs}
First, the case excluding the sharing of position IDs is discussed.
The sharing of position IDs has the effect of linking the mention embeddings and KG embeddings.
As shown in Figure~\ref{fig:sec7:our_model}, with position sharing, the position ID of \texttt{[KG1]} is 1 of the drug mention 1 and the position ID of \texttt{[KG2]} is 6 of the drug mention 2. When this sharing is disabled, the position IDs of \texttt{[KG1]} and \texttt{[KG2]} are the values following from the IDs of the \texttt{[SEP]} token.
From Table~\ref{tab:sec7:ablation}, when position sharing is excluded, the F-score is reduced by 0.30 percent points from the full model.

\paragraph{w/o freezing KG embeddings}
In the proposed model, the BERT embeddings the KG embeddings are frozen and the attention weight is trainable.
Table~\ref{tab:sec7:ablation} shows that when embedding freezing is disabled, 0.58 percent points of F-score is lower than the full model.
The author thinks the reason embedding freezing is effective is that if the embedding is not frozen, there will be a gap between the KG embeddings when drugs that appear on the train set and those that appear only on the test set.

\paragraph{w/o \texttt{CLS} representation}
When $\textbf{h}_\texttt{CLS}$ was excluded from the input vector of the middle layer, the F-score decreased by 0.67 percent points.
The $\texttt{CLS}$ token representation holds information of the entire sentence. It is effective to use the representation of the \texttt{[CLS]} token.

\paragraph{w/o mention representation}
When $\textbf{h}_{m_1}$ and $\textbf{h}_{m_2}$ were excluded from the input vector of the middle layer, the F-score decreased by 0.70 percent points.
In addition to the \texttt{[CLS]} token representation, it is effective to use the drug mention representation.

\paragraph{w/o KG embeddings}
Finally, the case without KG embeddings is discussed.
In the baseline model, \texttt{[KG1]} and \texttt{[KG2]} tokens are not fed to the BERT architecture and KG embeddings are not used.
Except for this point, the model structure is the same as the proposed model.

The use of heterogeneous KG information increased the F-score by 1.94 percent points. This result showed that the drug heterogeneous information is useful for DDI extraction from literature.

\begin{table}[t]
    \centering
    \begin{tabular}{lrrrr}\hline
         Method & P & R & F (\%) & $\Delta$ (pp) \\\hline
         PubMedBERT + HKG             & 83.68 & 85.42 & 84.48 & -\\
         w/o Protein nodes            & 83.22 & 85.04 & 84.08 & 0.40\\
         w/o MeSH category nodes      & 83.28 & 83.75 & 83.38 & 1.10\\
         w/o ATC nodes                & 83.33 & 84.27 & 83.73 & 0.75\\
         w/o Pathway nodes            & 83.98 & 84.85 & 84.30 & 0.18\\
         w/o Textual nodes            & 83.75 & 84.20 & 83.86 & 0.62\\
         w/o Molecular structure nodes& 83.36 & 84.30 & 83.71 & 0.77\\
         \hline
    \end{tabular}
    \caption[The ablation study on node types]{The ablation study on node types. The performance with 5-fold cross-validation on the training set is showed.\label{table:ablation_entity_type}}
\end{table}
\subsubsection{Ablation Study on Heterogeneous KG Node Types}
Table~\ref{table:ablation_entity_type} shows the ablation study on the effect of individual KG node type.
As shown in Table~\ref{table:ablation_entity_type}, all types of nodes in the heterogeneous KG contribute to the performance improvement of DDI extraction from the literature.
The results show the importance of simultaneously considering multiple pieces of drug-related information.
Among the types of nodes, the MeSH categorical information contributed most to the performance improvement while the pathway information contributed least. 

\subsubsection{Verification of DDI Label Leakage from KGs}
The constructed heterogeneous KG contains triples of \textit{interact} relations between Drug nodes and Drug nodes.
This section examines whether the \textit{interact} triples directly affects the performance improvement in DDI extraction from the literature and whether there is a leakage of DDI labels.
First, the percentage of drug pairs in the DDIExtraction-2013 train data set that are registered as \textit{interact} relations in the KG is shown.
39.79\% (11,060 / 27,792) of the drugs pairs are included in the constructed KG and others are not included in the KG.
It should be noted that the DDI labels in the DDIExtraction-2013 data set are \textsl{Mechanism}, \textsl{Effect}, \textsl{Advice} and \textsl{Int.}, whereas there is only on relation label \textsl{interact} in the KG.
Table~\ref{table:leakage} shows a comparison of DDI extraction performance for drug pairs that are included in the KG and drug pairs that are not included in the KG.
As shown in Table~\ref{table:leakage}, both for drug pairs that are registered in the KG and drug pairs that are not registered in the KG, the proposed model (baseline + HKG) outperformed the baseline model.
When the pairs are included in the KG, the improvement of F-score is 1.1 percent, 0.97 pp, whereas when the pairs are not included in the KG, the improvement of F-score is 3.4 percent, 2.79 pp.
Therefore, the performance improvement of the proposed model is greater when the drug pairs is not registered in the KG than when they are registered in the KG.
The author believes that these results indicate that the proposed model does not improve DDI extraction performance because of the leakage of DDI labels from the constructed KG.

\begin{table}[t]
    \centering
    \begin{tabular}{lrrr}\hline
         Method & P & R & F (\%) \\\hline
         All drug pairs \\ \hline
         baseline                   & 82.19 & 83.23 & 82.54 \\
         baseline + HKG             & 83.68 & 85.42 & 84.48\\ \hline \\ 
         Drug pairs that are included in KGs as \textit{interact} triples \\\hline
         baseline                   & 85.11	& 85.74	& 85.25 \\
         baseline + HKG             & 85.94	& 86.73	& 86.22 \\ \hline\\ 
         
         Drug pairs that are NOT included in KGs as \textit{interact} triples \\\hline
         baseline                   & 79.83	& 80.94	& 80.23 \\
         baseline + HKG             & 81.83	& 84.32	& 83.02 \\ 
         \hline
    \end{tabular}
    \caption{The DDI extraction performance for drug pairs that are included in the constructed heterogeneous KG and drug pairs that are not included in KG\label{table:leakage}}
\end{table}

\subsubsection{Learning Curve}
\begin{figure}[t]
\begin{center}
\includegraphics[width=.9\linewidth]{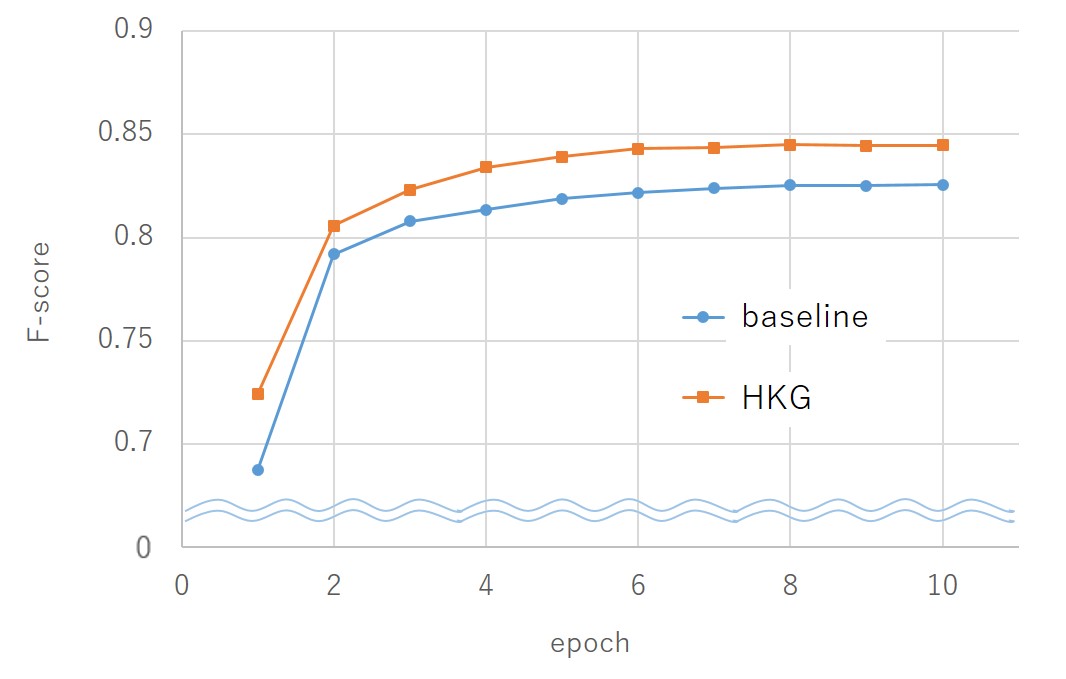}
\end{center}
\caption[Learning curve of baseline model and the proposed model on 5-fold cross-validation data sets]{Learning curve of baseline model and the proposed model on 5-fold cross-validation data sets. 
The average of F-scores over the five validation data sets is reported.\label{fig:sec7:learning_curve}}
\end{figure}

The learning curve of baseline model and proposed model is shown in Figure~\ref{fig:sec7:learning_curve}.
Figure~\ref{fig:sec7:learning_curve} plots the average of the F-scores of the five validation data sets per epoch.
Learning curves show that the proposed method always has a higher F-score than the baseline model.
From the 4th epoch, the model using KG embeddings outperforms the baseline in F-score more than 2 percent points.

\subsubsection{Analysis of Prediction Results}

\begin{figure*}[t]
    \centering
    \includegraphics[width=0.9\linewidth]{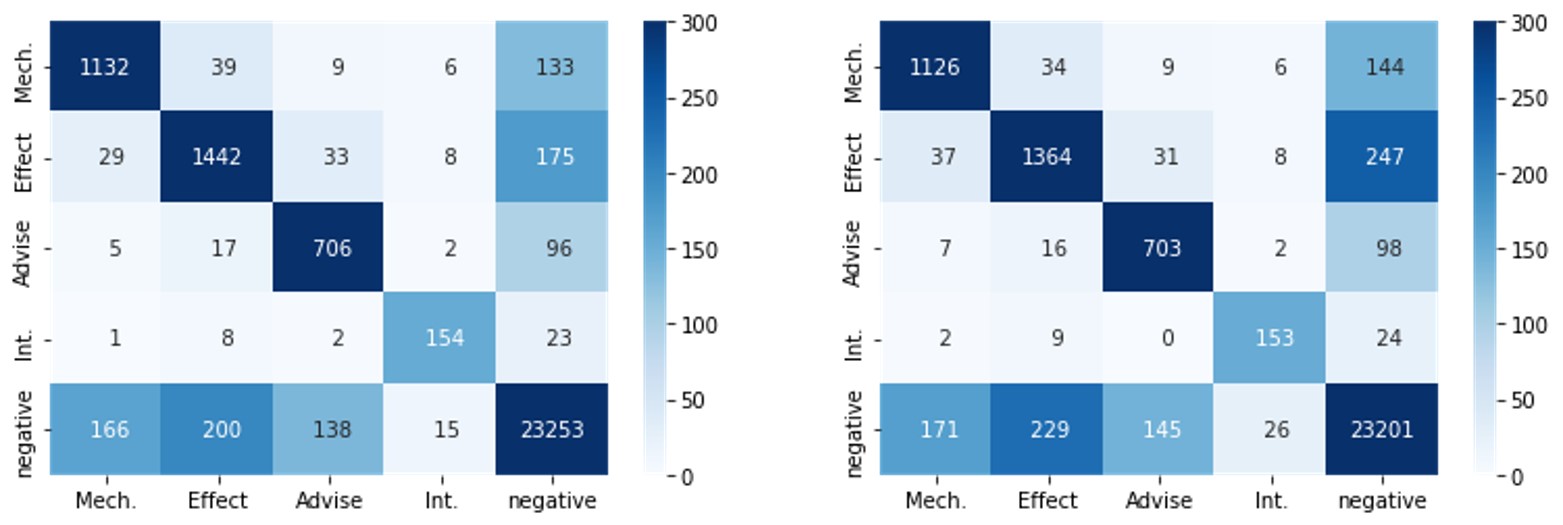}
    \caption[The confusion matrices (left: baseline, right: proposed method)]{The confusion matrices (left: baseline, right: proposed method). The vertical axis shows the actual labels and the horizontal axis shows the predicted labels. The numbers indicates the total count of five validation data sets.
    \label{fig:sec7:heatmap}}
\end{figure*}

The confusion matrices of the baseline and the proposed model are shown in Figure~\ref{fig:sec7:heatmap}.
The numbers indicate the total count of five validation data sets.
The proposed method reduced the all patterns of errors (non-diagonal components in the table) compared to the baseline model.
In particular, errors in which the model incorrectly predicts the \textsl{Effect} interaction as negative and errors in which the model incorrectly predicts the negative as \textsl{Effect} are greatly reduced.
For \textsl{Mechanism}, \textsl{Advice}, and \textsl{Int.} relations, the use of heterogeneous KG information also reduced the number of cases of false negative or false positive relations.
On the other hand, there were cases in which the use of heterogeneous KG information slightly increased errors by classifying relations into wrong types, e.g., incorrectly classifying a \textsl{Mechanism} relation as \textsl{Effect}.

In addition, four examples of prediction results are shown.
Example 1, 2, and 3 in Table~\ref{tab:case_study} are cases correctly predicted by using heterogeneous KG information but incorrectly predicted by the baseline model.
In Examples 1 and 2, many drug entities appear, and DRUG1 and DRUG2 are included in parentheses. 
As shown in Example 3, the baseline model tends to predict the cases where the distance between DRUG1 and DRUG2 is extremely short as negative, but the proposed model correctly predicts them.
From these examples, heterogeneous KG representation of the drug entities may be helpful to predict correct relations when the prediction is difficult only from their surrounding contexts in the sentences.

Example 4 is a case incorrectly predicted by using heterogeneous KG information while correctly predicted by the baseline model.
According to the annotation guideline of DDIExtraction-2013 data set, an interaction should only be annotated when it occurs in the text. Example 4 shows some studies about given interactions were performed, however, the sentence does not provide any evidence.
In such a case, background knowledge of drugs may have disturbed correct prediction.

\begin{table*}[t]
    \centering
    \begin{tabular}{p{16cm}}
        \hline
        Example 1 \\ 
        \textbf{Text}: \textit{In patients receiving nonselective DRUGOTHER (DRUGOTHER) (e.g., \textbf{DRUG1}) in combination with DRUGOTHER (e.g., DRUGOTHER, DRUGOTHER, DRUGOTHER, DRUGOTHER, \textbf{DRUG2}), there have been reports of serious, sometimes fatal, reactions.} \\
        DRUG1: \textit{selegiline hydrochloride}, DRUG2: \textit{venlafaxine}\\
        \textbf{Gold label}: Effect~~~~\textbf{Baseline}: negative~~~~\textbf{Proposed model} Effect: 
        \\\hline 
        Example 2 \\
        \textbf{Text}: \textit{DRUGOTHER: In a study of 7 healthy male volunteers, \textbf{DRUG1} treatment potentiated the blood glucose lowering effect of DRUGOTHER (a DRUGOTHER similar to \textbf{DRUG2}) in 3 of the 7 subjects.} \\
        DRUG1: \textit{acitretin}, DRUG2: \textit{chlorpropamide}\\
        \textbf{Gold label}: negative~~~~\textbf{Baseline}: Effect~~~~\textbf{Proposed model} negative: 
        \\\hline
        Example 3 \\ 
        \textbf{Text}: \textit{Caution should be exercised when considering the use of \textbf{DRUG1} and \textbf{DRUG2} in patients with depressed myocardial function.} \\ 
        DRUG1: \textit{BREVIBLOC}, DRUG2: \textit{verapamil}\\
        \textbf{Gold label}: Advice~~~~\textbf{Baseline}: negative~~~~\textbf{Proposed model} Advice: 
        \\\hline 
        Example 4 \\
        \textbf{Text}: \textit{To determine whether \textbf{DRUG1} has a direct effect on the distribution of \textbf{DRUG2}, the elimination and distribution of DRUGOTHER was studied in six patients, five lacking kidney function and one with a partially impaired renal function, in the presence or absence of DRUGOTHER.} \\
        DRUG1: \textit{probenecid}, DRUG2: \textit{cloxacillin}\\
        \textbf{Gold label}: negative~~~~\textbf{Baseline}: negative~~~~\textbf{Proposed model} Mechanism: 
        \\\hline
    \end{tabular}
    \caption{Case studies of the proposed model}
    \label{tab:case_study}
\end{table*}

\subsection{Summary}
This chapter first added the molecular structure information of drugs to the KG data set constructed in the previous chapter. The results on the link prediction task showed that the MRR and Hits@k was improved by considering the molecular structure information of drugs.

Then, a method to use the heterogeneous KG representations for the relation extraction task was proposed.
The proposed model incorporates heterogeneous KG embeddings into the input sentence in the form of levitated markers and considers the relationship between contexts and KG information through an attention mechanism.
In the experiment, a significant improvement of 1.70 percent points on the DDIExtraction-2013 data set was achieved by using heterogeneous KG information.

\section{Conclusions} \label{sec:conclusions}
\subsection{Summary}
This thesis proposed a novel approach that focuses on heterogeneous domain information for relation extraction from the literature. In case studies, the proposed method can perform DDI extraction integrally considering heterogeneous information about drugs, and showed the usefulness of utilizing various information about drugs for the DDI extraction task. 
After describing an attention mechanism for
relation extraction in Chapter 3 and reviewing an approach to using a single kind of domain information in Capter 4, this thesis devised neural relation extraction models that can consider heterogeneous domain information.

Chapter 5 proposed a novel neural method for relation extraction from text using large-scale raw text information and two kinds of domain information, which are the drug descriptions and the drug molecular structure information. The results show that the large-scale raw text information with SciBERT greatly improves the performance of DDI extraction from the literature on the DDIExtraction-2013 data set.
In addition, either of the drug description and the molecular structure information can further improve the performance for specific DDI types, and their simultaneous use can improve the performance on all the DDI types. 

Chapter 6 constructed a new heterogeneous knowledge graph containing textual information \OurKG{} from several databases. The combinations of three methods to use textual information and four scoring functions on the link prediction task are compared.
The utility of text information and the best combination for the link prediction depends on the target relation types were found.
In addition, when the averaged MRR for all relation types was in focus, a method that combines SimplE and text information achieved the highest MRR, and this result showed the usefulness of text information in the link prediction task in the drug domain.

Chapter 7 proposes a neural architecture that integrates sentence information and knowledge graph information constructed in Chapter 6. The molecular structure information of drugs is added to the KG data set constructed in the previous chapter. The results on the link prediction task showed that the MRR and Hits@k were improved by considering the molecular structure information of drugs.
Then, a method to use the heterogeneous KG representations for the relation extraction task was proposed.
The proposed model incorporates heterogeneous KG embeddings into the input sentence in the form of levitated markers and considers the relationship between contexts and KG information.
In the experiment, an improvement of 1.70 percent points on the DDIExtraction-2013 data set was achieved by adding heterogeneous KG information.

\subsection{Future Work}
\subsubsection{Employing the Deep Neural Entity Linking Method}
As mentioned in Chapter 7, the coverage of linking drug mentions in the input sentence to drug entries in the KG is only about 90\%, which has become a bottleneck in the overall DDI extraction model.
In recent years, many deep neural entity linking models~\cite{dongfang2022asimple} have been proposed, and these models can classify the input mentions into KG entries even if the number of KG entries is more than several hundred thousand.
The deep neural linking model has been reported to perform better than the model with simple string matching, and 
higher coverage and more accurate linking are expected by employing the deep neural linking model in the model.

In addition, when the issues about memory usage and training time are solved, a method that jointly train entity linking and DDI extraction can be realized.
These ideas are left for future work.

\subsubsection{Joint Learning of BERT and KG Embeddings}
In the proposed method, the learning of KG embeddings is decoupled from the learning of DDI extraction.
If the KG embeddings and BERT model are jointly updated, it is expected that DDI label information can be taken into account when updating KG embeddings and the learning KG representation can be more suitable for DDI extraction.

Many GNN-based text classification methods~\cite{yao2019graph, zhang-etal-2020-every} have been proposed that perform the sentence classification task by placing words in the input sentence on a graph space and learning the graph representation.
A possible extension of these models is to place the input sentences in a heterogeneous KG and train the word/sentence entities and other heterogeneous entities jointly to perform DDI extraction.

\subsubsection{End-to-End DDI Extraction}
DDI extraction from the literature consists of two parts: drug mention recognition and DDI extraction between the mention pair, and in this thesis, the author focuses on the DDI extraction part.
Efficient simultaneous training of the two tasks may improve the performance of both tasks.
It is important to analyze the impact of using heterogeneous information in the DDI extraction part on the mention recognition part.

Furthermore, drug mention recognition is also closely correlated with drug mention linking to KG entries.
KG entries information can help to improve the drug mention recognition performance.
The construction of a framework of combining the drug mention recognition part, drug mention linking part and DDI extraction part is left for future work.

\subsubsection{Extension to Other Tasks}
This thesis limits the experiment to DDI extraction and drug-protein interaction extraction.
The main idea of the proposed model consists of two parts: representation of heterogeneous items in the form of KG and integration of representation embeddings by an attention mechanism.
The proposed method can be extended to other tasks, since the method is not domain specific and domain information from many other fields can be applied to the input format of the model.
The author would like to apply the proposed model to other tasks and test the versatility of the method.

\section*{Publications}
\begin{itemize}
    \item \textbf{Using drug descriptions and molecular structures for drug-drug interaction extraction from literature}~\cite{asada-bioinformatics}\\
    Masaki Asada, Makoto Miwa, and Yutaka Sasaki, Bioinformatics volume 37, issue 12, pages 1739-1746, 2021.
    \item \textbf{Representing a heterogeneous pharmaceutical knowledge-graph with textual information}~\cite{Asada2021-uv}\\
    Masaki Asada, Nallappan Gunasekaran, Makoto Miwa, and Yutaka Sasaki, Frontiers in Research Mtrics and Analytics, volume 6, pages 1-13, 2021.
    \item \textbf{TTI-COIN at BioCreative VII Track 1 - Drug-protein interaction extraction with external database information}~\cite{iinuma-etal-2021-tticoin}\\
    Naoki Iinuma, Masaki Asada, Makowo Miwa, and Yutaka Sasaki, In Proceedings of BioCreative VII workshop, volume 1, pages 49-53, 2021.
    \item \textbf{Integrating heterogeneous knowledge graphs into drug-drug interaction extraction from the literature}~\cite{asada-bioinformatics2}\\
    Masaki Asada, Makoto Miwa, and Yutaka Sasaki, Bioinformatics, 2022.
\end{itemize}

\clearpage

\bibliography{references}
\bibliographystyle{unsrt}

\end{document}